\documentclass{article}

\PassOptionsToPackage{numbers, compress}{natbib}

\usepackage[preprint]{neurips_2023}

\usepackage[utf8]{inputenc} 
\usepackage[T1]{fontenc}    
\usepackage{url}            
\usepackage{booktabs}       
\usepackage{amsfonts}       
\usepackage{nicefrac}       
\usepackage{xcolor}         
\usepackage{graphicx}
\usepackage{multirow}
\usepackage{tikz}
\usepackage[edges]{forest}
\usepackage{natbib}
\definecolor{hidden-draw}{RGB}{20,68,106}
\definecolor{hidden-pink}{RGB}{255,245,247}
\usepackage{amsmath}
\usepackage{setspace}
\usepackage{makecell}
\usepackage{graphicx}
\usepackage{arydshln}
\usepackage{colortbl}
\usepackage{mathtools}
\usepackage{bm}
\usepackage{algorithm}
\usepackage{algorithmic}
\usepackage{amsmath}
\usepackage{amssymb}
\usepackage{bbm}
\definecolor{mycolor_blue}{HTML}{E7EFFA}
\definecolor{mycolor_green}{HTML}{E6F8E0}
\definecolor{mycolor_gray}{HTML}{ECECEC}
\definecolor{pearDark}{HTML}{2980B9}
\usepackage[pagebackref=false,breaklinks=true,colorlinks=True,urlcolor=blue,citecolor=pearDark,bookmarks=false]{hyperref}
\usepackage{cleveref}
\usepackage{graphicx} 		
\usepackage{subfigure}		
\usepackage{pifont}

\title{
\raisebox{-1.1ex}{\protect\includegraphics[height=2.7\fontcharht\font`\B]{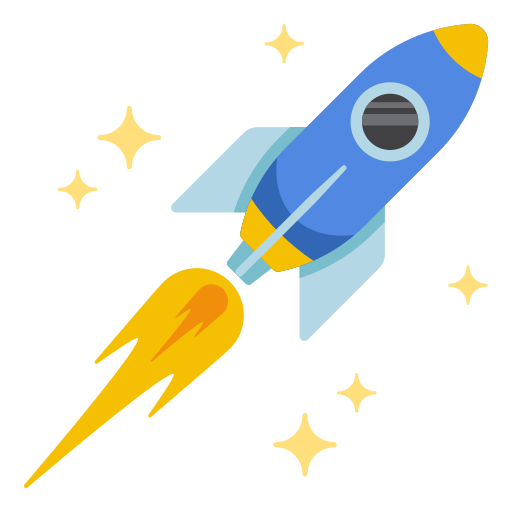}}
Efficient Diffusion Models: A Comprehensive Survey from Principles to Practices}

\author{%
      Zhiyuan Ma \textsuperscript{1}\thanks{Zhiyuan Ma is the project leader.},
      Yuzhu Zhang\textsuperscript{2},
      Guoli Jia\textsuperscript{1},
      Liangliang Zhao\textsuperscript{1},
      Yichao Ma\textsuperscript{2},
  \\ 
  \textbf{
      Mingjie Ma\textsuperscript{2},
      Gaofeng Liu\textsuperscript{3},
      Kaiyan Zhang\textsuperscript{1},
      Jianjun Li\textsuperscript{2},
      Bowen Zhou\textsuperscript{1,4}\thanks{Corresponding author.}   
  }
  \\
  \\
  \textsuperscript{1}Tsinghua University,
  \textsuperscript{2}HUST,
  \textsuperscript{3} SJTU,
  \textsuperscript{4} Shanghai AI Lab
  \\
}

\begin{document}

\maketitle

\begin{abstract}
As one of the most popular and sought-after generative models in the recent years, diffusion models have sparked the interests of many researchers and steadily shown excellent advantage in various generative tasks such as image synthesis, video generation, molecule design, 3D scene rendering and multimodal generation, relying on their dense theoretical principles and reliable application practices. The remarkable success of these recent efforts on diffusion models comes largely from progressive design principles and efficient architecture, training, inference, and deployment methodologies. However, there has not been a comprehensive and in-depth review to summarize these principles and practices to help the rapid understanding and application of diffusion models. In this survey, we provide a new efficiency-oriented perspective on these existing efforts, which mainly focuses on the profound principles and efficient practices in architecture designs, model training, fast inference and reliable deployment, to guide further theoretical research, algorithm migration and model application for new scenarios in a reader-friendly way. \url{https://github.com/ponyzym/Efficient-DMs-Survey}

\end{abstract}




\section{Introduction}
\begin{figure}[!t]
\centering
\includegraphics[width=\linewidth]{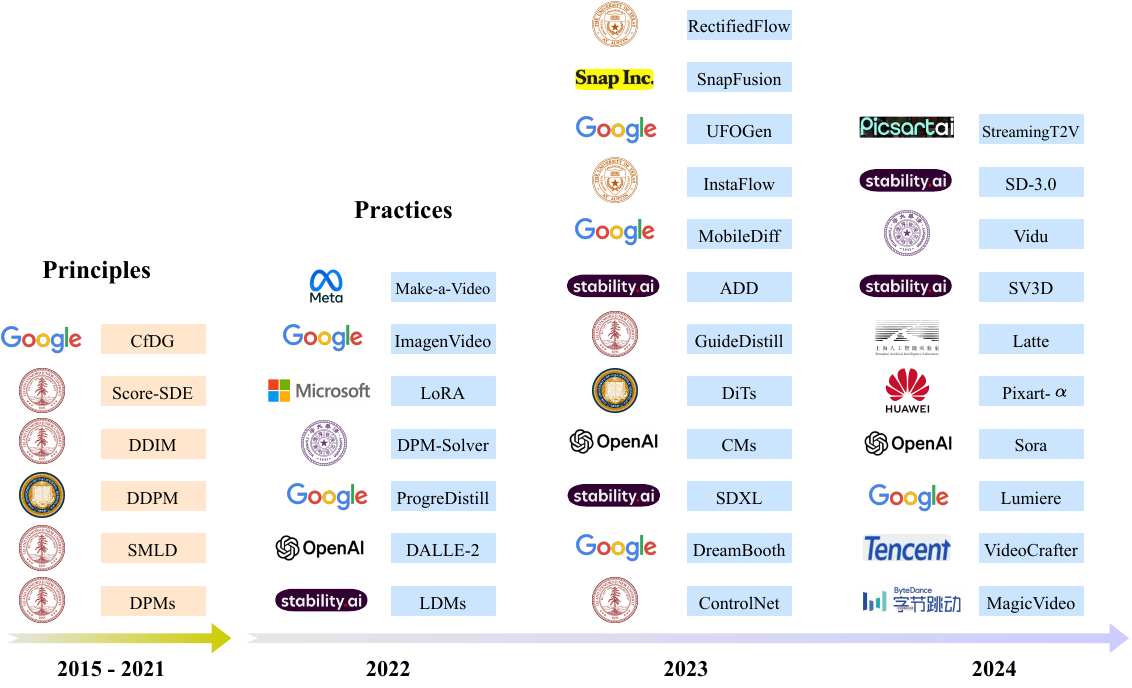}
\caption{The timeline of efficient DMs.}
\label{fig_1}
\end{figure}

Recent years have witnessed the remarkable success of diffusion models (DMs)~\cite{sohl2015deep,ho2020denoising,song2020denoising}, accompanied by a range of visually stunning generative contents emerging. After surpassing GAN on image synthesis~\cite{dhariwal2021diffusion}, DMs have shown a promising algorithm in a wide variety of downstream applications such as image synthesis~\cite{saharia2022photorealistic, ramesh2022hierarchical,ding2022cogview2,podell2023sdxl,zhang2023adding,ruiz2023dreambooth}, video generation~\cite{he2022latent,zhou2022magicvideo,wang2023modelscope,blattmann2023stable,chen2023videocrafter1,chen2024videocrafter2,chai2023stablevideo,wang2024magicvideo,bar2024lumiere}, audio synthesis~\cite{kong2020diffwave,huang2023make,yang2023diffsound}, 3D rendering and generation~\cite{poole2022dreamfusion,shi2023mvdream,lin2023magic3d,zhu2023hifa,voleti2024sv3d} etc., and have emerged as the new state-of-the-art generative models family. Behind these attractive works, the DMs have denser theoretical basis than other generative families such as Variational AutoEncoders (VAEs) and Generative Adversarial Networks (GANs) and a lot of previous efforts have focused on sampling procedure~\cite{zheng2023fast,salimans2021progressive,song2023consistency,luo2023latent}, conditional guidance~\cite{nichol2021glide,rombach2022high,guo2023animatediff,esser2023structure}, likelihood maximization~\cite{kim2022maximum,chen2023score,lu2022maximum,song2021maximum} and generalization ability~\cite{li2023generalized,gu2022vector,mou2024t2i} to improve their efficiency and performance for more powerful generative abilities. 
Standing on the shoulders of these extensive works on the principles and practices of DMs, we have almost seen DMs become a competitive counterpart to LLMs and almost together become the two most brilliant diamonds in the generative AI community today. However, for LLMs, there are already many comprehensive reviews that explain their efforts in efficient architecture design, model training, supervise fine-tuning, preference aligning as well as corresponding applications, but in the field of DMs, existing surveys~\cite{luo2022understanding,croitoru2023diffusion,yang2022diffusion,cao2022survey} still have a significant limitation in comprehensive and in-depth summarize these previous principles and practices (refer to Figure.~\ref{fig_1}), for helping rapid understanding and application in future works.

Besides, a noteworthy trend is that, driven by the advantages of self-attention and deep scalable architecture, LLMs have acquired powerful language emergence capabilities. However, current DMs still face a scalability dilemma~\cite{ma2024neural}, which will play a critical role in supporting large-scale deep generative training and giving rise to emergent abilities~\cite{wei2022emergent} similar to LLMs~\cite{achiam2023gpt}.
Representatively, the recent emergence of Sora~\cite{videoworldsimulators2024} has pushed the intelligent emergence capabilities of generative models to a climax by treating video models as world simulators. While unfortunately, Sora is still a closed-source system and the mechanism for the intelligence emergence is still not very clear. 

\tikzstyle{my-box}=[
    rectangle,
    draw=hidden-draw,
    rounded corners,
    text opacity=1,
    minimum height=1.5em,
    minimum width=5em,
    inner sep=2pt,
    align=center,
    fill opacity=.5,
    line width=0.8pt,
]
\tikzstyle{leaf}=[my-box, minimum height=1.5em,
    fill=hidden-pink!80, text=black, align=left,font=\normalsize,
    inner xsep=2pt,
    inner ysep=4pt,
    line width=0.8pt,
]

\begin{figure*}[ht!]
    \centering
    \resizebox{\textwidth}{!}{
        \begin{forest}
            forked edges,
            for tree={
                grow=east,
                reversed=true,
                anchor=base west,
                parent anchor=east,
                child anchor=west,
                base=center,
                font=\large,
                rectangle,
                draw=hidden-draw,
                rounded corners,
                align=left,
                text centered,
                minimum width=4em,
                edge+={darkgray, line width=1pt},
                s sep=3pt,
                inner xsep=2pt,
                inner ysep=3pt,
                line width=0.8pt,
                ver/.style={rotate=90, child anchor=north, parent anchor=south, anchor=center},
            },
            where level=1{text width=12em,font=\normalsize,}{},
            where level=2{text width=14em,font=\normalsize,}{},
            where level=3{text width=16em,font=\normalsize,}{},
            where level=4{text width=18em,font=\normalsize,}{},
            where level=5{text width=18em,font=\normalsize,}{},
            [
                \textbf{Efficient DMs}, ver
                [
                        \textbf{Principles}(\S\ref{sec:2}), fill=blue!10  
                        [
                            \textbf{Foundational Diffusion}
                            \\\textbf{Theories and Models}(\S\ref{subsec:2.1}), fill=blue!10  
                            [
                                Reverse-SDE~\cite{anderson1982reverse}{,}
                                DPMs~\citep{sohl2015deep}{,} 
                                VDMs~\citep{kingma2021variational}{,}
                                \\DDPM~\citep{ho2020denoising}{,}
                                iDDPM~\citep{nichol2021improved}{,} 
                                DDIM~\citep{song2020denoising}{,}
                                \\DDRM~\cite{kawar2022denoising}{,}
                                PNDM~\cite{liu2021pseudo}{,}
                                INDM~\cite{kim2022maximum}{,}
                                \\D3PM~\citep{austin2021structured}{,}
                                EDM~\cite{karras2022elucidating}{,}
                                CDM~\cite{bansal2024cold}, leaf, text width=18em
                            ]
                        ]
                        [
                            \textbf{Score-based Matching}(\S\ref{subsec:2.2}), fill=blue!10  
                            [
                                NCSN~\citep{song2019generative}{,}
                                LSGM~\citep{vahdat2021score}{,}
                                Score-SDE~\citep{song2020score}{,}
                                \\SSM~\citep{song2020sliced}{,}
                                ScoreFlow~\citep{song2021maximum}{,}
                                ScoreAppr.~\cite{chen2023score}, leaf, text width=18em
                            ]
                        ]
                        [
                            \textbf{Latent Modeling}(\S\ref{subsec:2.3}), fill=blue!10  
                            [
                                LDM~\cite{rombach2022high}{,}
                                LSGM~\citep{vahdat2021score}{,}
                                LCM~\cite{luo2023latent}, leaf, text width=18em
                            ]
                        ]
                        [
                            \textbf{Conditional Guidance}(\S\ref{subsec:2.4}), fill=blue!10  
                            [
                                GLIDE~\cite{nichol2021glide}{,}
                                CfDG~\cite{ho2021classifier}{,}
                                SDG~\cite{liu2021more}{,}
                                \\ADM~\cite{dhariwal2021diffusion}{,}
                                LDM~\cite{rombach2022high}{,}
                                DALL-E2~\cite{ramesh2022hierarchical}, leaf, text width=18em
                            ]
                        ]
                ]
                [
                        \textbf{Mainstream Network}
                        \\\textbf{Architecture}(\S\ref{sec:3}), fill=yellow!10   
                        [
                            \textbf{VAE}(\S\ref{subsec:3.1}), fill=yellow!10
                        [   VQVAE~\cite{vqvae}
                            VQGAN~\cite{VQGAN}{,}
                            C-ViViT~\cite{C-ViViT}{,}\\
                            TATS~\cite{tats}{,}
                            MAGViT~\cite{magvit}{,}
                            CV-VAE~\cite{cv-vae}{,}\\
                            MAGViT-V2~\cite{magvitv2}, leaf, text width=18em
                        ]
                        ]
                        [
                            \textbf{Backbone}(\S\ref{subsec:3.2}), fill=yellow!10
                            [
                                LDM~\cite{rombach2022high}{,}
                                SDXL~\cite{podell2023sdxl}{,}
                                U-ViT~\cite{bao2023all}{,}
                                \\ DiT~\citep{peebles2023scalable}{,}
                                FiT~\cite{lu2024fit}{,}
                                SiT~\cite{ma2024sit}{,}
                                DiM~\cite{teng2024dim}{,}
                                \\ZigMa~\cite{hu2024zigma}{,}
                                Dimba\cite{fei2024dimba}{,}
                                Latte~\cite{ma2024latte}{,}
                                \\SD3.0~\cite{esser2024scaling}{,}
                                Pixart-$\alpha$\cite{chen2023pixart}{,}
                                CogvideoX~\cite{yang2024cogvideox}{,} 
                                \\ Sora~\cite{videoworldsimulators2024}{,} Moive Gen~\cite{moive-gen}, leaf, text width=18em
                            ]
                        ]
                        [
                         \textbf{Text Encoder}(\S\ref{subsec:3.3}), fill=yellow!10
                          [
                            CLIP~\cite{radford2021clip}{,}
                            T5~\cite{raffel2020exploring}{,}
                            mCLIP~\cite{mclip}{,} mT5~\cite{mt5}{,}
                            \\ Lllama~\cite{llama,llama2}{,}
                            ChatGLM3~\cite{chatglm}
                            , leaf, text width=18em
                         ]
                        ]
                    ]                  
                [
                        \textbf{Efficient Training}
                        \\\textbf{and Fine-tuning}(\S\ref{sec:4}), fill=green!10
                        [
                            \textbf{ControlNet Training}
                            \\\textbf{/Fine-tuning}(\S\ref{sec:ControlNet}), fill=green!10
                              [
                               ControlNet~\cite{zhang2023adding}{,}
                               Controlnet-XS~\cite{zavadski2023controlnet}{,}
                               \\ControlnetXt\cite{peng2024controlnext}{,}
                               Controlnet++\cite{li2024controlnet++},
                               leaf, text width=18em
                              ]
                        ]
                        [
                            \textbf{Adapter Training}
                            \\\textbf{/Fine-tuning}(\S\ref{sec:adapter}), fill=green!10
                            [
                            T2I-Adapter~\cite{mou2024t2i}{,}
                            IP-Adapter~\cite{ye2023ip}{,}
                            \\X-Adapter~\cite{ran2024x}{,}
                            Sur-Adapter~\cite{zhong2023adapter}{,}
                            \\SimDA~\cite{xing2024simda}{,}
                            CTRL-Adapter~\cite{lin2024ctrl}, leaf, text width=18em
                            ]
                        ]
                        [
                            \textbf{Low Rank Adaption}
                            \\\textbf{Training/Fine-tuning}(\S\ref{sec:LoRA}), fill=green!10
                            [
                            LoRA~\cite{hu2021lora}{,}
                            LoRA-Composer~\cite{yang2024lora}{,}
                            \\LCM-LoRA~\cite{luo2023lcm}{,}
                            Concept-Sliders~\cite{gandikota2023concept}, leaf, text width=18em
                            ]
                        ]
                        [
                            \textbf{Preference Optimization}(\S\ref{sec:PreferenceOptimization}), fill=green!10
                            [
                            DDPO~\cite{black2023training}{,}
                            HPS~\cite{wu2023human}{,}
                            DreamTuner~\cite{hua2023dreamtuner}{,}
                            \\ImagenReward~\cite{xu2024imagereward}{,}
                            Diffusion-DPO~\cite{wallace2024diffusion}{,}
                            \\RAFT~\cite{dong2023raft}{,}
                            AHF~\cite{lee2023aligning},leaf, text width=18em
                            ]
                        ]
                        [
                            \textbf{Personalized Training}(\S\ref{sec:PersonalizedTraining}), fill=green!10
                            [
                            Textual Inversion~\cite{gal2022image}{,}
                            DreamBooth~\cite{ruiz2023dreambooth}{,}
                            \\BLIP-Diffusion~\cite{li2024blip}{,}
                            ELITE~\cite{wei2023elite}{,}
                            \\Mix-of-show~\cite{gu2024mix}{,}
                            MoA~\cite{ostashev2024moa}{,}
                            OMG~\cite{kong2024omg},leaf, text width=18em
                            ]
                        ]
                    ]
                    [
                            \textbf{Efficient Sampling}
                            \\\textbf{and Inference}(\S\ref{sec:5}), fill=yellow!10
                            [
                                \textbf{Training-Free Methods}(\S\ref{subsec:training-free}), fill=yellow!10
                                [
                                SDE Solver~\cite{songscore, dockhornscore, jolicoeuradversarial, jolicoeur2021gotta, chung2022come, song2019generative, karras2022elucidating}{,} \\
                                ODE Solver~\cite{songdenoising, liupseudo, lu2022dpm, zhanggddim, zhangfast, zhao2023unipc, xue2024accelerating}{,} \\
                                Trajectory Optimization~\cite{watson2021learning, watson2022learning, zhang2023redi}, leaf, text width=18em
                                ]
                            ]
                            [
                                \textbf{Training-based Methods}(\S\ref{subsec:training-based}), fill=yellow!10
                                [
                                Distribution Based Distillation \\ \cite{luhman2021knowledge, salimansprogressive, meng2023distillation, song2023consistency, luo2023latent, yin2024one}{,} \\
                                Trajectory Based Distillation \\ \cite{berthelot2023tract, liuflow, liu2023instaflow, zheng2023fast, fan2023optimizing, yan2024perflow}{,} \\
                                Adversarial Based Distillation~\cite{sauer2023adversarial, sauer2024fast}{,} \\
                                GAN Objective~\cite{xiaotackling, xu2023semi, xu2024ufogen}{,} \\
                                Truncated Diffusion~\cite{lyu2022accelerating, zheng2022truncated}, leaf, text width=18em
                                ]
                            ]
                    ]
                    [
                        \textbf{Efficient Deployment}
                        \\\textbf{and Usage}(\S\ref{sec:6}), fill=green!10
                        [
                            \textbf{Deployment as a Tool}(\S\ref{subsec:dep-tool}), fill=green!10
                            [
                            ComfyUI{,}
                            Automatic1111's SD WebUI, leaf, text width=18em
                            ]
                        ]
                        [
                            \textbf{Deployment as a Service}(\S\ref{subsec:dep-service}), fill=green!10
                            [
                            SnapFusion~\cite{li2024snapfusion}{,}
                            MobileDiffusion~\cite{zhao2024mobiledif}{,}\\
                            DistriFusion~\cite{li2024distrifusion}{,}
                            PipeFusion~\cite{wang2024pipefusion}{,}\\
                            AsyncDiff~\cite{chen2024asyncdiff}, leaf, text width=18em
                            ]
                        ]
                    ]
            ]
        \end{forest}}
    \vspace{-0mm}
    \caption{Organization of efficient diffusion models advancements.}
    \label{fig:efficient MLLMs structure}
    \vspace{0mm}
\end{figure*}
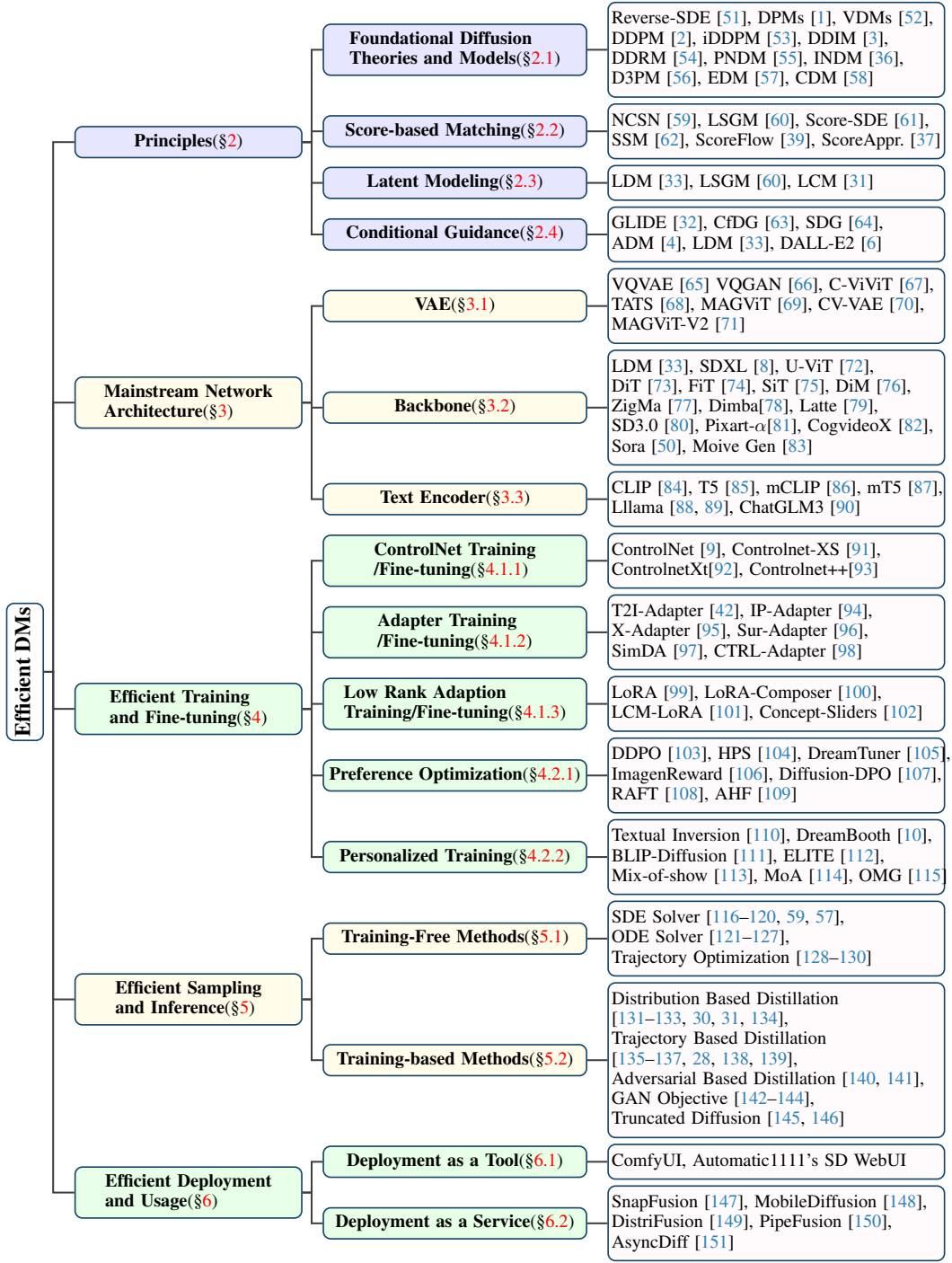

In this survey, we aim to present an exhaustive organization of the recent advancements in the rapidly evolving field of efficient DMs to promote the intelligence emergence of generative models, as depicted in Figure.\ref{fig:efficient MLLMs structure}. We organize the literature in a taxonomy consisting of six primary categories, encompassing various aspects of efficient DMs, including \textbf{principles}, \textbf{efficient architecture}, \textbf{efficient training and fine-tuning}, \textbf{efficient sampling and inference}, \textbf{deployment}, and \textbf{applications}.

\begin{itemize}
    \item[\textbf{•}] \textbf{Principle} focuses on the dense theoretical foundation of DMs to explain and reveal the essential reasons for its generative effectiveness by sorting out relevant theories, such as dynamic modeling, score matching, latent projecting, and conditional guidance, to promote the development of new theories and guide various efficient generative practices.
    \item[\textbf{•}] \textbf{Efficient Architecture} explores the mainstream backbone networks of the DMs, including: U-Net, DiT, U-ViT, MamBa, etc., and analyzes their design structures to compare their respective advantages and disadvantages, in order to guide the emergence of more powerful new deep scalable architectures.
    
    \item[\textbf{•}] \textbf{Efficient Training and Fine-tuning} sorts out the efficient training, finetuning and preference optimization Methods of DMs such as Low Rank Adaption, Consistency Training, Adversarial Training, Adapter Training, etc., and aims to help researchers and developers make appropriate choices for specific low-resource or personalized training tasks.
    \item[\textbf{•}] \textbf{Efficient Sampling and Inference} surveys the most commonly used efficient sampling and inference strategies in diffusion models, covering two categories: learning-free and learning-based methods. By comparing their acceleration performance on various generative tasks, we will provide a theoretical basis for the study of faster sampling methods.
    \item[\textbf{•}] \textbf{Efficient Deployment} summarizes the latest solutions for deploying the current DMs on mobile devices and on the web, which will facilitate the operation of the DMs in various cross-platform, low-resource environments and promote the birth of various applications.
    \item[\textbf{•}] \textbf{Application} investigates the practical applications of efficient DMs in various domains, emphasizing the balance between generative performance, efficiency and computational cost.
\end{itemize}
To sum up, this survey delves into these research endeavors, exploring various theories, methods and strategies for making DMs more design-, training- and computation-efficient. We review the development history of efficient DMs, provide a taxonomy of the strategies for efficient DMs, and comprehensively compare the performance of existing efficient DMs. Through this investigation, we aspire to provide a comprehensive understanding of the current state-of-the-art and efficient generative models. Furthermore, this survey serves as a roadmap, highlighting potential avenues for future research and applications, and fostering a deeper comprehension of the challenges and opportunities that lie ahead in the domain of efficient DMs.
In addition to the survey, we have established a GitHub repository where we compile the papers featured in the survey, organizing them with the same taxonomy at \url{https://github.com/ponyzym/Efficient-DMs-Survey}. We will actively update it and incorporate new research in the future.

\section{Efficient Diffusion Models: Foundational Principles}
\label{sec:2}

The diffusion models~\cite{sohl2015deep,ho2020denoising,nichol2021improved,song2020score} are modeled as a family of unsupervised latent variable models inspired by considerations from nonequilibrium thermodynamics~\cite{sohl2015deep}, which are straightforward to define and efficient to train for generating high-quality samples. We will organize the theoretical contexts of the diffusion models and summarize the core principles below.

\subsection{Definition and Theory Preliminaries}
\label{subsec:2.1}
\paragraph{Discrete Definition}
Assuming the data distribution is $q{(\mathbf{x}_0)}$, the discrete DMs~\citep{sohl2015deep,ho2020denoising} are defined as a forward data perturbation process $q(\mathbf{x}_{1:T}|\mathbf{x}_0)$ and a learnable reverse denoising process $p_{\theta}(\mathbf{x}_{0:T})$, both of them are implemented based on Markov steps for progressive add-noising or denoising,
\begin{equation}
\label{eq: Eq.1}
   q(\mathbf{x}_{1:T}|\mathbf{x}_0)\coloneqq \prod_{t=1}^{T}q(\mathbf{x}_t|\mathbf{x}_{t-1}),\quad p_{\theta}(\mathbf{x}_{0:T})\coloneqq p(\mathbf{x}_T)\prod_{t=1}^{T}p_\theta(\mathbf{x}_{t-1}|\mathbf{x}_{t}).
\end{equation} 
Note that the two symmetric processes are carried out in different fashions. The former leverages an artificially noise-adding scheduler to gradually convert $\mathbf{x}_{0}$ into $\mathbf{x}_{T}$, while the latter usually starts from $p(\mathbf{x}_{T})=\mathcal{N}(\mathbf{x}_{T};\mathbf{0},\mathbf{I})$ and adopts a score matching model $s_\theta$~(Sec.~\ref{subsec:2.2}) to gradually estimate the posterior distribution $p_\theta(\mathbf{x}_{t-1}|\mathbf{x}_{t})$ until $\mathbf{x}_{0}$ is predicted. Specifically, they can be described as:
\begin{equation}
\label{eq: Eq.2}
  q(\mathbf{x}_t|\mathbf{x}_{t-1})\coloneqq \mathcal{N}(\mathbf{x}_{t};\sqrt{\alpha_t}\mathbf{x}_{t-1},\beta_t\mathbf{I}) ,\quad p_\theta(\mathbf{x}_{t-1}|\mathbf{x}_{t})\coloneqq\mathcal{N}(\mathbf{x}_{t-1};\mu_\theta(\mathbf{x}_{t},t),\sigma_\theta(\mathbf{x}_{t},t)).
\end{equation} 
Where $\alpha_t=1-\beta_t$ for facilitating computation and expression. The training objective of $p_\theta$ amounts to minimize the negative log-likelihood of the model,
\begin{equation}
\label{eq: Eq.3}
  L\coloneqq -\mathbb{E}_q\big[\log p_\theta(\mathbf{x}_0)\big]\le -\mathbb{E}_q\Big[\log p(\mathbf{x}_T)-\sum_{t\ge 1}\log \frac{p_\theta(\mathbf{x}_{t-1}|\mathbf{x}_{t})}{q(\mathbf{x}_t|\mathbf{x}_{t-1})}\Big]=L_{vb}.
\end{equation} 
The above variational bound $L_{vb}$ can be rewritten into a tractable form,
\begin{equation}
\label{eq: Eq.4}
\mathbb{E}_q\Big[\underbrace{D_{\text{KL}}(q(\mathbf{x}_T|\mathbf{x}_0)||p(\mathbf{x}_T))}_{L_T}+\sum_{t> 1}\underbrace{D_{\text{KL}}(q(\mathbf{x}_{t-1}|\mathbf{x}_t,\mathbf{x}_0)||p_\theta(\mathbf{x}_{t-1}|\mathbf{x}_t))}_{L_{t-1}}\underbrace{-\log p_\theta(\mathbf{x}_0|\mathbf{x}_1)}_{L_0}\Big]
\end{equation} 
\paragraph{Continuous Definition} Score-SDE~\cite{song2020score} is the first to define continuous-time DMs from the perspective of stochastic differential equations (SDE), which can be simplified as: Let $p_\text{data}(x)$ denote the data distribution, the diffusion models start by exerting a perturbation kernel $p_\sigma(\tilde{x}|x)\coloneqq \mathcal{N}(\tilde{x};x,\sigma^2\mathbf{I})$ onto $p_\text{data}(x)$ for forward process. Then they continues to leverage a reverse ODE (also dubbed as the \emph{Probability Flow} (PF) ODE by~\citep{song2020score}) for inverted denoising, which retain the same marginal probability densities as the forward SDE. The forward and reverse diffusion process in continuous-time form can be expressed as: 
\begin{equation}
\label{eq: Eq.5}
d\mathbf{x}_t=\bm\mu(\mathbf{x}_t,t)dt+\sigma(t)d\mathbf{w}_t, \quad\frac{d\mathbf{x}_t}{dt}=\bm\mu(\mathbf{x}_t,t)-\frac{1}{2}\sigma(t)^2\cdot\Big[\nabla_{x} \log p_t(\mathbf{x}_t)\Big]
\end{equation}
where $\bm\mu(\mathbf{x}_t,t)$ and $\bm\sigma(t)$ are the drift and diffusion coefficient-terms respectively, and $\{\mathbf{w}_t\}_{t\in[0,T]}$ denotes the standard Brownian motion. Moreover, $\nabla_x\log p_t(\mathbf{x}_t)$ denotes the gradient of the log-likelihood of $p_t(\mathbf{x}_t)$, which can be estimated by a score matching network $\mathbf{s}_\theta(\mathbf{x}_t,t)$.

\subsection{Score-based Matching Principle}
\label{subsec:2.2}
Score matching is a popular method for estimating unnormalized statistical models, such as energy-based and flow-based, and it is also well suited for estimating the gradients $\nabla_x\log p_t(\mathbf{x}_t)$ of aforementioned diffusion models. Given samples ${\mathbf{x}_1, \mathbf{x}_2, \cdots, \mathbf{x}_N}\subset \mathbb{R}^D$ from a data distribution $p_\text{data}(x)$, our task is to learn an unnormalized density, $\tilde{p}_m(\mathbf{x};\bm\theta)$, where $\theta$ is from the parameter space $\Theta$. The model’s partition function is denoted as $Z_{\bm\theta}$, which is assumed to be existent but intractable. Let $\tilde{p}_m(\mathbf{x};\bm\theta)$ be the normalized density determined by our model, we have:
\begin{equation}
\label{eq: Eq.6}
p_m(\mathbf{x};\bm\theta)=\frac{\tilde{p}_m(\mathbf{x};\bm\theta)}{Z_{\bm\theta}},\quad Z_{\bm\theta}=\int \tilde{p}_m(\mathbf{x};\bm\theta)d\mathbf{x}.
\end{equation} 

\subsection{Latent Modeling Principle}
\label{subsec:2.3}
The latent space projection is proposed by~\cite{rombach2022high} to compress the input images $\textbf{x}_0$ into a perceptual high-dimensional space to obtain $\textbf{z}_0$ by leveraging a pretrained VQ-VAE model~\cite{esser2021taming}. The VQ-VAE is also adopted by almost all current diffusion models, it consists of an encoder $\mathcal{E}$ and a decoder $\mathcal{G}$. The mathematical definition is: Given an input image $x \in \mathbb{R}^{H\times W\times 3}$, the VQ-VAE first compress the image $x$ into a latent variable $\hat{z}$ by encoder $\mathcal{E}$, i.e., $\hat{z}=\mathcal{E}(x)$ and $\hat{z} \in \mathbb{R}^{h\times w\times d}$, where $h$ and $w$ respectively denote scaled height and width (scaled factor $f=H/h=W/w=8$), and $d$ is the dimensionality of the compressed latent variable. After going through the diffusion step described in Eq.~\ref{eq: Eq.1} or Eq.~\ref{eq: Eq.5}, the latent variable $\hat{z}$ is updated and finally reconstructed into $\hat{x}$ by decoder $\mathcal{G}$, 
\begin{equation}
\label{Eq.7}
	\hat{x}=\mathcal{G}_{\pi}(\text{LDM}_{\mathcal{F}_{\theta}(\cdot)}(\mathcal{E}_{\pi}(x))),
\end{equation}
where LDM($\cdot$) represents the latent diffusion models (including Unet-based or Transformer-based Sec. ~\ref{subsec:3.2}), $\theta$ denotes the parameters of LDM, and $\pi$ denotes the parameters of the VQVAE that are frozen to train our diffusion models. 

\subsection{Conditional Guidance Principle}
\label{subsec:2.4}

\paragraph{Condition-guided Vision Generation.} The core of text-conditional diffusion models is to integrate the semantics of text condition $\bm c$ into noise prediction model $\bm{\epsilon}_\theta(\textbf{z}_t,t)$ to generate visual contents conforming to text semantics, i.e., $\bm{\epsilon}_\theta(\textbf{z}_t,t,\bm c)$. The classifier-free guidance technique has recently been widely adopted in text-guided image generation as,
\begin{equation}
    \tilde{\bm{\epsilon}}_\theta(\textbf{z}_t,t,\bm c,\varnothing)=w\cdot \bm{\epsilon}_\theta(\textbf{z}_t,t,\bm c)+(1-w)\cdot \bm{\epsilon}_\theta(\textbf{z}_t,t,\bm{\varnothing})
\end{equation}
where $w=7.5$ is default linear parameter for weighting the unconditional guidance objective and conditional guidance objective in Stable Diffusion, $t$ is time input, $\bm{c}$ is text condition, $\bm\varnothing$ denotes null text embedding initialized by zero vector and $\theta$ is model parameters. Note that all of these parameters will be individually or jointly optimized for controlled image editing in following variants. 

\paragraph{Condition-guided Vision Editing.} Compared with the condition-guided diffusion models, image editing methods usually own more stringent restrictions, which aim to conduct semantic-guided editing while preserving original pixel characteristics. 
For ControlNet~\cite{zhang2023adding}, parameter $\theta$ is split into $\theta_\text{locked}$ and $\theta_\text{copy}$ for prior preservation and semantic-guided editing, in which $\varnothing$ is trained by a zero convolution layer and condition $\bm{c}$ is split into text prompt $\bm{c}_t$ and image's feature map $\bm{c}_f$. This variant can be formalized as $\tilde{\bm{\epsilon}}_{\theta_\text{locked},\theta_\text{copy}}(\textbf{z}_t,t,\bm{c}_t,\bm{c}_f,\varnothing_\text{zero})$ (\textbf{variant 1}). 
Then, in order to achieve more accurate editing, Prompt-to-Prompt~\cite{hertz2022prompt} introduces a fixed time hyper-parameter $\tau$ to determine when to manipulate the cross-attentive parameters $\theta_{\textbf{M}_t}$ into edited $\theta_{\textbf{M}_t^*}$, which can be formulated as 
$\tilde{\bm{\epsilon}}_\theta(\textbf{z}_t,t,\tau,\bm c,\bm c^*)=w\cdot \bm{\epsilon}_{\theta}(\theta_{\textbf{M}_t^*,t<\tau};\textbf{z}_t,t,\tau,\bm c^*)+(1-w)\cdot \bm{\epsilon}_\theta(\theta_{\textbf{M}_t,t\ge \tau};\textbf{z}_t,t,\tau,\bm{c})$, where $w$ can be viewed as a reweight hyper-parameter 
(\textbf{variant 2}). Afterwards, Null-Text-Inversion~\cite{mokady2023null} optimizes the zero embedding $\varnothing$ into time-aware embedding $\varnothing_t$ with pivot supervision from DDIM inversion process, which can be simply denoted as $\tilde{\bm{\epsilon}}_\theta(\textbf{z}_t,t,\bm{c},\varnothing_t)$ (\textbf{variant 3}). Later, to further realize the subject-binding and prior preservation, DreamBooth~\cite{ruiz2023dreambooth} introduces rare token identifiers ``\texttt{[V]}'' associated with visual subjects and exploits an additional class-specific prior preservation item for training as $\tilde{\bm{\epsilon}}_\theta(\textbf{z}_t,t,\bm c,\bm c_{\texttt{[V]}})=w\cdot \bm{\epsilon}_\theta(\textbf{z}_t,t,\bm c_{\texttt{[V]}})+\lambda\cdot w^{'}\cdot \bm{\epsilon}_\theta(\textbf{z}_t,t,\bm{c})$ (\textbf{variant 4}). Moreover, to enable non-grid editing~\cite{kawar2023imagic,ma2024adapedit}, Imagic~\cite{kawar2023imagic} optimizes text embedding $\bm c$ and leverages an interpolation technique to implement variable guidance, which is controlled by a linear hyper-parameter $\eta$ as $\tilde{\bm{\epsilon}}_\theta(\textbf{z}_t,t,\bm c^*)$ (\textbf{variant 5}), where $\bm c^*=\eta\cdot \bm{c}_\text{tgt}+(1-\eta)\cdot\bm{c}_\text{opt}$.


\section{Mainstream Network Architectures}
\label{sec:3}

\begin{figure}[t!]
	\centering
	\includegraphics[width=1.00\linewidth]{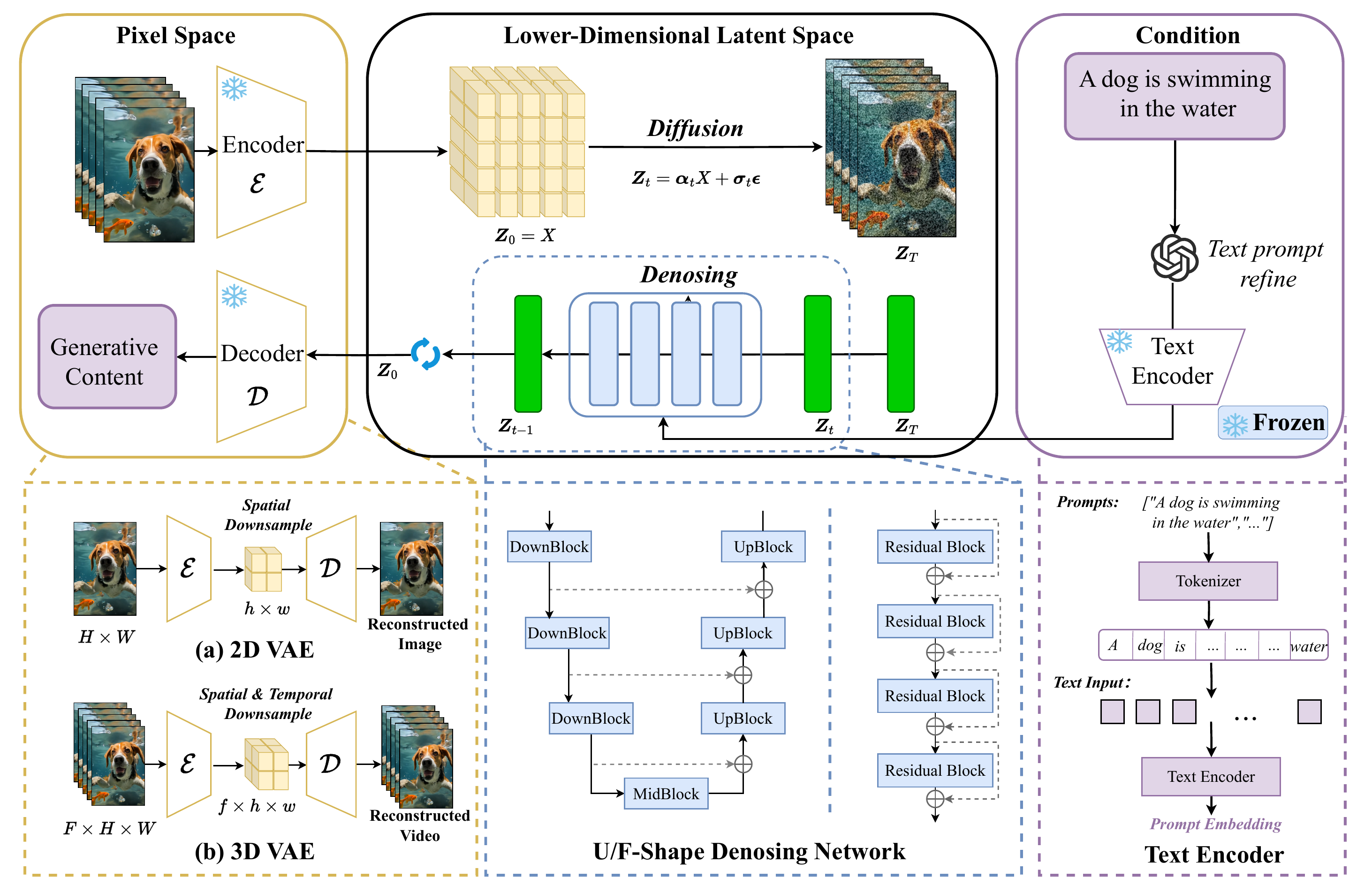}
	\caption{A universal pipeline of the diffusion based models for visual content generation. A pre-trained VAE (with encoder and decoder structures) compresses the input image or video into a latent space. Diffusion models add noise to the latent features and train a neural network  (e.g. U-Net or Transformer) for de-noising. User-input text instructions are refined by a large language model and then encoded by a trained text encoder into an embedding space, which is injected into the diffusion model to control content generation.}
 \label{fig: modules}
\end{figure}

As shown in Figure \ref{fig: modules}, following the Latent diffusion model (LDM)~\cite{rombach2022high}, most recent text-conditional visual generation models consist of three main modules: a variational auto-encoder (VAE) is trained and served as a latent compressor, which encodes images or videos from a high-dimensional pixel space into latent space.  The model performs diffusion and denoising in the compressed latent space. A neural network is optimized for learning the probability distribution required for each denoising step.  A text encoder that encodes the input text into a text embedding as a condition to control and guide for the generation of the image or video content.

\begin{table*}[h]
\renewcommand{\arraystretch}{1.2}
\centering
\setlength{\tabcolsep}{5pt}
\footnotesize
\scalebox{0.7}{
\begin{tabular}{ccccccc}
    \toprule
        \textbf{Methods} & \textbf{Year} & \textbf{Organization} & \textbf{Backbone} & \textbf{VAE} & \textbf{Text Encoder} & \textbf{\# Params} \\ \hline
        ADM~\cite{dhariwal2021diffusion} & 2021 & OpenAI & \multirow{8}{*}{Unet} & \multirow{4}{*}{None} & - & 554M \\ 
        CDM~\cite{ho2022cascaded} & 2021 & Google & ~ & ~ & - & -\\ 
        DALL-E 2~\cite{ramesh2022hierarchical} & 2022 & OpenAI & ~ & ~ & CLIP & 6.5B \\ 
        Imagen~\cite{saharia2022photorealistic} & 2022 & Google & ~  & ~ & T5-XXL & 3B \\ \cdashline{5-7}[1pt/1pt] 
        LDM~\cite{rombach2022high}  & 2022 & LMU Munich & ~ &  \multirow{5}{*}{2D VAE} & CLIP ViT-L & 400M+55M(VAE)\\ 
        SD1.5~\cite{rombach2022high}  & 2022 & LMU Munich & ~ &  ~ & CLIP ViT-L & 860M\\ 
        SD2.0~\cite{rombach2022high}  & 2022 & LMU Munich & ~ & ~ & OpenCLIP ViT-H & 865M\\ 
        SDXL~\cite{podell2023sdxl}  & 2023 & Stability AI & ~  & ~ & CLIP ViT-L \& OpenCLIP ViT-bigG & 2.6B\\ 
        Playground-v2.5~\cite{li2024playground} & 2024 & Playground & ~ & ~ & CLIP & - \\  \cdashline{1-7}[1pt/1pt] 
        UViT~\cite{bao2023all}  & 2022 & Tsinghua University &  \multirow{16}{*}{Transformer} & \multirow{11}{*}{2D VAE} & CLIP ViT-L & 501M+84M(VAE) \\ 
        DiT~\cite{peebles2023scalable}  & 2022 & UC Berkeley &  ~ & ~  & CLIP ViT-L & 675M+84M(VAE)\\ 
        PixArt-$\alpha$~\cite{chen2023pixart} & 2023 & Huawei Noah’s Ark Lab & ~ & ~ & T5-XXL & 600M \\ 
        FiT~\cite{lu2024fit}  & 2024 & Shanghai AI Lab &  ~ & ~  & CLIP ViT-L & - \\ 
        SiT~\cite{ma2024sit}  & 2024 & New York University &  ~ & ~  & CLIP ViT-L & 675M\\ 
        Latte~\cite{ma2024latte}  & 2024 & Shanghai AI Lab  & ~ & ~  & T5-XXL & 673.68M \\ 
        Hunyuan-DiT~\cite{li2024hunyuan}  & 2024 & Tencent Hunyuan & ~  & ~ & mCLIP \& mT5-XL & 1.5B\\ 
        LuminaT2X~\cite{gao2024lumina}  & 2024 & Shanghai AI Lab & ~ & ~ & LLama2-7B & 7B\\  
        Kolors~\cite{kolors}  & 2024 & Kuaishou & ~ & ~ & ChatGLM3-6B-Base & 2.6B\\  
        SD3.0~\cite{esser2024scaling}  & 2024 & Stability AI & ~  & ~ & CLIP ViT-L \& OpenCLIP ViT-bigG \& T5-XXL & 8B \\
        Flux.1~\cite{Flux}  & 2024 & BlackForestLabs & ~  & ~ & CLIP ViT-L \& OpenCLIP ViT-bigG \& T5-XXL & 12B \\
        \cline{5-7}
        Sora~\cite{liu2024sora}  & 2024 & OpenAI & ~ & \multirow{5}{*}{3D VAE} & - & - \\ 
        Open-Sora~\cite{opensora}  & 2024 & Hpcaitech & ~ & ~ & T5-XXL & 1.2B \\ 
        Open-Sora-Plan~\cite{opensora-plan} & 2024 & Peking University & ~ & ~ & T5 \& mT5 & - \\ 
        EasyAnimate~\cite{xu2024easyanimate} & 2024 & Alibaba Group & ~ & ~ & mCLIP \& mT5-XL & 1.5B \\ 
        CogvideoX~\cite{yang2024cogvideox}  & 2024 & Zhipu AI & ~ & ~  & T5-XXL & 2B/5B \\ 
        Moive Gen~\cite{moive-gen} & 2024 & Meta & & TAE & MetaCLIP \& UL2 \& ByT5 & 30B \\
    \bottomrule
\end{tabular} 
}
\caption{Comparison of modules and parameters in different diffusion generative Models.}
\label{tab: module-compare}
\vspace{-10pt}
\end{table*}

\subsection{VAE for Latent Space Compression}
\label{subsec:3.1}
\begin{figure}[ht!]
	\centering
	\includegraphics[width=1.00\linewidth]{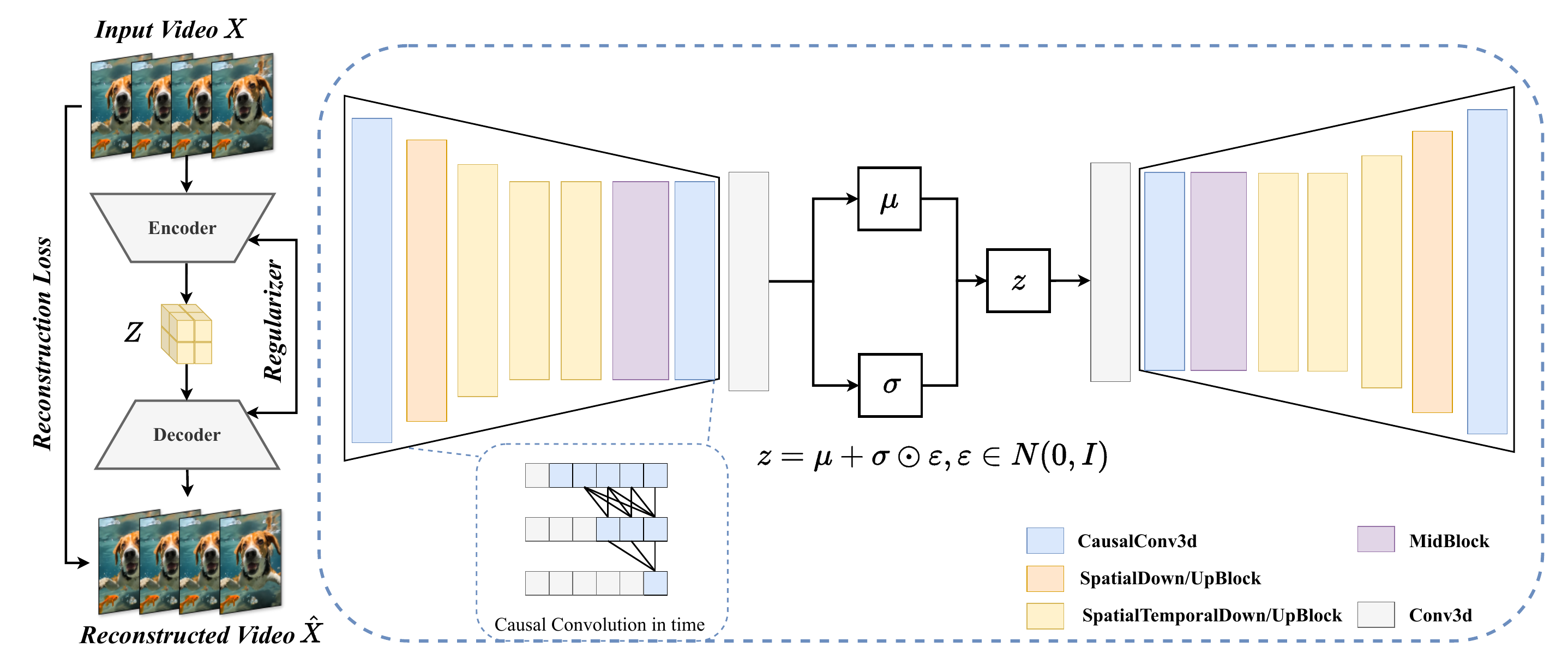}
	\caption{A standard encoder-decoder architecture of 3D Variational Autoencoders (VAEs) are utilized for video compression.}
    \label{fig: VAE}
\end{figure}

\begin{table*}[]
\renewcommand{\arraystretch}{1.2}
\centering
\footnotesize
\scalebox{0.9}{
\begin{tabular}{ccccccccc}
    \toprule
        \multirow{2}{*}{\textbf{Models}} & \multirow{2}{*}{\textbf{Compress Ratio}} & \multicolumn{3}{c}{\textbf{WebVid}} & \multicolumn{3}{c}{\textbf{Panda-70M}}  \\ 
         & & PSNR$\uparrow$ & SSIM$\uparrow$ & LPIPS$\downarrow$ & PSNR$\uparrow$ & SSIM$\uparrow$ & LPIPS$\downarrow$ \\    \hline
        SD2.1 VAE~\cite{rombach2022high} & 1$\times$8$\times$8 & 30.19 & 0.8379 & 0.0568 & 30.40 & 0.8894 & 0.0396  \\ 
        SVD VAE~\cite{blattmann2023stable} & 1$\times$8$\times$8 & 31.15 & 0.8686 & 0.0547 & 31.00 & 0.9058 & 0.0379 \\ 
        CV-VAE~\cite{cv-vae} & 4$\times$8$\times$8 & 30.76 & 0.8566 & 0.0803 & 29.57 & 0.8795 & 0.0673 \\ 
        Open-Sora VAE~\cite{opensora} & 4$\times$8$\times$8 & 31.12 & 0.8569 & 0.1003 & 31.06 & 0.8969 & 	0.0666 \\ 
        Open-Sora-Plan VAE~\cite{opensora-plan} & 4$\times$8$\times$8 & 31.16 & 0.8694 & 0.0586 & 30.49 & 0.8970 & 0.0454 \\  
    \bottomrule
\end{tabular} 
}
\caption{Comparison of VAE performance in common image and video generation diffusion models.}
\label{tab: vae-compare}
\vspace{-10pt}
\end{table*}

Diffusion and denoising in high-dimensional RGB pixel space~\cite{ho2020denoising, dhariwal2021diffusion,  saharia2022photorealistic, ramesh2022hierarchical, ho2022cascaded} results in an expensive training cost and affects the speed of inference. To make the diffusion model accessible while reducing its significant resource consumption, LDM~\cite{rombach2022high} observes that most bits of an image contribute to perceptual details and retain semantic and conceptual composition even after aggressive compression. LDM removes pixel-level redundancy by training a VAE that compresses the input image from the pixel space to the latent space. Then diffusion and denoising are performed in latent space, which significantly reduces the cost of training and reasoning for DMs. Figure~\ref{fig: VAE} illustrates the structure of standard VAEs for image/video compression, which include normal variational auto-encoders (VAEs)~\cite{kingma2013auto}, quantized VAEs such as VQVAE~\cite{vqvae} or VQGAN~\cite{VQGAN} and their variants~\cite{lee2022autoregressive}, where the GAN discriminator loss is added to achieve reconstruction quality for higher compression
More importantly, the trained VAE is a generalized compression model, and its latent space can be used to train multiple generative models and applied to other downstream tasks.
Following the LDM, the posterior image generation approach~\cite{ho2022imagen,li2024playground, podell2023sdxl,bao2023one,peebles2023scalable,ma2024lmd,lu2024fit,ma2024sit,chen2023pixart,li2024hunyuan,kolors,esser2024scaling} compresses/decompresses the image in latent space by training a VAE using the encoder and decoder of the VAE. The parameters of the VAE are frozen during diffusion model training and inference. Some diffusion models~\cite{singer2022make,ho2022imagen,zhang2023show} are used to generate videos by directly learning pixel distributions. Video contains not only spatial information but also a lot of temporal information, so there are more computational challenges in video generation. In addition, the diffusion video generation models exemplified by Sora~\cite{liu2024sora} use VAE to compress the video and then train and reason in latent space. These video generation models~\cite{ma2024latte,opensora,opensora-plan,xu2024easyanimate} are usually derived from Stable Diffusion's image 2D VAEs, since training 3D from scratch is quite challenging. Time compression is simply achieved by uniform frame sampling while ignoring motion information between frames. Table~\ref{tab: vae-compare} compares the performance parameters of VAEs commonly used in the community
Some methods use hybrid 2D-3D VAEs~\cite{magvit,xu2024easyanimate,opensora,opensora-plan, yang2024cogvideox,cv-vae,C-ViViT} or full 3D VAEs~\cite{magvitv2}. e.g., MAGViT~\cite{magvit} uses 3D VQGAN with 3D and 2D downsampling layers, and MAGViT-V2~\cite{magvitv2} uses a full 3D convolutional encoder with overlapping downsampling. In order to trade off lower memory and computational cost with slightly lower reconstruction quality, the latest video generation model, Moive Gen~\cite{moive-gen}, uses interleaved 2D-1D convolutional encoders in its VAE.

\subsection{Denoising Neural Network Backbone}
\begin{figure}[h!]
	\centering
	\includegraphics[width=1.00\linewidth]{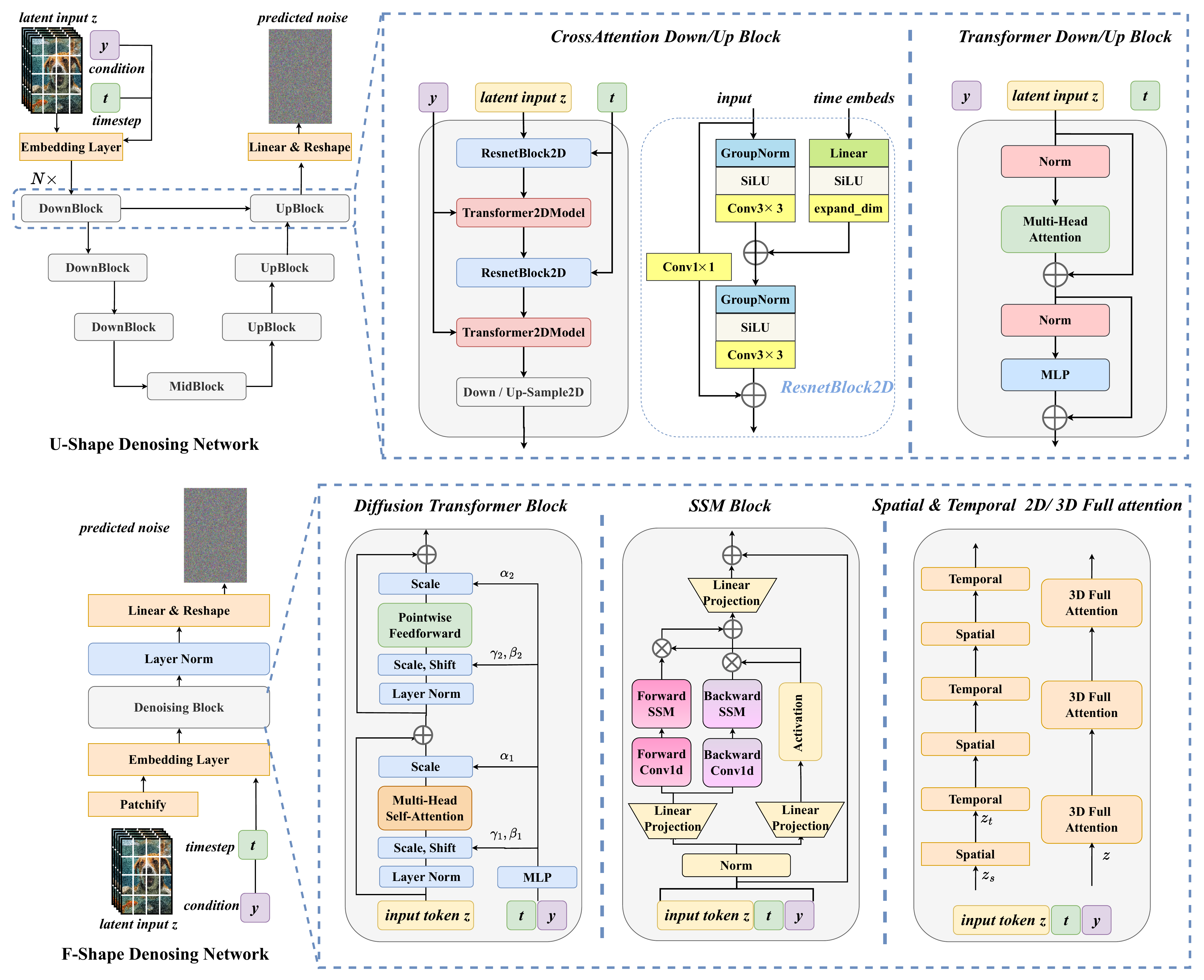}
	\caption{The mainstream neural network backbones serving as denoisers in diffusion models, which including U-shaped denoising networks (U-Net based and U-ViT based) and F-shaped denoising networks (DiT-based and SSM-based).}
 \label{fig: backbone}
\end{figure}

As shown in Figure.~\ref{fig: backbone}, the neural networks within the diffusion models mainly serve as \emph{residual-style} noise predictors in the de-noising stage~\cite{ma2024neural}, which can be categorized into the following mainstream architectures:
\label{subsec:3.2}

\textbf{U-Net based Backbone.}
DDPM~\cite{ho2020denoising} as the seminal work that introduces U-Net~\cite{ronneberger2015u} as the backbone for the diffusion model to predict the probability distribution at each step of the de-noising process. In which the U-Net follows from PixelCNN++~\cite{salimans2017pixelcnn++} and utilizes an encoder-decoder architecture, where the spatial pixels of the image are downsampled by convolutional operations at each layer of the encoding process while extracting the features. The spatial resolution is progressively restored in the decoder stage, while the feature information extracted by the encoder and decoder is fused via skip connections. U-Net~\cite{ronneberger2015u} is capable to process the image features at different scales, which helps in the gradual de-noising process. Specifically, Song et al.~\cite{songscore} improved the performance of unconditional image generation tasks by making further changes for U-net in the Sore-based diffusion model. Prafulla et al.~\cite{dhariwal2021diffusion} improved the U-Net architecture in the diffusion model by increasing the width and depth of the network, and increasing the number of attention heads, etc., achieved better performance than GAN on the image generation tasks. Other models~\cite{ho2022cascaded,nichol2021glide,ramesh2022hierarchical} with U-Net based architectures perform diffusion and denoising directly in high-dimensional RGB pixel space, incurring high training costs and affecting the inference speed. Based on the LDM~\cite{rombach2022high}, SDXL~\cite{podell2023sdxl} uses more attention blocks and a larger cross-attention context, thus including more parameters in the U-Net. VDM~\cite{vdm} extends LDM to the video generation task by introducing 3D convolutional layers.

\textbf{Transformer based Backbone.}
Transformer has shown the dominance in the fields of Natural Language Processing~\cite{devlin2018bert,radford2018improving,ma2022glaf,ma2021intention}, Computer Vision~\cite{dosovitskiy2020vit} and Multi-modality~\cite{ma2022cmal,ma2022unitranser,ma2023hybridprompt} with its scalability and ability to model long-range dependencies of the attention mechanism. This trend is also held in many autoregressive image generation models~\cite{chen2020generative, ramesh2021zero}. However, before U-ViT~\cite{bao2023all} and DiT~\cite{peebles2023scalable} were proposed, advanced diffusion models for image generation tasks still adopt a convolutional U-Net architecture. U-ViT~\cite{bao2023all} introduced the Transformer Block in a U-shaped structure as a backbone for diffusion models, which treats all inputs as tokens and utilizes a long skip connection between the shallow and deep layers.
DiT~\cite{peebles2023scalable} introduced Vision Transformer~\cite{dosovitskiy2020vit} as a backbone to replace U-Net, and further demonstrates the scalability of Transformer for image generation tasks. Recent works have also demonstrated the superior performance of the diffusion generation model of the DiT architecture for image~\cite{chen2023pixart} and video generation tasks~\cite{ma2024latte, liu2024sora}. Specifically, PixArt-$\alpha$~\cite{chen2023pixart} simplicates computationally intensive class conditional branching in Diffusion Transformer by joining the cross-attention module to inject textual conditions that are encoded through T5~\cite{raffel2020exploring}. Latte~\cite{ma2024latte} expands the DiT architecture to the video generation task by extracting spatio-temporal tokens from the input video, and introducing temporal and spatial transformer blocks to model the video distribution in latent spaces, respectively. Futhermore, following DiT, Latte uses AdaLAN for time-step class information injection.  Notably, the emergence of Sora~\cite{liu2024sora} demonstrates the substantial scalability of the Transformer architecture for generating high-quality video content. There are a number of recent image~\cite{li2024hunyuan,esser2024scaling,Flux} and video~\cite{opensora,opensora-plan,yang2024cogvideox,xu2024easyanimate,moive-gen} generation models that have also verified the scalability of transformer in diffusion modeling under large-scale training.

\textbf{SSM based Backbone.}
The transformer-based diffusion models suffer from the quadratic complexity of the attention mechanism, making them consume huge computational cost for long sequence generation tasks (e.g., high-resolution image synthesis, video generation, etc.). Advances in state-space modeling (SSM)~\cite{gu2023modeling,qi2024exploring} show a new direction to achieve a trade-off between computational efficiency and model flexibility. Some recent SSM-based approaches~\cite{gu2020hippo,gu2021efficiently,gupta2022diagonal} have been proposed and proven their efficiency on multiple tasks and modalities in modeling long sequence dependencies. Mamba~\cite{gu2023mamba} combines SSM architectures and proposes hardware-aware algorithms that enable efficient training and inference. DiM~\cite{teng2024dim} introduces Mamba as a diffusion backbone for high-resolution image generation. Specifically, DiM avoids unidirectional causality between patches by designing the Mamba block to perform the four scanning directions alternately. In consideration of the lack of spatial continuity in mamba scanning schemes, ZigMa~\cite{hu2024zigma} allows mamba blocks applicable to 2D images by incorporating continuity-based inductive bias in the images. In addition by performing spatio-temporal decomposition of 3D sequences, which is extended to video generation task.  

In addition to the above types of mainstream diffusion model architectures, there are some other diffusion model architectures for image and video generation. Diffusion-RWKV~\cite{fei2024diffusion} introduces the RWKV~\cite{peng2023rwkv} architecture as the Backbone for Diffusion models. The RWKV consists of an input layer, a series of stacked residual blocks and an output layer. Each residual block consists of temporal mixing and channel mixing sub-blocks. RWKV improves on the standard RNN architecture by parallelizing computation during training similarly to RNN. It includes enhancements to the linear attention mechanism and designs the receptance weight key value (RWKV) mechanism. DiG~\cite{zhu2024dig} introduces a Gated Linear Attention Transformer (GLA)~\cite{yang2023gated} and proposes Diffusion GLA model. DiG achieves high efficiency in terms of training speed and GPU memory for high resolution image generation.

\begin{figure}[ht!]
	\centering
	\includegraphics[width=1.00\linewidth]{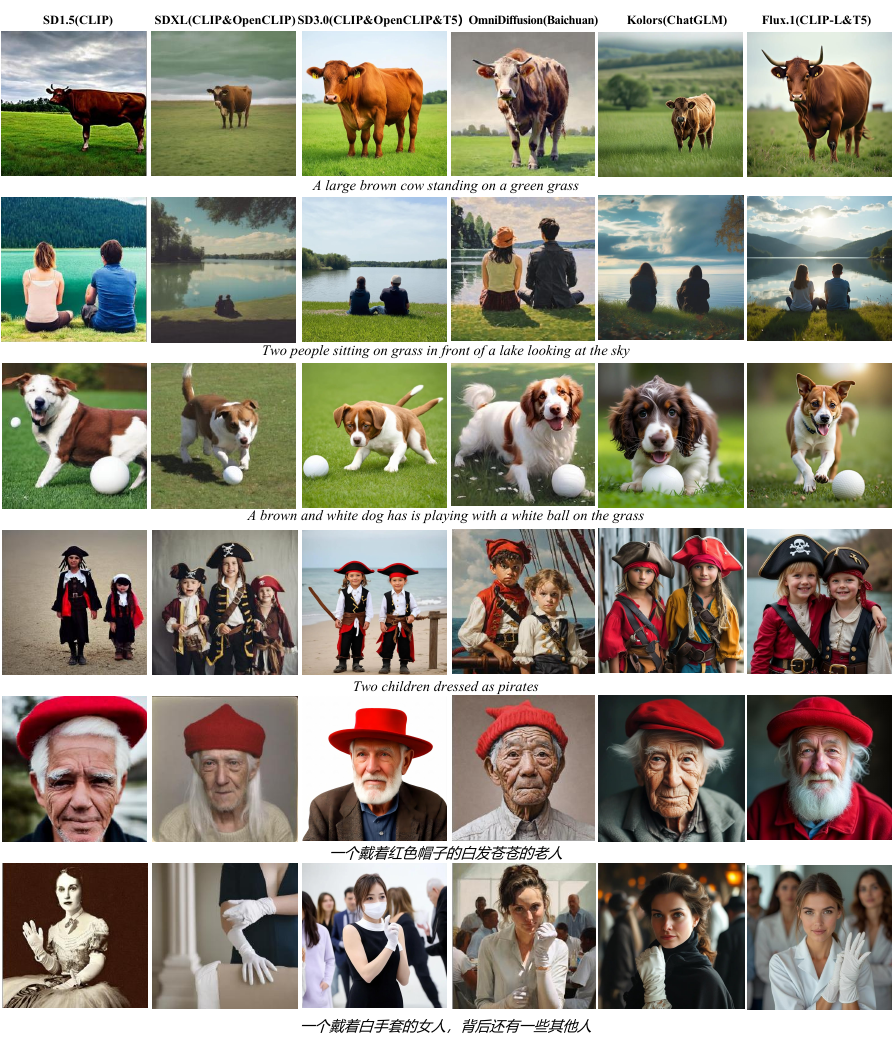}
	\caption{Comparison of  generated images from diffusion models with different text encoders. The last two rows are Chinese prompts, which are used to test image generation models with the text encoders that support the multilingual condition (i.e., OmniDiffusion~\cite{omniDiffusion}, Kolors~\cite{kolors}). For models that do not support the multilingual text condition, the given prompts are translated into the corresponding language to generate images.}
 \label{fig: text-encoder}
\end{figure}

\subsection{Text encoder}
\label{subsec:3.3}
The text encoder is used to capture the complex semantics within the input text prompts, which is a critical component of the text-conditional visual generation model and directly affects the generated content. Early text-to-image approaches used text encoders which are trained on paired text-images, and they can be trained from scratch~\cite{nichol2021glide,ramesh2021zero} or pre-trained (e.g., CLIP~\cite{radford2021learning}). CLIP uses contrast learning, and is trained to align the embedding representations of the text and images. These text encoders can encode visual and textual semantics after being trained using paired text-images. After tokenizer and embedding, the input text prompt is injected as a condition into the diffusion model generative backbone.  As shown in the table~\ref{tab: module-compare}, some classical text to image diffusion models~\cite{ramesh2022hierarchical,rombach2022high,bao2023all,peebles2023scalable,lu2024fit,ma2024sit,li2024hunyuan,esser2024scaling, Flux} use the text branch of CLIP models for text representation. Typically, the parameters of these text encoders are frozen thus their computational and memory consumption during diffusion model training can be ignored. 
CLIP series models focus on the global representation of an image by aligning the embedding space of the image and the text, however, it is difficult to understand the detailed description. Large language models are trained on larger text corpora and have stronger text comprehension and generation capabilities. Imagen~\cite{ho2022imagen} compared CLIP with a pre-trained large language model (BERT~\cite{kenton2019bert}, T5~\cite{raffel2020exploring}) as text encoders. In addition, they found that scaling the size of the text encoder can improve the quality of text-to-image generation and that using the T5-XXL encoder achieves better image-text alignment and image fidelity. Some approaches merge both CLIP and T5 encoders to improve the ability of text comprehension. Some image diffusion models~\cite{AltDiffusion, PAI-Diffusion, Taiyi-Diffusion, li2024hunyuan} focus on understanding the multilingual prompt and generating images. HunyuanDiT~\cite{li2024hunyuan} combines a bilingual CLIP~\cite{mclip} and a multilingual T5~\cite{mt5} text encoder to improve Chinese comprehension. Some recent image~\cite{omniDiffusion,kolors} and video~\cite{yang2024cogvideox} generation models use large language models (e.g., Baichuan~\cite{baichuan}, Llama~\cite{llama,llama2}, and ChatGLM~\cite{chatglm}) to enhance semantic understanding of complex text. Figure~\ref{fig: text-encoder} provides a visual comparison that demonstrates how the understanding of complex texts by large language models affects the generation effects of diffusion models.

\section{Efficient Training and Fine-tuning}
\label{sec:4}
The efficient training strategies of diffusion models aim to reduce training time and resource consumption while maintaining performance improvements, making diffusion models more flexible in a wide range of downstream tasks. Here, we mainly lay emphasis on two aspects of efficient training: parameter efficiency and label efficiency.
Parameter-efficient methods focus on optimizing the architecture of trainable modules to reduce the number of parameters required for high performance. Meanwhile, label-efficient methods aim to minimize the amount of training data needed, which is especially critical when high-quality labeled datasets are limited or unavailable. In this section, we provide a brief overview of various techniques and approaches that enhance parameter efficiency and label efficiency, and discuss their significance in downstream tasks of diffusion models.

\subsection{Parameter-Efficient Methods}
Parameter-efficient training methods aim to adapt pre-trained models to new tasks by updating only a small number of parameters, rather than the entire model, thereby prevent overfitting while improving performance. Following the definition in ~\cite{ding2022delta}, given the pretrained parameters of a diffusion model $\theta=\{w_{1},w_{2},\dots,w_{n}\}$, the fine-tuning task aims to obtain the parameters $\theta'=\{w_{1},w_{2},\dots,w_{m}\}$ on a given dataset $D$. The parameter update is defined as $\Delta \theta = \theta - \theta'$. Compared to full fine-tuning, where $|\Delta \theta| = |\theta|$, efficient training is achieved when $|\Delta \theta| \ll |\theta|$, where $|\cdot|$ denotes the number of parameters.

\begin{figure}[h]
\centering
\includegraphics[width=1.00\linewidth]{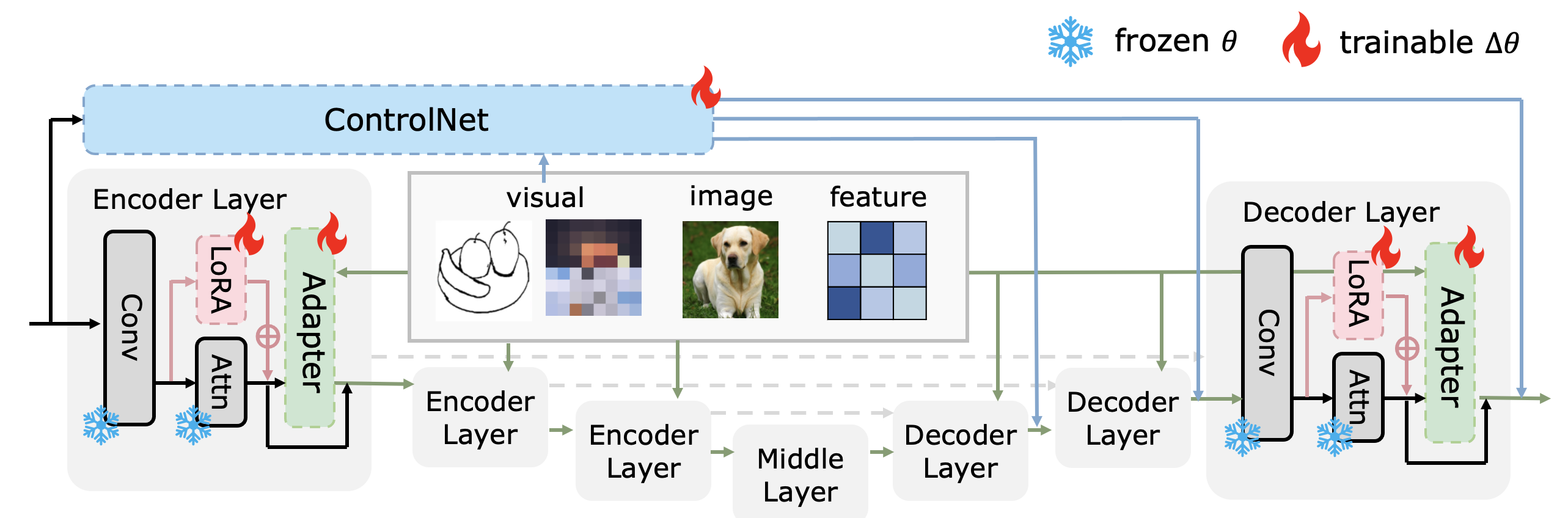}
\caption{A generic training framework for parameter-efficient training approaches in diffusion models. The model leverages frozen base parameters while introducing trainable components through ControlNet, LoRA, and adapter modules  with visual, image, and feature inputs progressively encoded to produce the final output.}
\label{fig:parameter-efficent_framework}
\end{figure}

As shown in  Figure \ref{fig:parameter-efficent_framework}, parameter-efficient training techniques can be categorized into three types: ControlNet~\cite{zhang2023adding}, low-rank adaption (LoRA)~\cite{hu2021lora}, and adapter~\cite{ xing2024simda, guo2024i2v}. These approaches add and update lightweight modules, enabling efficient adaptation to new tasks. In the following subsections, we will analyze the application advantages of these techniques across various downstream tasks.

\subsubsection{ControlNets}
\label{sec:ControlNet}
\begin{figure}[h]
\centering
\includegraphics[width=1.00\linewidth]{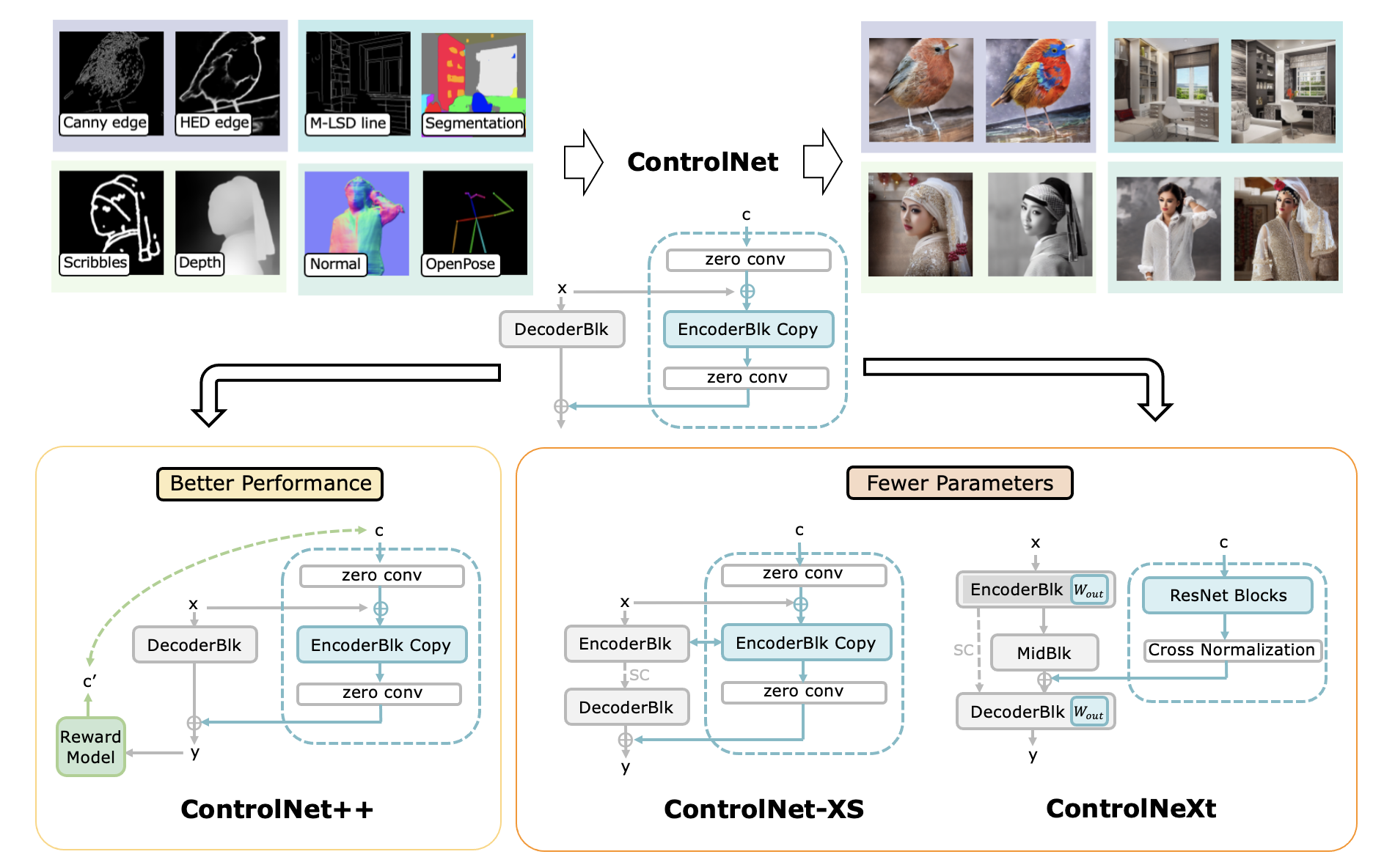}
\caption{An illustration of ControlNet and its extensions, demonstrating its ability to guide image generation using various control signals such as edges, depth, segmentation, and poses.}
\label{fig:ControlNet}
\end{figure}

Despite the impressive text-to-image capabilities~\cite{nichol2021glide, balaji2022ediff, ding2022cogview2, ramesh2022hierarchical, rombach2022high, saharia2022photorealistic}, diffusion models often struggle with spatial compositional control~\cite{isola2017image}, particularly in tasks such as depth-to-image and pose-to-image. To address these limitations, ControlNet~\cite{zhang2023adding} introduces visual features into the multi-resolution layers of a pre-trained UNet, thereby enabling more controllable generation. This advancement has spurred further research, resulting in several efficient variants of ControlNet~\cite{zavadski2023controlnet, peng2024controlnext, li2024controlnet++}. As illustrated in Figure \ref{fig:ControlNet}, these improvements focus on two main aspects: reducing the number of parameters in ControlNet while maintaining or improving its performance, and enhancing finer-grained control without increasing the number of parameters.

A branch of works prefer to reduce the parameter count of ControlNet. ControlNet-XS~\cite{zavadski2023controlnet} found that with high-frequency and large-bandwidth communication between the control blocks and generative network, the control module requires fewer parameters to achieve better results, speeding up both inference and training. ControlNeXt~\cite{peng2024controlnext} introduces a lightweight convolutional module to extract control features, replaces zero-convolution with cross normalization to align the parameter distributions with those of main denoising branch and achieve faster and more stable training convergence.

Another line of works enhance ControlNet's controllability over the generated output while maintaining the same parameter count, ControlNet++~\cite{li2024controlnet++} employs a pre-trained discriminative reward model to effectively bridge the gap between conditions and generated images, improving the quality and pixel-level relevance of the output when control signals are reflected in the generated images.

\subsubsection{Adapters}
\label{sec:adapter}
\begin{figure}[h]
\centering
\includegraphics[width=1.00\linewidth]{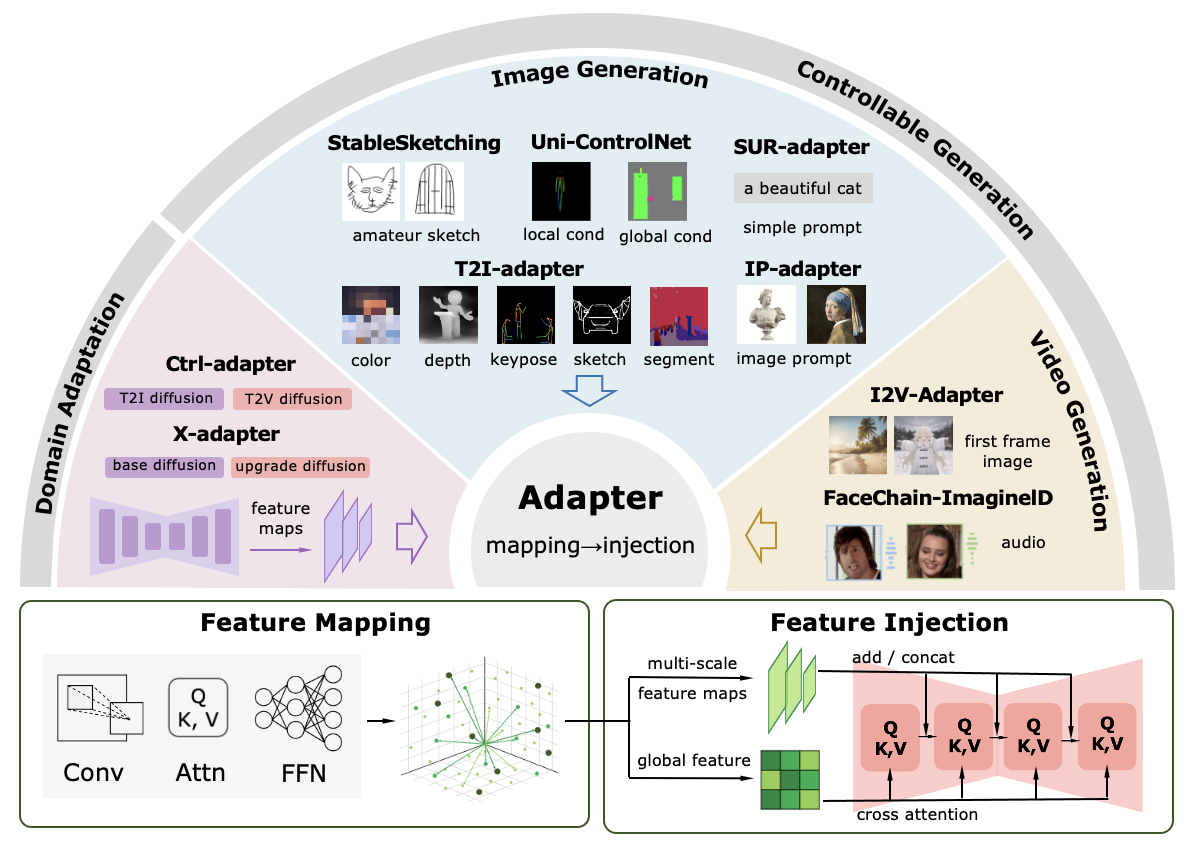}
\caption{An overview of the Adapter framework for diffusion models, illustrating various adapters, which facilitate feature mapping and injection processes, allowing the diffusion model to handle diverse input types.}
\label{fig:Adapter}
\end{figure}

Compared to ControlNet~\cite{zhang2023adding}, which achieves additional spatial control by fine-tuning duplicated encoders, adapter-based methods~\cite{ye2023ip, mou2024t2i, xing2024simda, lin2024ctrl, guo2024i2v} boast more flexible and lightweight architectures as shown in Figure \ref{fig:Adapter} and reduce the need for extensive data and computational resources as detailed in Table \ref{tab:adapter}. 
As crucial parts that allow models to perform a variety of downstream tasks, adapters establish the intrinsic connection between the conditional inputs and their corresponding images, making them commonly used for controllable generation~\cite{zhong2023adapter, zhao2024uni, koley2024s} and domain adaptation~\cite{rebuffi2018efficient}.

\begin{table}[H]
\centering
\scalebox{0.7}{
\begin{tabular}{cccccccc}
\toprule
\textbf{Model} & \multicolumn{2}{c}{\textbf{Dataset}}& \textbf{Condition}& \textbf{Params} & \textbf{Hardware} & \textbf{Time} \\

\midrule
\multicolumn{7}{c}{\textbf{\textit{Controllable Generation}}} \\  
\midrule

\multirow{3}{*}{T2I-Adapter~\cite{mou2024t2i}} & COCO17 & 164K images & sketch map  & \multirow{3}{*}{77M} & \multirow{3}{*}{4 32G V100} & \multirow{3}{*}{3d} \\

 & COCO-Stuff & 164K images & segmentation map & & & \\
 
 & LAION-Aesthetics  & 600K T-I pairs & keypoints/color/depth & & & \\
 
StableSketching~\cite{koley2024s} & Sketchy database & 12.5K images & abstract sketch & - & - & - \\

\multirow{2}{*}{Uni-ControlNet~\cite{zhao2024uni}} & \multirow{2}{*}{LAION} & \multirow{2}{*}{10M T-I pairs}  & global   condition & 47M & \multirow{2}{*}{-} & \multirow{2}{*}{-} \\

 & & & local condition & 412M & & \\
 
SUR-Adapter~\cite{zhong2023adapter} & \begin{tabular}[c]{@{}c@{}}Lexica/civitai/\\Stable   Diffusion Online \end{tabular} & 57K T-I-T sets & simple   prompt & 20M & RTX 3090 & 5K steps \\

IP-Adapter~\cite{ye2023ip} & LAION \&  COYO & 10M T-I pairs & image & 1.5M & 8 V100 & 1M steps \\


I2V-Adapter~\cite{guo2024i2v} & WebVid & 10M videos & first frame   image & - & - & - \\

FaceChain-ImaginelD~\cite{xu2024facechain} &\begin{tabular}[c]{@{}c@{}}MEAD/HDTF/\\VoxCeleb2 \end{tabular} & - & audio & - & 8 V100& 2.5d\\
 
\midrule
\multicolumn{7}{c}{\textbf{\textit{Domain Adaptation}}}\\
\midrule

X-Adapter~\cite{ran2024x} & LAION & 300K images & spatial feature & 213M & 4 A100 & 2 epochs \\

\multirow{2}{*}{Ctrl-Adapter~\cite{lin2024ctrl}} & Panda & 200K videos & \multirow{2}{*}{spatial feature} & \multirow{2}{*}{184M} & \multirow{2}{*}{80G A100}  & \multirow{2}{*}{10h}  \\

 & LAION POP & 300K images & & & &  \\

\bottomrule
\end{tabular}
}
\caption{Comparison of various adapters and their applications.}
\label{tab:adapter}
\end{table}

\textbf{Controllable generation} Adapters effectively map diverse conditions into meaningful regions within the conditional space of diffusion models. For image generation, T2I-Adapter~\cite{mou2024t2i} captures conditional features and maps control feature to internal knowledge of the T2I model, achieving visual control of image generation. StableSketching~\cite{koley2024s} transforms semantic information from abstract sketch into textual conditional embedding and further constrains control features to pixel-perfect and textually meaningful regions in embedding space. SUR-Adapter~\cite{zhong2023adapter} effectively navigates simple prompt features toward a more information-dense region within the conditional space, enabling the generation of highly detailed images from simple prompts. IP-Adapter~\cite{ye2023ip} maps image features into a decoupled conditional space, enabling the model to generate images that resemble the input image. In the field of video synthesis, I2V-Adapter~\cite{guo2024i2v} aligns each frame of the video with the semantic information of the image condition, enhancing the overall coherence across frames. FaceChain-ImagineID~\cite{xu2024facechain} introduces a textual inversion adapter to convert speech text embeddings into token embeddings. Simultaneously, a spatial conditional adapter maps facial mesh, identity features, and masked adjacent frame features into the conditional space, maintaining audio-visual consistency and spatial coherence throughout the video. In summary, adapters play a crucial role in injecting a wide array of conditions into  diffusion models, significantly enhancing the control and quality of generated content in images and videos.

\textbf{Domain adaptation} Adapters in domain adaptation serve to align feature representations, enable task-specific adjustments, and facilitate efficient and effective transfer of knowledge from a source domain to a target domain. X-Adapter~\cite{ran2024x} establishes a mapping relationship between the spatial features of the base diffusion model and those of the upgraded diffusion model. Ctrl-Adapter~\cite{lin2024ctrl} integrates the features from a pre-trained image ControlNet into the framework of a target video diffusion model, facilitating multi-conditional control in video generation. Overall, by establishing mappings between different feature spaces, adapters enhance the flexibility of diffusion models across diverse applications.

\subsubsection{Low Rank Adaption}
\label{sec:LoRA}

Based on the recent observations~\cite{aghajanyan2020intrinsic, li2018measuring} that over-parameterized models operate in a low-dimensional subspace, LoRA~\cite{hu2021lora} learns parameter offsets using low-rank matrices and assumes that the weight update $\Delta W$ during fine-tuning can be represented as a low-rank decomposition of two smaller matrices $A \in \mathbb{R}^{d \times r}$ and $B \in \mathbb{R}^{r \times k}$, such that $\Delta W = A \times B$. The fine-tuned weight matrix becomes $W = W_0 + A \times B$ , where $W_0$ is the original pretrained weight. By restricting $A$ and $B$ to have low-rank $r$, where $r \ll \min(d, k)$, LoRA reduces the number of trainable parameters and computational overhead during fine-tuning. Instead of freezing diffusion models and inserting new trainable modules to prevent catastrophic forgetting~\cite{zhang2023adding, mou2024t2i}, LoRA allows the learned weight update to be merged back into the original model after training, avoiding the need for additional inference time. Therefore, it has been widely applied to various downstream tasks, as shown in the table \ref{tab:lora}.

\begin{figure}[h]
\centering
\includegraphics[width=1.00\linewidth]{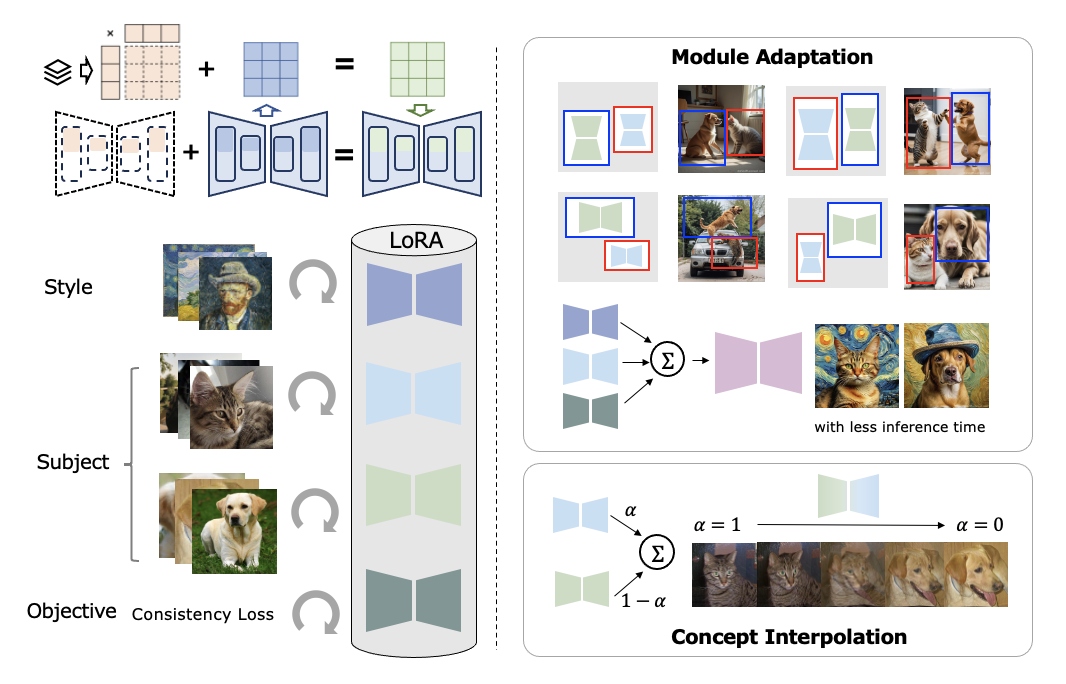}
\caption{An illustration of the mechanism combining LoRA and diffusion models, where LoRA fine-tunes the diffusion model to adapt to various customized tasks such as style, subject and other objective, and exceeds in modular adaptation and conceptual interpolation.}
\label{fig:LoRA}
\end{figure}

\begin{table}[H]
\centering
\scalebox{0.7}{
\begin{tabular}{ccccc}
\toprule
\textbf{Method} & \textbf{Year} & \textbf{Base Model} & \textbf{Downstream Task} & \textbf{Code} \\
\midrule
\multicolumn{5}{c}{\textbf{\textit{Image Generation}}} \\  
\midrule

 & & & Image   Generation, Depth Map Guided & \\
 & & & Image Generation, Canny Edge Guided & \\
\multirow{-3}{*}{Control-LoRA} & \multirow{-3}{*}{2023} & \multirow{-3}{*}{ControlNet}  & Recolor & \multirow{-3}{*}{\href{https://huggingface.co/stabilityai/control-lora}{[code]} } \\
 &  & Dreamshaper 7 & & \href{https://huggingface.co/latent-consistency/lcm-lora-sdv1-5}{[code]}  \\
 & & SSD 1B & & \href{https://huggingface.co/latent-consistency/lcm-lora-ssd-1b}{[code]} \\
\multirow{-3}{*}{LCM-LoRA~\cite{luo2023lcm}} & \multirow{-3}{*}{2023} & SDXL v1.0 & \multirow{-3}{*}{\begin{tabular}[c]{@{}c@{}}Fast   Image Generation\\ (Text-to-Image, Inpainting, styled-Generation)\end{tabular}} & \href{https://huggingface.co/latent-consistency/lcm-lora-sdxl}{[code]} \\
Concept   Sliders~\cite{gandikota2023concept} & 2023 & SD & Customized Attribute Editing & \href{https://github.com/rohitgandikota/sliders}{[code]}\\
LoRA-Composer~\cite{yang2024lora} & 2024 & SD & Multi-Concept Customization & \href{https://github.com/Young98CN/LoRA_Composer}{[code]}\\
ZipLoRA~\cite{shah2025ziplora} & 2023 & SDXL v1.0 & Subject \& Style Composed Customization &  \href{https://github.com/mkshing/ziplora-pytorch}{[code]} \\
Mix-of-Show~\cite{gu2024mix} & NeurIPS 2023 & SD v1.5 & Multi-Concept Customization & \href{https://github.com/TencentARC/Mix-of-Show}{[code]}\\
C-LoRA~\cite{smith2023continual} & 2023 & SD & Continual Concept   Customization & - \\
Intrinsic LoRA~\cite{du2024intrinsic}  & 2023 & SD(v1.1, v1.2, v1.5) & Image Normals, Depth, Albedo, Shading Generation & - \\
Smooth Diffusion~\cite{guo2024smooth} & CVPR 2024 & SD & Latent Space Interpolation, Image Inversion, Image Editing & \href{https://github.com/SHI-Labs/Smooth-Diffusion}{[code]} \\
DiffMorpher~\cite{zhang2024diffmorpher} & CVPR 2024 & SD v2.1 & Image Morphing & \href{https://github.com/Kevin-thu/DiffMorpher}{[code]} \\

\midrule
\multicolumn{5}{c}{\textbf{\textit{Video Generation}}} \\  
\midrule

AnimateDiff~\cite{guo2023animatediff} & ICLR 2024 & SD & Personalized   Style \& Motion Guided, Animation Generation & \href{https://github.com/guoyww/AnimateDiff}{[code]} \\
DragVideo~\cite{deng2023dragvideo}   & 2023      & AnimateDiff        & Sample-Specific,   Video Generation                         & \href{https://github.com/rickyskywalker/dragvideo-official}{[code]}\\
MagicStick~\cite{ma2023magicstick}  & 2023      & SD & Scenes-Specific, Video Generation                           & - \\                                                                                

\midrule
\multicolumn{5}{c}{\textbf{\textit{3D Synthesis}}} \\  
\midrule

ProlificDreamer~\cite{wang2024prolificdreamer} & NeurIPS 2023 & SD & Rendered 2D   Image Generation, Text \& Camera Pose Guided & \href{https://ml.cs.tsinghua.edu.cn/prolificdreamer/}{[code]} \\
Boosting3D~\cite{yu2023boosting3d} & 2023 & SD & Rendered Image   Generation, Text \& Camera Pose Guided & - \\
3DFuse~\cite{seo2023let} & 2023 & SD      & Rendered Image   Generation, Text \& Sparse Depth Map Guided & \href{https://github.com/KU-CVLAB/3DFuse}{[code]} \\
DreamControl~\cite{huang2024dreamcontrol} & CVPR 2024 & SD v1.5 & Rendered Image   Generation, Text \& Normal Map Guided &\href{https://github.com/tyhuang0428/DreamControl}{[code]} \\

\bottomrule
\end{tabular}
}
\caption{The statistics for LoRA methods utilized in recent research}
\label{tab:lora}
\end{table}

\textbf{Module adaptation} Benefiting from the low-rank property, multiple LoRA parameters, which are fine-tuned on different datasets or downstream tasks, can be directly combined to produce composition capability as shown in Figure \ref{fig:LoRA}. 
LCM-LoRA~\cite{luo2023lcm} can generate images in a specific style while supporting fast inference with minimal steps, by linearly combining the style-related LoRA parameter and acceleration LoRA parameter. 
AnimateDiff~\cite{guo2023animatediff} trains individual LoRAs to specialize in distinct motion patterns. During inference, these specialized LoRAs can be synergistically combined, enabling the generation of diverse and complex motion effects.
LoRA-Composer~\cite{yang2024lora} integrates multiple concept-specific LoRAs into the image generation process, ensuring each concept is accurately rendered in terms of position, size, and distinctive features.
These methods enable multiple LoRAs to seamlessly generate different concepts in various regions of an image, or to combine distinct characteristics during the image generation process, fully exploiting the composable nature of LoRAs.

\textbf{Concept interpolation} LoRA approximates the update direction within a compact and structured parameter space through low-rank decomposition. As shown in Figure \ref{fig:LoRA}, when performing linear interpolation between LoRA parameters for different concepts, the resulting intermediate parameters smoothly blend the features of the original parameter sets.
Concept Sliders~\cite{gandikota2023concept} subsequently modifies the concept along specific parameter direction by scaling the guidance coefficient in training loss and the hyperparameters of LoRA.
DiffMorpher~\cite{zhang2024diffmorpher} discovers LoRA has the capability to encapsulate image semantic identity, and achieves image morphing by performing linear interpolation on the LoRA parameters adapted to different concepts. 
Collectively, these advancements demonstrate that LoRA offers significant flexibility and control in the field of image generation and editing, allowing creators to achieve smoother transitions between different concepts.

\subsection{Label-Efficient Methods}

The scarcity of data can negatively impact the generation quality of diffusion models, leading to the development of two key strategies for efficient adaptation to downstream tasks with minimal labeling. One strategy is preference optimization, which trains annotation models (like reward models) to replace human annotations and uses reinforcement learning to continuously supervise the training of diffusion models to meet human preferences. The other is personalized training, which optimizes the learning process to extract the most salient features from small datasets while preserving the generative capabilities of diffusion models.

\label{subsec:4_2}
\begin{figure}[h]
\centering
\includegraphics[width=1.00\linewidth]{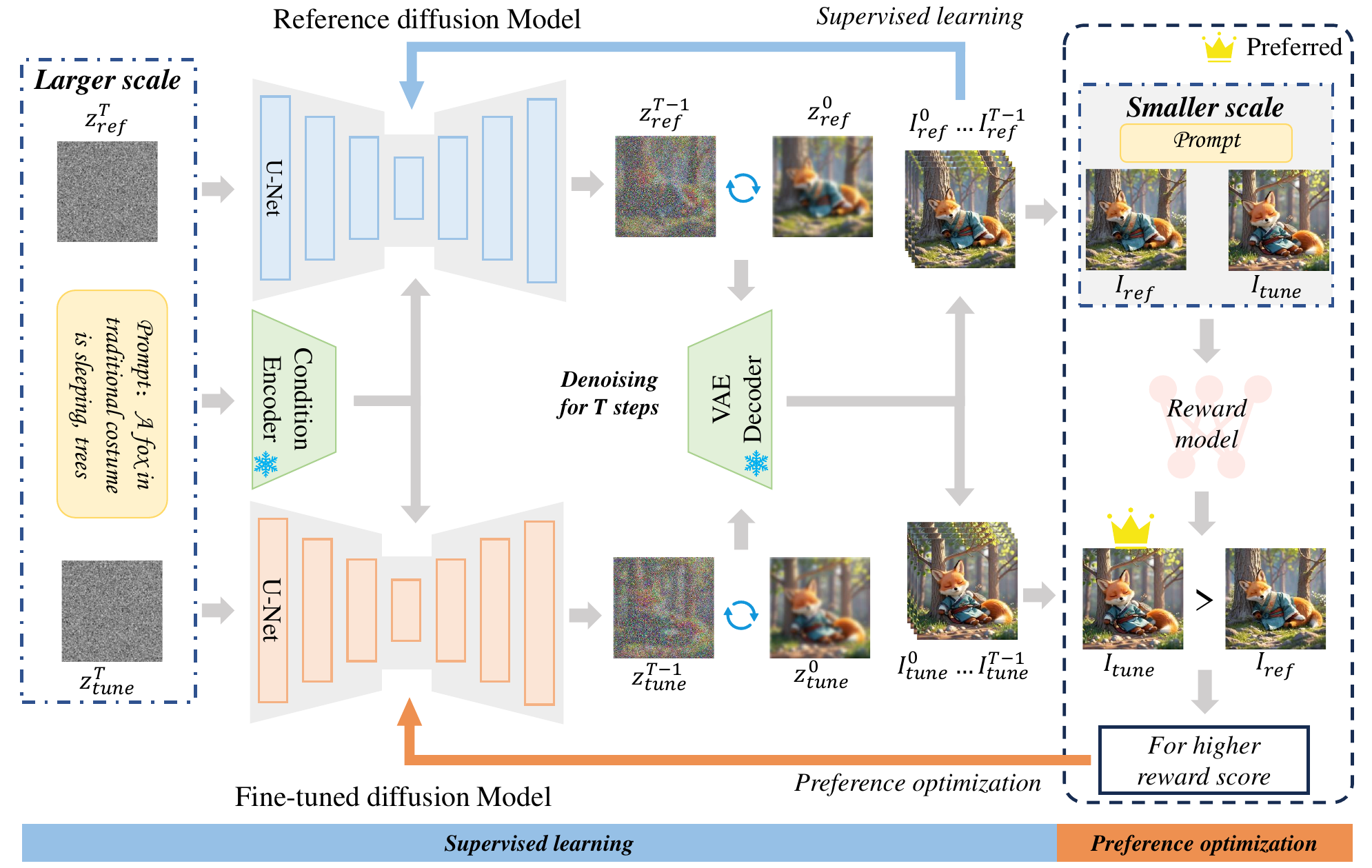}
\caption{A generic training framework for preference optimization. The reward model scores the generated images directly without human annotation, saving annotation cost and time.}
\label{fig:preference_optimization_framework}
\end{figure}
\subsubsection{Preference Optimization}
\label{sec:PreferenceOptimization}

\begin{figure}[t!]
\centering
\includegraphics[width=1.00\linewidth]{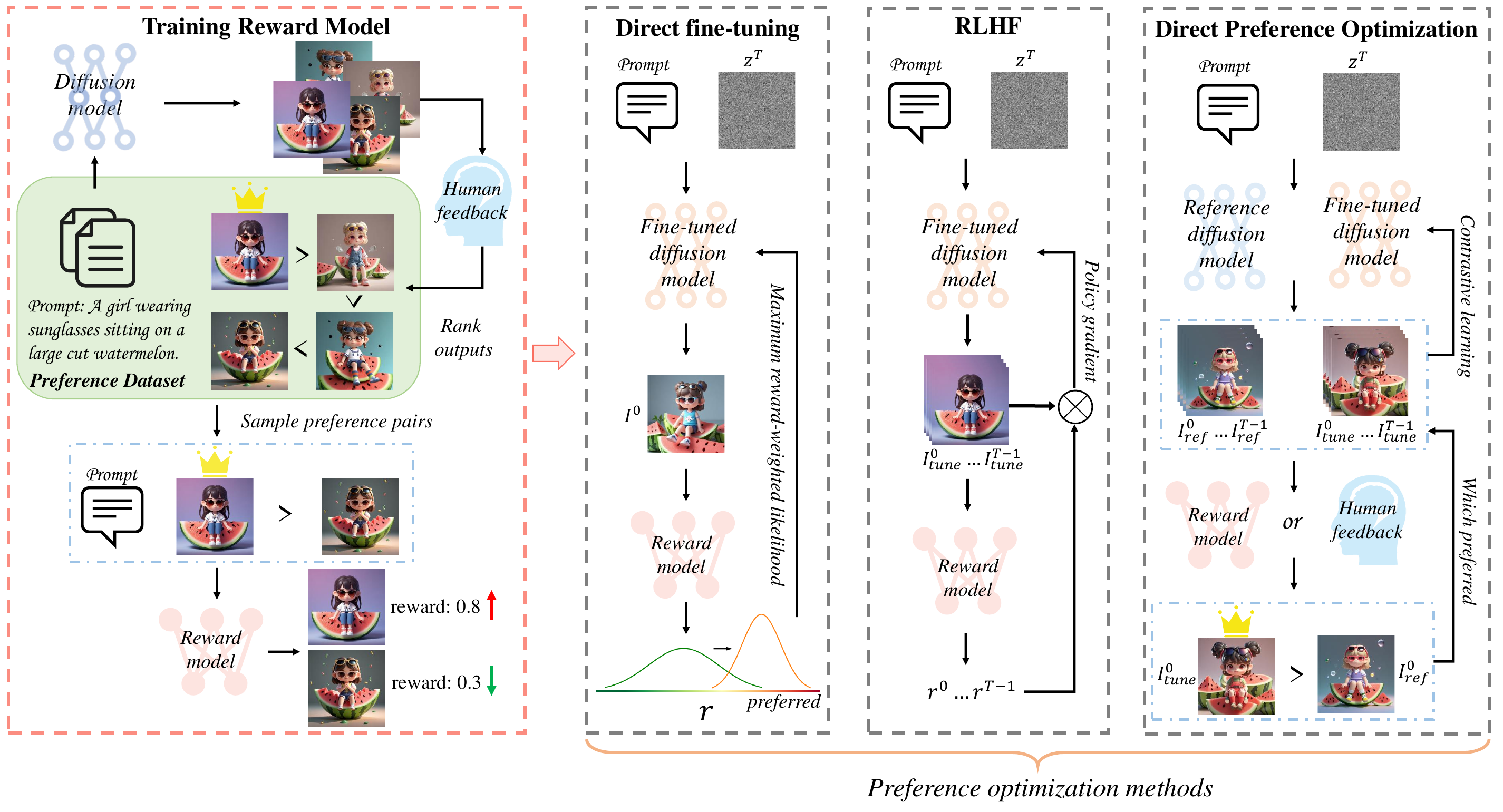}
\caption{The illustrations of preference optimization paradigms. The trained reward model can be used in subsequent various preference optimization methods.}
\label{fig: rl_rlhf_dft}
\end{figure}

Diffusion models mainly utilize the variational lower bound on the log-likelihood expressed in Equation~\ref{eq: Eq.2} to approximate the target data distribution. While a decrease in training loss indicates that the model is learning certain patterns, it does not necessarily mean that the generated images meet human aesthetic standards. Therefore, the preference training framework shown in Figure \ref{fig:preference_optimization_framework} has become a critical approach for aligning models with human expectations. 
The current process for preference optimization in image generation tasks is generally divided into two steps. First, human aesthetic preferences are formalized into a reward model~\cite{wu2023human,lee2023aligning,xu2024imagereward,kirstain2023pick}, which reduces the cost and time required for labeling data in the subsequent stages.
Secondly, the direct fine-tuning on preferred outputs~\cite{dong2023raft,wu2023human,lee2023aligning,xu2024imagereward} or reinforcement learning from human feedback (RLHF)~\cite{black2023training,fan2024reinforcement,yang2024using,wallace2024diffusion} are employed to optimize the diffusion model against the reward model. Figure.\ref{fig: rl_rlhf_dft} illustrates the paradigms of each category.
These methods avoid the complex computational burden associated with direct supervised training on large datasets labeled with preference tags, making preference optimization an efficient method.

\textbf{Reward model}. It is crucial that the reward model effectively encodes human preferences, as this directly impacts the diffusion model's ability to correctly learn and reflect individual aesthetics. The general idea of human preference modeling is to maximize the difference that the reward score of a preferred image $I_w$ with prompt condition $T$ is greater than the other outputs $I_l$ for any sample from the preference dataset, formulated as follows: 
\begin{equation}
\label{eq: L_reward}
\mathrm{L}_{reward}=-\mathbb{E}_{(T,I_{w},I_{l})\sim\mathcal{D}}[\mathrm{log}(\sigma(R(T,I_{w})-R(T,I_{l})))].
\end{equation}
where $\sigma$ indicates the activation function and $R$ represents the reward model. HPS~\cite{wu2023human} fine-tunes the CLIP model using training data that includes text prompts and multiple images (one preferred and the others non-preferred), enabling the model to obtain a human preference score. AHF~\cite{lee2023aligning} creates text-image groups with binary feedback datasets to train the CLIP model, applying the mean squared error (MSE) loss for accuracy and the cross-entropy loss to improve generalization to unseen data. ImageReward~\cite{xu2024imagereward} employs a scoring system where higher rankings yield higher scores to train the BLIP model, utilizing a text-images dataset with ratings and rankings, allowing for a finer distinction in image quality. Pick-a-Pic~\cite{kirstain2023pick} is a large dataset where each instance includes a prompt, two generated images, and a label indicating preference or tie. It is employed to fine-tune the Pick Score, a reward model based CLIP-H, with the objective of minimizing the KL divergence between the preference label and the softmax normalized scores of the two images.

\textbf{Direct fine-tuning} has achieved remarkable preformance by leveraging a reward model for supervised learning. RAFT~\cite{dong2023raft}, in each iteration, uses a reward model to filter the K samples generated by the diffusion model, selects the best-of-K sample for fine-tuning the model, thereby avoiding the overfitting problem when fine-tuning with datasets devoid of preference labels. AHF~\cite{lee2023aligning} introduces the negative reward-weighted log-likelihood into the loss function of preference optimization to improve the image-text alignment of the model. ImageReward~\cite{xu2024imagereward}, during the refinement phase of diffusion models utilizes ReFL loss and regularization with pre-training loss to prevent rapid overfitting and stabilize fine-tuning. HPS~\cite{wu2023human} suggests incorporating a special identifier in the prompts during fine-tuning to distinguish preferred images. During inference, these special identifier serve as negative prompts for classifier-free guidance, effectively preventing the generation of non-preferred images.

\textbf{Reinforcement learning from human feedback} (RLHF) uses policy gradient to optimize human-preferred policy aimed at maximizing the reward model's scoring of generated images. DDPO~\cite{black2023training} reframes the denoising process of diffusion model as a multi-step Markov decision process and employs importance sampling techniques to optimize it. This algorithm serves as a versatile framework for optimizing any downstream objective, covering aspects such as compressibility, aesthetic quality, and text alignment. 
DPOK~\cite{fan2024reinforcement} introduces two critical improvements over DPPO. Firstly, it incorporates KL regularization into the loss function, effectively curbing the model's tendency to overfit to rewards. Secondly, by additionally training a value function, it not only significantly reduces the variance in gradient estimation but also further enhances the performance of the final reward. 
D3PO~\cite{yang2024using} overcomes the application obstacles of DPO~\cite{rafailov2024direct} in diffusion models without the need for pre-trained reward models. It trains through online learning, leveraging real-time preference annotations from experts on two images generated from the same text.
Diffusion-DPO~\cite{wallace2024diffusion} directly optimizes the policy that aligns more closely with human preferences during the single-step denoising process, effectively solving the problem of prolonged training time due to the need for multi-step reverse denoising in previous methods.

\subsubsection{Personalized Training}
\label{sec:PersonalizedTraining}
The primary challenge of personalized generation based on diffusion model lies in data scarcity, as high-quality training data is often difficult to obtain. To tackle this, personalized training methods~\cite{gal2022image, ruiz2023dreambooth, ye2023ip, wei2023elite, li2024photomaker, ruiz2024hyperdreambooth, li2024blip, wang2024instantid, kong2024omg} tailor the learning process to achieve high performance with less data by focusing on relevant and personalized information rather than generalizing across a broad dataset, significantly reducing the need for large amounts of individual data. In this section, we present two mainstream personalized synthesis methods and discuss their contributions to label-efficient approaches.

Fine-tuning-based personalization approaches have focused on fine-tuning pre-trained diffusion models to learn a placeholder token that captures the identity information of reference subjects. For instance, DreamBooth~\cite{ruiz2023dreambooth} conducts full fine-tuning, Textual Inversion~\cite{gal2022image} adjusts the embeddings of pseudo-words, and Custom-Diffusion~\cite{kumari2023multi} optimizes the key, value mapping matrices within cross-attention layers. Moreover, LoRA~\cite{hu2021lora} introduces a minimal number of trainable parameters and trains individually on a few customized datasets, facilitating the widespread utilization of LoRA for customization. Recent efforts~\cite{kong2024omg,gu2024mix,yang2024lora,ostashev2024moa} focus on achieving multi-concept customization by combining LoRA weights from different concepts, aiming to improve identity preservation, handle occlusions, and enhance foreground-background harmony.

However, fine-tuning-based approaches often require training for thousands of steps to customize concepts and most of them rely solely on a placeholder token embedding which proves insufficient for effectively decoupling specific concepts from their background layouts. To address this, encoder-based methods~\cite{wei2023elite,gal2023designing,hua2023dreamtuner,li2024blip,wang2024instantid,li2024photomaker,shi2024instantbooth} utilize additional image encoders to inject the reference image details for subject generation. ELITE~\cite{wei2023elite} and DreamTuner~\cite{hua2023dreamtuner} adopt a strategy that progressively extracts visual information of target features, from coarse to fine, enabling more precise and controllable subject-driven image generation. Meanwhile, BLIP Diffusion~\cite{li2024blip} uses a multimodal encoder (\textit{i.e.}. Q-former~\cite{li2022blip}) to filter out background information, focusing on learning the intricate details of the intended concepts.

\section{Efficient Sampling and Inference}
\label{sec:5}
%
Representative diffusion models often require numerous iterations for de-noising~\cite{ho2020denoising}, which hinders their practical application~\cite{yang2023diffusion}.
Consequently, researchers devote to the efficient sampling methods~\cite{salimansprogressive, meng2023distillation, song2023consistency, luo2023latent, luo2023lcm, sauer2023adversarial, liuflow, liu2023instaflow, xu2024ufogen, lu2022dpm, karras2022elucidating} that can reduce the number of iterations during the inference stage while maintaining the model's ability to generate high-quality images.
We summarize four types of methods and illustrate them for efficient sampling and inference below.

\begin{figure}[!t]
\centering
\includegraphics[width=1.0\linewidth]{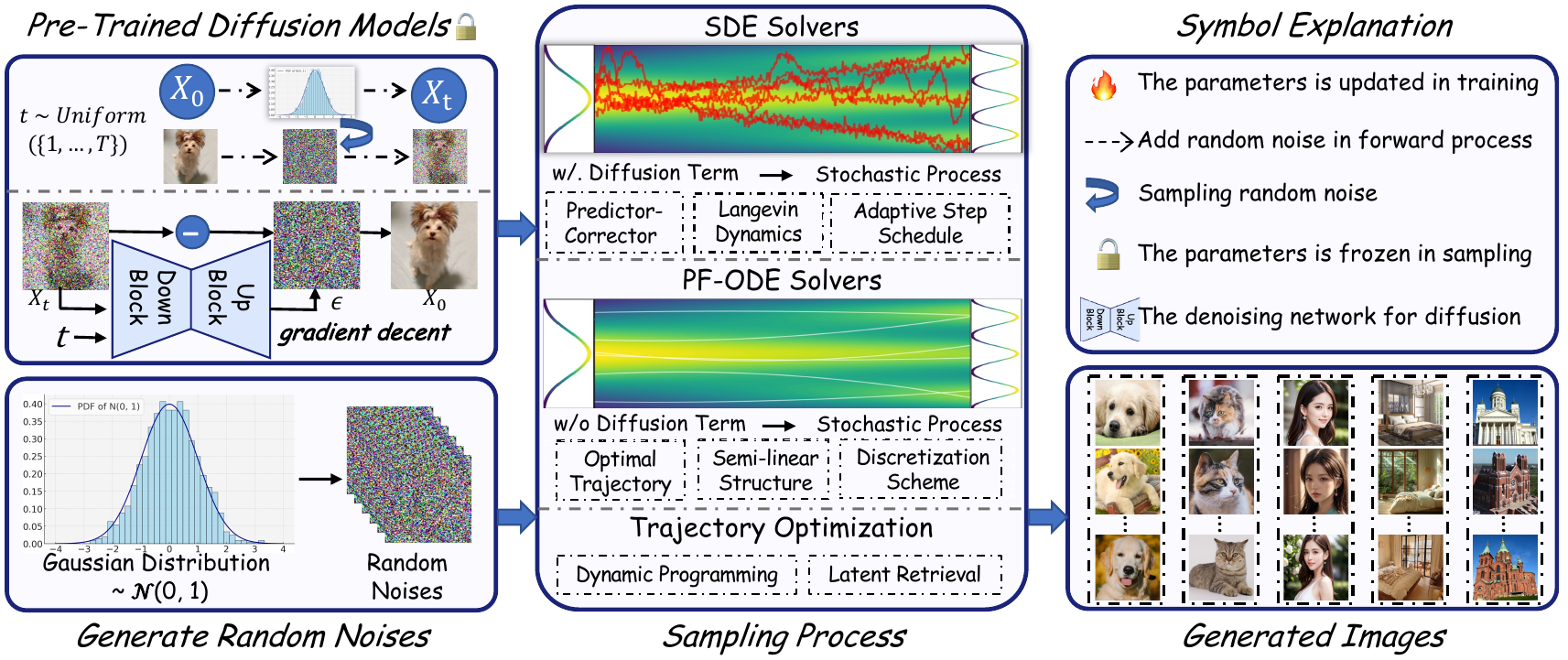}
\caption{The illustration of training-free methods.}
\label{fig:training-free}
\end{figure}

\subsection{Training Free Method}
\label{subsec:training-free}
As illustrated in \ref{subsec:2.1}, the DMs can be defined as a continuous-time process from the perspectives of SDE and PF-ODE.
Many works~\cite{songscore, songdenoising, karras2022elucidating, lu2022dpm} accelerate sampling process by solving the discretized differential equations.

\textbf{SDE solver} is a numerical method used to approximate the solution of an Stochastic Differential Equation (SDE). It discretizes the continuous-time SDE into multiple time steps, enabling efficient sampling from noise to data. The SDE is fundamental to both the forward and reverse processes in generative modeling. Song et. al.~\cite{songscore} unified previous generative models into a common mathematical framework via SDEs in Eqs.~\ref{eq: Eq.5}. Specifically, the forward and reverse processes in DDPM~\cite{ho2020denoising} and SMLD~\cite{song2019generative} are discretizations of the following SDEs:
\begin{equation}
    \text{DDPM: } \begin{cases}
        \text{forward: } d\mathbf{x}_t=-{\frac 1 2}\beta(t)\mathbf{x}_tdt+\sqrt{\beta(t)}dw,\\
        \text{reverse: }\ d\mathbf{x}_t=-{\frac 1 2}\beta(t)[\mathbf{x}_t-\nabla_{\mathbf{x}_t} \log p_t(\mathbf{x})]dt+\sqrt{\beta(t)}d\bar{w};
    \end{cases}
\end{equation}
\begin{equation}
    \text{SMLD: } \begin{cases}
        \text{forward: } d\mathbf{x}_t=\sqrt{{\frac {d(\sigma^2(t))} {dt}}} dw,\\
        \text{reverse: }\ d\mathbf{x}_t= -{\frac {d(\sigma^2(t))} {dt}}\nabla_{\mathbf{x}_t} \log p_t(\mathbf{x}) dt+ \sqrt{{\frac {d(\sigma^2(t))} {dt}}}d\bar{w}.
    \end{cases}
\end{equation}
By carefully designing the SDE and its discretization scheme, the SDE solver seeks to balance the number of steps and approximation errors, thereby improving both the efficiency and quality of outputs in diffusion models.

Noise-Conditional Score Networks (NCSNs)~\cite{song2019generative} generate new data points through Langevin dynamics, using score matching to estimate the gradient of the data distribution. NCSNs identified three issues when data lies on a low-dimensional manifold: (1) the score function is undefined in low data-density regions; (2) due to the sparsity of training data, score estimation in low-density regions is inaccurate; and (3) in Langevin dynamics, it is difficult to effectively mix different modes of the distribution. To address these problems, NCSNs introduce multi-level noise to perturb the data and adopt Annealed Langevin Dynamics (ALD), where sampling starts with the score corresponding to the highest noise level, and the noise is gradually reduced until convergence to the original data distribution. Building upon this, Jolicoeur-Martineau et. al.~\cite{jolicoeuradversarial} discussed the inconsistencies in noise scaling within ALD and proposed Consistent Annealed Sampling (CAS), a score-based MCMC method that ensures noise levels follow a predefined schedule, providing a more stable alternative to ALD.

\begin{table*}[t!]
\footnotesize
\centering
\label{tab:training-free sampling}
\scalebox{0.9}{
\begin{tabular}{c r ccccccccc}
\toprule

\multicolumn{2}{c}{\multirow{2}{*}{\textbf{\textit{SDE Solver}}}} & \multicolumn{9}{c}{CIFAR-10} \\

\cmidrule(lr){3-11}
&  & 35 & 50 & 100 & 232 & 275 & 500 & 1000 & 1160 & 2000  \\ \midrule

Score SDE~\cite{songscore} & ICLR20 & - & - & - & - & - & - & 3.21 & - & 3.10 \\ 
CLD~\cite{dockhornscore} & ICLR21 & - & 52.70 & - & - & 3.24 & 2.41 & 2.27 & - & 2.23 \\ 
DSM-ALS~\cite{jolicoeuradversarial} & ICLR21 & - & - & - & 7.50 & - & - & - & 5.60 & - \\ 
Gotta Go Fast~\cite{jolicoeur2021gotta} & ArXiv21 & - & 72.29 & - & - & 2.74 & - & - & - & - \\ 
NCSN~\cite{song2019generative} & NeurIPS19  & - & - & - & - & - & - & 25.32 & - & - \\ 
EDM~\cite{karras2022elucidating} & NeurIPS22  & 1.97 & - & - & - & - & - & - & - & - \\ 
\midrule 

\multicolumn{2}{c}{\multirow{2}{*}{\textbf{\textit{ODE Solver}}}} & \multicolumn{3}{c}{CIFAR-10} & \multicolumn{3}{c}{CelebA}  & \multicolumn{3}{c}{LSUN} \\

\cmidrule(lr){3-5}  \cmidrule(lr){6-8}  \cmidrule(lr){9-11}
&  & 1 & 2 & 4 & 1 & 2 & 4 & 1 & 2 & 4  \\ \midrule

DDIM~\cite{songdenoising} & ICLR21 & 13.68 & 6.84 & 4.67 & 17.33 & 13.73 & 9.17 & 19.95 & 8.89 & 6.75  \\
PNDM~\cite{liupseudo} & ICLR22 & 7.05 & 4.61 & 3.68 & 7.71 & 5.51 & 3.34 & 8.69 & 9.13 & 9.89 \\
DPM-Solver~\cite{lu2022dpm} & NeurIPS22 & 6.37 & 4.28 & 3.90 & 7.15 & 4.40 & 4.23 & 6.10 & 3.09 & 2.53 \\
gDDIM~\cite{zhanggddim} & ICLR23 & 41.7 & 3.03 & 2.59 & - & - & - & - & - & - \\
DEIS~\cite{zhangfast} & ICLR23 & 4.17 & 3.33 & 3.36 & 6.95 & 3.41 & 2.95 & - & - & - \\
UniPC~\cite{zhao2023unipc} & NeurIPS23 & 3.87 & - & - & - & - & - & 3.54 & - & - \\
NonUniform~\cite{xue2024accelerating} & CVPR24 & 3.50 & - & - & - & - & - & -  & - & -\\
\midrule

\multicolumn{2}{c}{\multirow{2}{*}{\textbf{\textit{Trajectory Optimization}}}} & \multicolumn{3}{c}{CIFAR-10} & \multicolumn{3}{c}{ImageNet}  & \multicolumn{3}{c}{MS-COCO} \\

\cmidrule(lr){3-5}  \cmidrule(lr){6-8}  \cmidrule(lr){9-11}
&  & 5 & 10 & 20 & 5 & 10 & 20 & 20 & 30 & 40  \\ \midrule

GGDM~\cite{watson2022learning} & ICLR22 & 13.77 & 8.23 & 4.72 & 55.14 & 37.32 & 20.69 & - & - & - \\
ReDi~\cite{zhang2023redi} & ICML23 & - & - & - & - & - & - & 25.50 & 24.70 & 25.20 \\
\bottomrule

\end{tabular}
}

\caption{Three types of \textbf{training-free} methods are summarized. We further present the generation performance of these methods in terms of efficiency, i.e., Neural Function Evaluations (NFEs $\downarrow$) and quality i.e., Fréchet Inception Distance (FID $\downarrow$). }

\end{table*}

Also building on Langevin dynamics, Dockhorn et al. introduced Critically-damped Langevin Diffusion (CLD)~\cite{dockhornscore}. As proved in~\cite{songscore}, the score function learnt by the neural network is uniquely determined by the forward process, CLD thus posits that a smoother forward process can lead to faster and more efficient sample generation. Inspired by statistical mechanics, CLD introduces a novel SDE by incorporating a velocity variable $v_t$, enabling diffusion in the joint data-velocity space ($x_t - v_t$). In CLD, noise is only injected into $v_t$, thereby avoiding the oscillations of under-damped systems and the slow dynamics of over-damped systems. Additionally, CLD only needs to learn the gradient of the velocity distribution $\nabla_{v_t} \log p_t(v_t | x_t)$ given the data, which is arguably simpler than learning the score function of the diffused data directly. This method combines Hamiltonian dynamics with the Ornstein-Uhlenbeck process, efficiently exploring the state space and ensuring convergence, thus enabling more efficient sampling and high-quality data generation.

The predictor-corrector method proposed in~\cite{songscore} solves the reverse-time SDE by alternating a numerical SDE solvers (``predictor'') and a score-based Markov Chain Monte Carlo (``corrector''). At each time step, the predictor, such as Euler-Maruyama and stochastic Runge-Kutta methods, approximates the reverse-time SDE, providing an estimate of the sample $x_t$ at the next time step $t$. Then a score-based corrector refines the marginal distribution of $x_t$. The predictor enables fast convergence and the corrector ensures sample diversity and quality. The resulting samples maintain the same time marginals as the solution to the reverse-time SDE, which allows them to closely align with the target distribution during the actual generation process. EDM~\cite{karras2022elucidating} combines a second-order deterministic ODE integrator with a Langevin-like ``churn'' perturbation of alternatively adding and removing noise. This approach improves the corrector from~\cite{songscore}, achieving state-of-the-art generation quality at the time. 

Another issue of numerical SDE solvers is that they require large number of score network evaluations. Jolicoeur-Martineau et. al.~\cite{jolicoeur2021gotta} devise an SDE solver with adaptive step sizes to accelerate the generation process. The step size is determined by comparing the outputs of a low-order solver and a high-order solver. At each step of the generation process, the solver generates both low-order sample $x'_l$ and high-order sample $x'_h$ from the previous sample $x'_{prev}$. The error between these two samples is then evaluated via:
\begin{equation}
    E_q = \left\| \frac {{x'_l}-{x'_h}} {\delta({x'_l},{x'_{prev}})}\right\|_2\ ,\  \delta(x'_l,x'_{prev})=\text{max}(\epsilon_{abs},\epsilon_{rel}\ \text{max}(|x'_l|,|x'_{prev}|)),
\end{equation}
where $\epsilon_{abs}$ and $\epsilon_{rel}$ are absolute and relative tolerance. If $x'_l$ and $x'_h$ are similar, then $x'_h$ is accepted and the step size will be increased.

In a more specific situation, CCDF~\cite{chung2022come} focuses on efficient sampling in conditional image generation tasks by leveraging the contraction property of the reverse diffusion path. It proposes that the generation process does not need to start from pure Gaussian noise but can significantly reduce sampling steps by starting from an initialization closer to the target. The input image is first perturbed with noise up to $t_0$ (where $t_0 < T$ , and this noise addition process is nearly ``free''), and then reverse denoising starts from $t_0$ to generate the conditional image. As a result, generating target images needs far fewer steps than $T$. In super-resolution (SR), inpainting, and MRI reconstruction tasks, the method achieves excellent results with only 10, 20, and 20 reverse diffusion steps, respectively.

\begin{figure}[!t]
\centering
\includegraphics[width=1.0\linewidth]{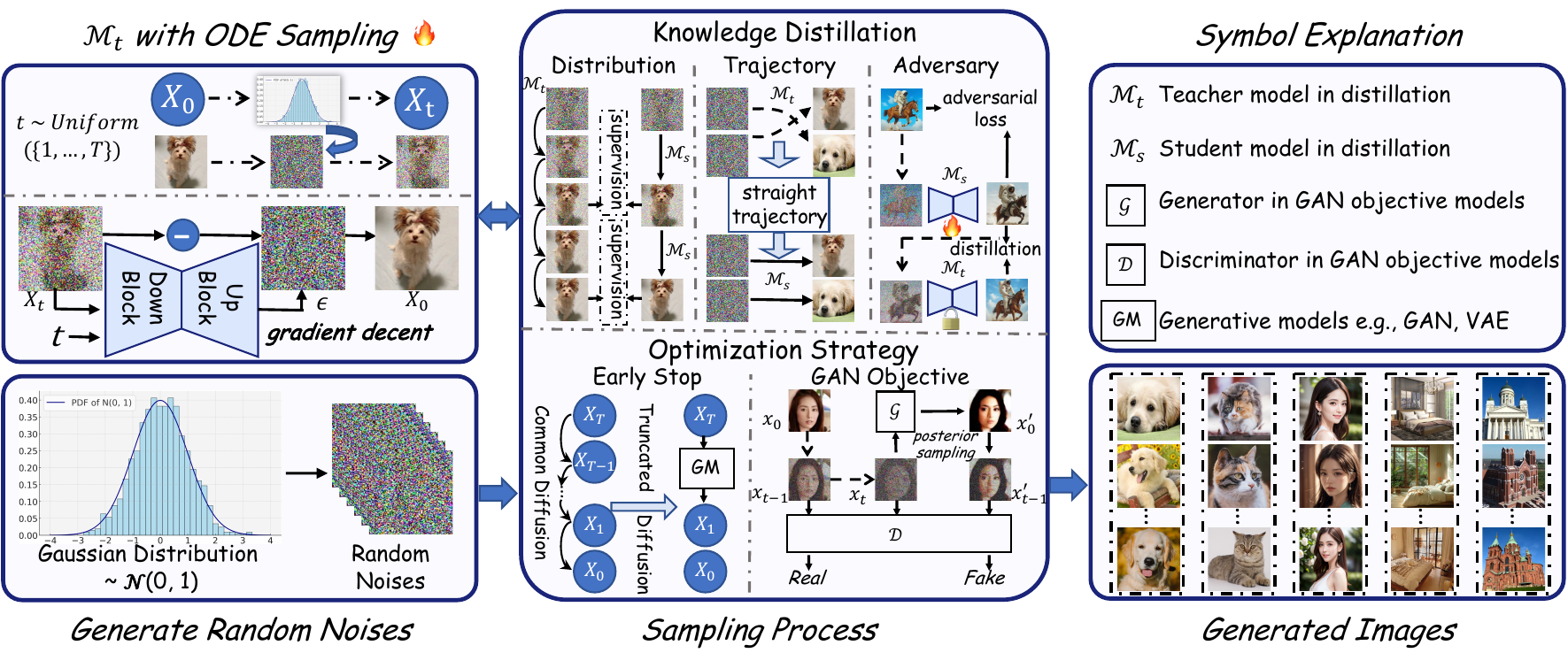}
\caption{The illustration of training-based methods.}
\label{fig:training-based}
\end{figure}

\textbf{PF-ODE solver} is one of the most commonly used strategy to accelerate the sampling process~\cite{songdenoising, liupseudo, karras2022elucidating, lu2022dpm, lu2022dpmplus, zhanggddim, zhangfast, zhang2023redi}.
Different from SDE Solver, the sampling process of PF-ODE solvers is deterministic, hence is suitable to serve as the teacher model in the knowledge distillation methods~\cite{dockhorn2022genie, salimansprogressive, luhman2021knowledge}.
Denoising Diffusion Implicit Models (DDIM)~\cite{songdenoising} is a notable faster diffusion sampling scheduler, which supports larger denoising steps via a non-Markovian diffusion processes.
Particularly, DDIM is a particular formulation of ODE, whose iteration can be rewrote as:

\begin{equation}
\sqrt{\frac{1}{\alpha_{t-1}}} \mathbf{x}_{t-1} = \sqrt{\frac{1}{\alpha_{t}}} \mathbf{x}_{t} + (\sqrt{\frac{1-\alpha_{t-1}}{\alpha_{t-1}}} - \sqrt{\frac{1-\alpha_{t}}{\alpha_{t}}} ) \epsilon^{(t)}_{\theta} (\mathbf{x}_{t}),
\label{eq:ODE1}
\end{equation}

After reparameterization, the equation can be transformed to the reverse of ODE.
Inspired by the observation that when the training dataset contains one sample, DDIM can exactly solve the corresponding SDEs/ODE, Zhang et.al. extend DDIM to general DMs, i.e.,   gDDIM~\cite{zhanggddim}.
Liu et.al.~\cite{liupseudo} discover two limitations of DDIM.
First, the denoising model and ODE are well-defined only in a limited area.
However, the sampling process with larger steps may generates samples away from the well-defined area, hence result in new errors.
Second, when the index t $\xrightarrow{}$ 0, the ODE equation tends to infinity in many higher-order numerical methods.
The phenomenon leads to additional error for the fine-grained  denoising steps.
To address these issues, PNDM~\cite{liupseudo} solve the ODE on certain manifolds, which is consists of gradient and transfer parts.
The former finds the gradient in each step, and the latter generates the result at the next step.
PNDM makes the sampling trajectory is more consistent with the pre-trained area, hence generating higher-quality images with skipped steps.
Further, DPM-Solver~\cite{lu2022dpm} and DEIS~\cite{zhangfast} calculate the exact solutions of the diffusion ODEs by semi-linear structure, hence the solvers support larger steps with less error.
Specifically, DPM-Solver finds that diffusion ODEs can be divided two parts, i.e. linear (drift coefficient) and non-linear (diffusion coefficient) functions.
Previous methods uniformly deal with these two parts, which causes discretization errors particularly on the linear part.
Actually, the part can be analytically computed.
For the non-linear part, DPM-Solver~\cite{lu2022dpm} simplifies the formulation by introducing log-SNR, which is a strictly decreasing function of t.
Next, Taylor expansion is utilized for approximating the non-linear part.
Moreover, DEIS~\cite{zhangfast} utilize high-order polynomial extrapolation to reduce the approximation error, which achieves better sampling quality.
Besides, to improve the quality of generated samples with accelerated sampling, UniPC~\cite{zhao2023unipc} utilizes the output $\epsilon_{\theta}(\mathbf{x}_t, t)$ at current timestep $t$ to correct the predicted sample.
NonUniform~\cite{xue2024accelerating} accelerates diffusion sampling by exploring discretization scheme for time steps, the orders of different steps can be different in the numerical ODE solvers.
\textbf{Retrieval based} methods retrieve trajectory from pre-computed knowledge base to accelerate the sampling process~\cite{zhang2023redi}.
Inspired by an crucial common sense that the previous sampling steps determine the layout of the images, and the following steps determine the details~\cite{mengsdedit, wu2023uncovering}.
ReDi~\cite{zhang2023redi} first proposes a retrieval based learning free acceleration strategy.
Specifically, the samples of first few steps are generated, which are utilized as the query in the retrieval process.
Sequentially, the top-$H$ keys that have highest similarity with the initialized query are selected.
Next, the linearly combined values are utilized as the remaining steps of the sampling process.

\begin{table*}[t!]
\footnotesize
\centering
\label{tab:training-based sampling}
\scalebox{0.9}{
\begin{tabular}{c r | ccc ccc ccc }
\toprule
\multicolumn{2}{c}{\multirow{2}{*}{\textbf{\textit{Distribution Based Distillation}}}} & \multicolumn{3}{c}{CIFAR-10} & \multicolumn{3}{c}{ImageNet}  & \multicolumn{3}{c}{MS-COCO} \\
\cmidrule(lr){3-5}  \cmidrule(lr){6-8}  \cmidrule(lr){9-11}
&  & 1 & 2 & 4 & 1 & 2 & 4 & 1 & 2 & 4  \\ \midrule
Denoising Student~\cite{luhman2021knowledge} &  ArXiv21 & 9.36 & - & - & - & - & - & - & - & -  \\
Progressive Distillation~\cite{salimansprogressive} & ICLR22 & 9.12 & 4.51 & 3.00 & 15.99 & 7.11 & 3.84 & 37.2 & 26.00 & 26.40 \\
Meng et al.~\cite{meng2023distillation} & CVPR23 & 7.34 & 4.23 & 3.58 & 22.74 & 4.14 & 2.79 & - & - & - \\
CM~\cite{song2023consistency} & ICML23 & 3.55 & 2.93 & - & 6.20 & 4.70 & - & 7.80 & 5.22 & -  \\
LCM~\cite{luo2023latent} & ArXiv23 & - & - & - & - & - & - & - & - & 23.49 \\
DMD~\cite{yin2024one} & CVPR24 & 2.62 & - & - & 2.62 & - & - & 11.49 & - & - \\
\midrule
\multicolumn{2}{c}{\multirow{2}{*}{\textbf{\textit{Trajectory Based Distillation}}}} & \multicolumn{3}{c}{CIFAR-10} & \multicolumn{2}{c}{ImageNet} & \multicolumn{2}{c}{LAION-A} & \multicolumn{2}{c}{MS-COCO} \\
\cmidrule(lr){3-5}  \cmidrule(lr){6-7}  \cmidrule(lr){8-9}  \cmidrule(lr){10-11}
&  & 1 & 2 & 4 & 1 & 4 & 4 & 8 & 4 & 8  \\ \midrule
TRACT~\cite{berthelot2023tract} & ArXiv23 & 3.78 & 3.32 & 2.93 & 7.43 & 4.97 & - & - & - & - \\
Rectified Flow~\cite{liuflow} & ICLR22 & 2.58 & - & - & - & - & - & - & - & - \\
InstaFlow~\cite{liu2023instaflow} & ICLR23 & - & - & - & - & - & 14.32 & 10.98 & 13.86 & 11.40 \\
DSNO~\cite{zheng2023fast} & ICML23 & 3.78 & - & - & 7.83 & - & - & - & - & - \\
SFT-PG~\cite{fan2023optimizing} & ICML23 & - & - & - & - & - & - & - & - & - \\
PeRFlow~\cite{yan2024perflow} & ArXiv24 & - & - & - & - & - & 8.60 & 8.52 & 11.31 & 14.16 \\
\midrule
\multicolumn{2}{c}{\multirow{2}{*}{\textbf{\textit{Adversarial Based Distillation}}}} & \multicolumn{3}{c}{CIFAR-10} & \multicolumn{2}{c}{ImageNet} & \multicolumn{2}{c}{LAION-A} & \multicolumn{2}{c}{MS-COCO} \\
\cmidrule(lr){3-5} \cmidrule(lr){6-7}  \cmidrule(lr){8-9}  \cmidrule(lr){10-11}
&  & 1 & 2 & 4 & 1 & 4 & 4 & 8 & 4 & 8  \\ \midrule
ADD~\cite{sauer2023adversarial} & ArXiv23 & - & - & - & - & - & - & 20.60 & 20.80 & 20.30    \\
LADD~\cite{sauer2024fast} & ArXiv24 & - & - & - & - & - & - & 19.70 & - & -  \\
\midrule
\multicolumn{2}{c}{\multirow{2}{*}{\textbf{\textit{GAN Objective}}}} & \multicolumn{3}{c}{CIFAR-10} & \multicolumn{3}{c}{CelebA-HQ-256} & \multicolumn{3}{c}{MS-COCO}  \\
\cmidrule(lr){3-5}  \cmidrule(lr){6-8}  \cmidrule(lr){9-11}
&  & 1 & 2 & 4 & 1 & 2 & 4 & 1 & 2 & 4  \\ \midrule
DDGAN~\cite{xiaotackling} & ICLR22 & 14.60 & 4.08 & 3.75 & - & 7.74 & - & - & - & - \\
SIDDMs~\cite{xu2023semi} & NeurIPS23 & - & - & 2.24 & - & 7.37 & - & 28.00 & - & 21.70 \\ 
UFOGen~\cite{xu2024ufogen} & CVPR24 & - & - & - & - & - & - & 22.50 & - & 22.10 \\
\midrule
\multicolumn{2}{c}{\multirow{2}{*}{\textbf{\textit{Truncated Diffusion}}}} & \multicolumn{3}{c}{CIFAR-10} & \multicolumn{3}{c}{ImageNet} & \multicolumn{3}{c}{LSUN-Bedroom} \\
\cmidrule(lr){3-5}   \cmidrule(lr){6-8}  \cmidrule(lr){9-11}
&  & 50 & 100 & 200 & 50 & 100 & 200 & 50 & 100 & 200  \\ \midrule
ES-DDPM~\cite{lyu2022accelerating} & ArXiv22 & - & 5.52 & 5.02 & - & 3.75 & 3.47 & - & 1.85 & 1.70  \\
TDPM~\cite{zheng2022truncated} & ArXiv22 & 2.94 & 2.88 & - & 1.77 & 1.62 & - & 4.34 & 3.98 & - \\
\bottomrule
\end{tabular}
}
\caption{Five types of \textbf{training-based} methods are summarized. NFEs and FID are presented to demonstrate the efficiency and quality of sampling methods. }
\end{table*}
\subsection{Training Based Method}
\label{subsec:training-based}

Knowledge Distillation~\cite{hinton2015distilling} is one of the most common sampling strategy in learning based method, which distills the knowledge from deterministic ODE (teacher) models to the accelerated sampling (student) models.
According to the learning objectives, these 
sampling strategies can be divided into three groups, i.e., distribution-based, trajectory-based, and GAN-based distillations.
\textbf{Distribution based distillation} strategy accelerate the sampling steps of student models by minimizing the image or latent distributions~\cite{luhman2021knowledge, salimansprogressive, meng2023distillation, song2023consistency, luo2023latent, luo2023lcm, yin2024one}.
Luhman et.al.~\cite{luhman2021knowledge} first propose Denoising Student to reduce the iterative denoising steps by knowledge distillation.
Specifically, the 100-step DDIM scheduler with pre-trained diffusion model is leveraged as the teacher model $\mathcal{M}_{t}$, which obtains deterministic $\mathbf{x}_{0}$ from the random $\mathbf{x}_{T}$.
Meanwhile, the student model $\mathcal{M}_{s}$ use one step denoising setting to accelerate the sampling process.
Next, in order to generate high quality images, the predicted distribution of student model $\mathcal{M}_{s}$ is aligned with the iterative denoised $\mathcal{x}_{0}$ from $\mathcal{M}_{t}$. 
The learning objective of $\mathcal{M}_{s}$ is formalized as:

\begin{equation}
\mathcal{L}_{s} = \mathbbm{E}_{\mathbf{x}_{T}}[D(\mathcal{M}_{s}(\mathbf{x}_{0} | \mathbf{x}_{T}), \mathcal{M}_{t}(\mathbf{x}_{0} | \mathbf{x}_{T}))],
\label{eq:dl1}
\end{equation}

$D$ is the function that measures the distance between distributions, which is implemented by KL divergence.
To inherit the learned knowledge, $\mathcal{M}_s$ is initialized with the original architecture and weight from the $\mathcal{M}_t$.
Compared with SOTA one step models e.g., NVAE~\cite{vahdat2020nvae}, BigGAN~\cite{brock2018large}, Denoising Student performs better generation ability on standard datasets.
Subsequently, considering the expensive time cost caused by the full number of sampling~\cite{luhman2021knowledge}, Progressive Distillation~\cite{salimansprogressive} is proposed to iteratively accelerate the sampling process.
During each iteration, the student model is trained to predict the noise after 2 DDIM sampling steps, and the optimized student model is utilized as the teacher model in the next iteration.
Hence the sampling number is reduced in exponential rate.
Moreover, Meng et.al.~\cite{meng2023distillation} design a two stage training method to apply the distillation strategy to classifier-free models.
In the first stage, following~\cite{ho2021classifier}, the denoised feature of teacher model is calculated by $\widetilde{\mathcal{M}}_{t}(\mathbf{z}_{t}, c) = (1+w)\mathcal{M}_{t}(\mathbf{z}_{t}, c) - w\mathcal{M}_{t}(\mathbf{z}_{t})$.
Then, the learning objective of student model is:

\begin{equation}
\mathbbm{E}_{w \sim p_{w}, t \sim U[0,1]} \left [ \omega(\lambda_{t}) ||\mathcal{M}_{s}(\mathbf{z_{t}}, c, w) - \mathcal{M}_{t}(z_{t}, c)||^{2}_{2} \right ],
\label{eq:dl2}
\end{equation}

where $p_w = U[w_{min}, w_{max}]$, $\omega(\lambda_{t})$ is the pre-specified weighting function~\cite{kingma2021variational}.
After distilling student model to fit classifier-free models, the second stage utilizes the progressively distillation strategy~\cite{salimansprogressive} to accelerate the sampling steps.
In addition to the above methods, consistency models~\cite{song2023consistency} (CMs) is a milestone in efficient inference and sampling, which proposes a remarkable consistency regularization, i.e., 

\begin{equation}
\mathcal{L} = \mathbbm{E} [\lambda (t_n) d(\mathbf{f_{\theta}} (\mathbf{x}_{t_{n+1}}, t_{n+1}), \mathbf{f_{\theta^{-}}} (\hat{x}^{\phi}_{t_n}, t_n) )],
\label{eq:os1}
\end{equation}

where $\lambda(t_n)$ denotes the weighting of $n$-th step, $d(\cdot)$ measures the distance of two distributions, which can be implemented by $L_1$, $L_2$ and LPIPS functions.
Given the distribution $\mathbf{x}$ in $t_{n+1}$-th step, $\hat{x}^{\phi}_{t_n}$ is acquired by running one discretization step of score based denoising model $\mathbf{s}_{\phi}$.
The $\mathbf{f_{\theta}}$ means the trained denoising network, the parameters $\theta$ of the network is updated by $\theta^{-}$ in an exponential moving average (EMA) manner.
Overall, CMs assums that the distribution at any time step in the PF-ODE trajectory can be directly mapped to the distribution at $t_0$.
Sequentially, LCM~\cite{luo2023latent} leverages augmented consistency function to align the diffusers with input text conditions, and further designs skipping-step technique to accelerate the convergence of denoising models.
Inspired by previous distribution matching methods~\cite{goodfellow2014generative}, DMD~\cite{yin2024one} finetunes the distilled model to learn the fake distribution of pretrained models, which enforce the generated images of student model is indistinguishable from the original teacher model.
\textbf{Trajectory based distillation} strategy accelerates sampling process by improving the trajectory of solving PF-ODE~\cite{liuflow, liu2023instaflow, zheng2023fast, fan2023optimizing, yan2024perflow}.
Rectified Flow~\cite{liuflow} proposes to rectify the trajectory from a non-linear path to a straight path, which is formally defined as:

\begin{equation}
\mathop{min}_{v} \int_{0}^{1} \mathbbm{E} [ || (X_{1} - X_{0}) - v(X_{t}, t) ||^{2} ] dt, \quad with X_{t} = t X_{1} + (1-t) X_{0},
\label{eq:os1}
\end{equation}

according to the equation, it can be observed that $X_{t}$ is the linear interpolation of $X_{0}$ and $X_{1}$, which models the shortest path between the samples.
To build the one-to-one correspondence between the samples from two distributions $\pi_{0}$ and $\pi_{1}$, they design reflow method, which first trains the sampling model using randomly selected $X_{0}$ and $X_{1}$.
Then, the first stage model is leveraged to provide accurate correspondence for training the second stage model.
Sequentially, InstaFlow~\cite{liu2023instaflow} is proposed to acquire a text conditional rectified flow models.
To further accelerate the sampling process, PeRFlow~\cite{yan2024perflow} trains a piecewise linear flow by creating $K$ times window, and follows the reflow operation to straightening each trajectory.
Similarly, DSNO~\cite{zheng2023fast} proposes a parallel decoding method, which is accomplished by Fourier neural operator (FNO)~\cite{lifourier}.
In addition to the strategy of using gradient descent algorithm based on trajectory for distillation, SFT-PG~\cite{fan2023optimizing} introduces reinforcement learning into efficient sampling.
To this end, the policy gradient is utilized to replace the gradient descent, and minimizing the integral probability metrics (IPM) to achieve better generation quality in few steps. 
\textbf{Adversarial based distillation} combines the advantages of GAN and diffusion models~\cite{sauer2023adversarial, sauer2024fast}.
Diffusion models have powerful generation capacity, which are able to generate high-quality images~\cite{rombach2022high, esser2024scaling} and videos~\cite{guoanimatediff, du2023learning}.
However, these models suffers from iteratively sampling process~\cite{zhanggddim, lu2022dpm++}, hindering their application in real-world scenes.
On the contrary, GAN models is able to generate images in single-step formulation, but often fall short of the quality, particularly artifacts~\cite{heusel2017gans, gal2022stylegan}.
Inspired by these observations, ADD~\cite{sauer2023adversarial} introduces a discriminator model~\cite{oquab2023dinov2} to optimize the accelerated sampling model.
The adversarial loss is defined as follows:

\begin{equation}
\begin{aligned}
\mathcal{L}^{D}_{adv} (\mathcal{M}_{s}(x_t, t), \phi) &= \mathbbm{E}_{x_0} \left[\sum _k \max(0,  1-\mathcal{D}_{\phi}(x_0)) + \gamma R1(\phi) \right] \\
&+ \mathbbm{E}_{\mathcal{M}_{s}} \left[\sum _k \max(0, 1+\mathcal{D}(\mathcal{M}_{s}(x_t, t))) \right],
\end{aligned}
\end{equation}

where $\mathcal{D}_{\phi}$ is the discriminator, R1 is the R1 gradient penalty~\cite{mescheder2018training}.
Meanwhile, in order to retaining the high quality generation capacity, a pre-trained diffusion model is utilized as teacher model.
Although ADD achieves fast sampling model, its denoising process is limited in the pixel level (RGB space) due to the discriminator.
Specifically, LDD utilizes DINOv2~\cite{oquab2023dinov2} as backbone of the discriminator, which cannot predict in latent space.
Moreover, the generated images are fixed to 518$\times$518 pixels.
To address the issues, LADD~\cite{sauer2024fast} unifies the teacher and discriminator, and input discriminator with latent features.
Therefore, LADD is able to produce high-resolution images with smaller storage cost.
\textbf{GAN objective} methods utilize multimodal conditional distribution to replace the rigorous Guassian distribution in the diffusion models, which is called denoising diffusion GAN~\cite{xiaotackling, xu2023semi, xu2024ufogen}.
DDGAN~\cite{xiaotackling} first proposes to train diffusion models with GAN objective, which inherits the fast sampling strength of GAN.
The most crucial observation is that only small steps achieves regorious Gaussian distribution, and larger steps result in multimodal (peak) distribution.
Therefore, to accelerate the sampling process, the multimodal conditional distribution is utilized to replace the unimodal  Gaussian distribution.
Please note that adversarial based distillation methods discriminate the generated samples and real images, while denoising diffusion GAN models use denoised latent as `real' samples.
However, DDGAN can not be applied to large scale dataset due to the non-scalability of GAN.
To this end, SIDDMs~\cite{xu2023semi} adds a loss term to explicitly match the conditional distribution.
Sequentially, UFOGen~\cite{xu2024ufogen} is proposed to achieve the one-step sampling.
Xu~et.al. thinks the failure of DDGAN and SIDDMs mainly caused by the posterior prediction in the denoising process.
In this way, the denoising diffusion GAN is able to directly match the distribution of $x_0$.
\textbf{Optimization strategy} contains methods that design acceleration strategies by introducing prior information during training and inference processes~\cite{watson2021learning, watson2022learning, dockhorn2022genie, shang2023post, he2023ptqd, lyu2022accelerating, zheng2022truncated}.
Watson et.al. introduce an dynamic programming algorithm to find the optimal discrete time schedules, which can be applied to any pre-trained DDPMs~\cite{watson2021learning}.
The method is based on the decomposability property of evidence lower bound (ELBO) that  the total ELBO is the sum of individual KL terms.
Then, they maintain two matrices $C, D \in R^{(K+1) \times (D+1)}$ to find the sampling path in K steps with minimum ELBO .
$C[k, t]$ denotes the minimum ELBO in iteration t with k steps, and $D[k, t]$ records the optimal path of the current step, the state transition equation of the dynamic programming can be formally defined as:
\begin{equation}
C[k, t] = \mathop{min} \limits _s (C[k-1, s] + L(t, s)),
D[k, t] = arg\mathop{min} \limits _s (C[k-1, s] + L(t, s)), 
\label{eq:os1}
\end{equation}
where $L(t, s)$ is the decomposed ELBO from $t$ to $s$. 
However, the metric used in~\cite{watson2021learning} has a mismatch with the quality of generated images, e.g., FID scores.
To address this issue, GGDM utilizes Kernel Inception Distance (KID) as perceptual loss to obtain high-fidelity images~\cite{watson2022learning}.
\textbf{Truncated diffusion} methods accelerate the sampling process by introducing early stop into training and inference process~\cite{lyu2022accelerating, zhengtruncated}.
The denoising process is started from non-Gaussian distributions, then we can perform only a few denoising steps to generate the high-quality images.
Specifically, the non-Gaussian distributions is obtained from existing generative models such as GAN~\cite{goodfellow2014generative, karras2019style} and VAE~\cite{kingma2013auto}, which is able to approximate the distribution of the data without expensive iteration process.
%



%

\section{Efficient Deployment and Usage}
\label{sec:6}

\begin{figure}[!t]
\centering
\includegraphics[width=1.0\linewidth]{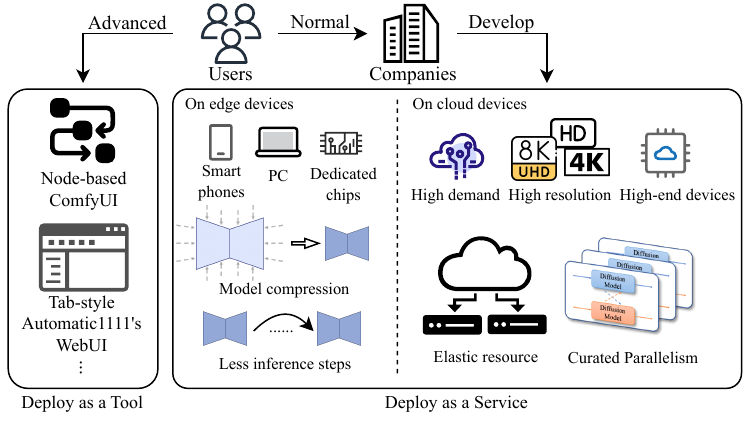}
\caption{Efficient deployment as a tool and as a service.}
\label{fig:deploy-base}
\end{figure}

The previous sections explored various efficient diffusion model techniques from a research perspective, focusing on model architecture, training and fine-tuning, sampling and inference optimizations. This section shifts focus to the \textbf{real-world deployment} and application of diffusion models. We divide the deployment and usage scenarios into two main categories: ``Efficient Deployment as a Tool’’ and ``as a Service’’, as shown in Figure.~\ref{fig:deploy-base}. The former is aimed at users who are already familiar with the fundamental processes of image generation using diffusion models, while the latter requires greater enterprise-level support to provide broader audiences with well-packaged, "one-click" image generation services.

\subsection{Efficient Deployment as a Tool}
\label{subsec:dep-tool}
In practical applications, the efficient deployment of diffusion models as tools is crucial for researchers, developers, and other AIGC practitioners. These users require a high degree of flexibility and control over the generation process to adjust and optimize model configurations across various scenarios. This type of deployment offers an environment for deep experimentation and customization, fully leveraging the potential of diffusion models. It is especially suited for tasks that require testing multiple model configurations, adjusting noise parameters, optimizing performance, or integrating custom components. Therefore, tool-based deployment typically emphasizes modular design, scalability, adaptability to diverse needs, and a high level of control.

In implementation, these tools must strike a balance between ease of use and technical depth. Professional users need an interface that is both intuitive and allows for in-depth adjustment of model parameters. Achieving this balance poses significant design challenges, requiring tools that cater to expert needs without overwhelming the user with complexity.

Taking \textbf{ComfyUI}\footnote{https://github.com/comfyanonymous/ComfyUI} as an example, it employs a ``node-based workflow interface'', allowing users to visually create and modify complex image generation processes. By connecting different nodes, users can construct each step of the model and flexibly adjust the parameters and hyperparameters of each module. This modular design is particularly well-suited for users who seek to refine and customize the generation process, especially researchers and developers who benefit from being able to track each stage of the workflow from input to output. ComfyUI's node-based architecture greatly facilitates the integration of custom models and new algorithms. Users can easily introduce new nodes, algorithms, or functional modules to experiment with. This is especially beneficial for developers, as they can flexibly swap components without needing to overhaul the entire system. Researchers, on the other hand, can quickly and conveniently compare the performance of different model components before and after adjustments. However, the flexibility of ComfyUI also makes its learning curve steeper, making it more suitable for users who have a deeper understanding of the overall diffusion model process.

In contrast, \textbf{Stable Diffusion WebUI}\footnote{https://github.com/AUTOMATIC1111/stable-diffusion-webui} (commonly referred to as \textbf{Automatic1111} or \textbf{WebUI}) offers a simple form-like interface. Users can quickly generate images by entering parameters such as prompts, number of steps, CFG scale, and image resolution. This design is particularly well-suited for users who want a fast and straightforward image generation process, especially beginners. Even though the detailed image generation workflow is hidden, WebUI still provides advanced features and customization options to meet the needs of more experienced users.
Through its plugin system, users can realize various features, such as ``inpainting’’ and personalized training tools like Textual Inversion and ControlNet. While it lacks flexibility of the node-based ComfyUI, the ease of using plugins makes it ideal for users who want to expand functionality without extensively modifying the model. Automatic1111's WebUI is more user-friendly and accessible, with its streamlined form-based interface allowing users to input parameters and generate images quickly, making it suitable for those looking for fast results. For users without a strong technical background, it offers a true "plug-and-play" experience.

These tools offer users extensive control over the generation process, from adjusting the number of diffusion steps to integrating custom plugins or models tailored for specific domains. They not only meet the needs of advanced users involved in research and development but also address the practical requirements of deployment in production environments.
When deployed in cloud environments, these tools typically provide scalable infrastructure to accommodate large-scale workflows. For instance, ComfyUI can seamlessly integrate with Amazon EKS, enabling dynamic scaling of GPU instances to meet the demands of large-scale parallel inference in the cloud. Additionally, an active user community contributes numerous resources to these tools, including comprehensive APIs and documentation, encouraging developers to create and share custom plugins. This open ecosystem not only enriches the tools' functionality but also opens up new possibilities for various applications, spanning from artistic creation to scientific research and industrial design.

\subsection{Efficient Deployment as a Service}
\label{subsec:dep-service}
Efficient Deployment as a Service is aimed at a broader user base, typically requiring neither advanced technical expertise nor local high-end computational resources. Service providers package comprehensive tools to simplify the complex processing of diffusion models into a "one-click" user experience. Their efforts are focused on optimizing the inference process and user interaction for real-world deployment scenarios on mobile and cloud platforms. The goal is to deliver faster, more stable inference services that meet the needs of everyday users, while also addressing cost control and privacy concerns.

In~\cite{chen2023speed}, Google optimizes GPU memory I/O to significantly reduce inference latency on mobile devices via two key improvements: enhanced attention modules and Winograd convolution. By using partially fused Softmax to reduce memory access for large intermediate matrices, along with FlashAttention to lower memory bandwidth pressure, the attention mechanism’s efficiency was greatly enhanced. Additionally, Winograd convolution accelerated the $3 \times 3$ convolution layers, striking a balance between computational efficiency and memory usage. Tests showed that on the \emph{Samsung S23 Ultra} and \emph{iPhone 14 Pro Max}, the latency for generating $512\text{px}$ resolution images was reduced by $52.2\%$ and $32.9\%$, respectively, with inference time dropping to under 12 seconds over 20 steps and memory usage capped at 2,093 MB.

Despite these improvements, latency remains high for interactive mobile applications. SnapFusion~\cite{li2024snapfusion} made a breakthrough by reducing inference time to under 2 seconds for text-to-image generation on mobile devices. To achieve this, SnapFusion optimized the UNet by removing redundant computations through an evolving-training framework. To further reduce inference steps, it introduced CFG-aware step distillation, greatly enhancing both efficiency and stability. Tests on the \emph{iPhone 14 Pro} demonstrated that SnapFusion can generate $512\text{px}$ images in just 2 seconds, and in experiments on the MS-COCO dataset, it achieved superior FID and CLIP scores using only 8 denoising steps, outperforming Stable Diffusion v1.5 with 50 steps.

To further optimizes both the architecture of diffusion models for mobile devices, MobileDiffusion~\cite{zhao2024mobiledif} redesigns the UNet by sharing projection matrices, replacing activation functions, and adopting separable convolutions to achieve a lightweight model. The VAE decoder is pruned for width and depth while increasing latent channels, accelerating decoding while maintaining reconstruction quality. For sampling, it introduces UFOGen's Diffusion-GAN hybrid training method~\cite{xu2024ufogen}, enabling one-step sampling. By leveraging adversarial fine-tuning and distillation techniques, the model generates high-quality images in just one step. On the iPhone 15 Pro, MobileDiffusion generates $512\text{px}$ images in under 0.2 seconds, while also supporting various downstream applications such as controlled generation (e.g., based on text, canny edge or depth map), personalized generation (e.g., Style-LoRA, Object-LoRA), and in-painting.

\begin{figure}[!t]
\centering
\includegraphics[width=1.0\linewidth]{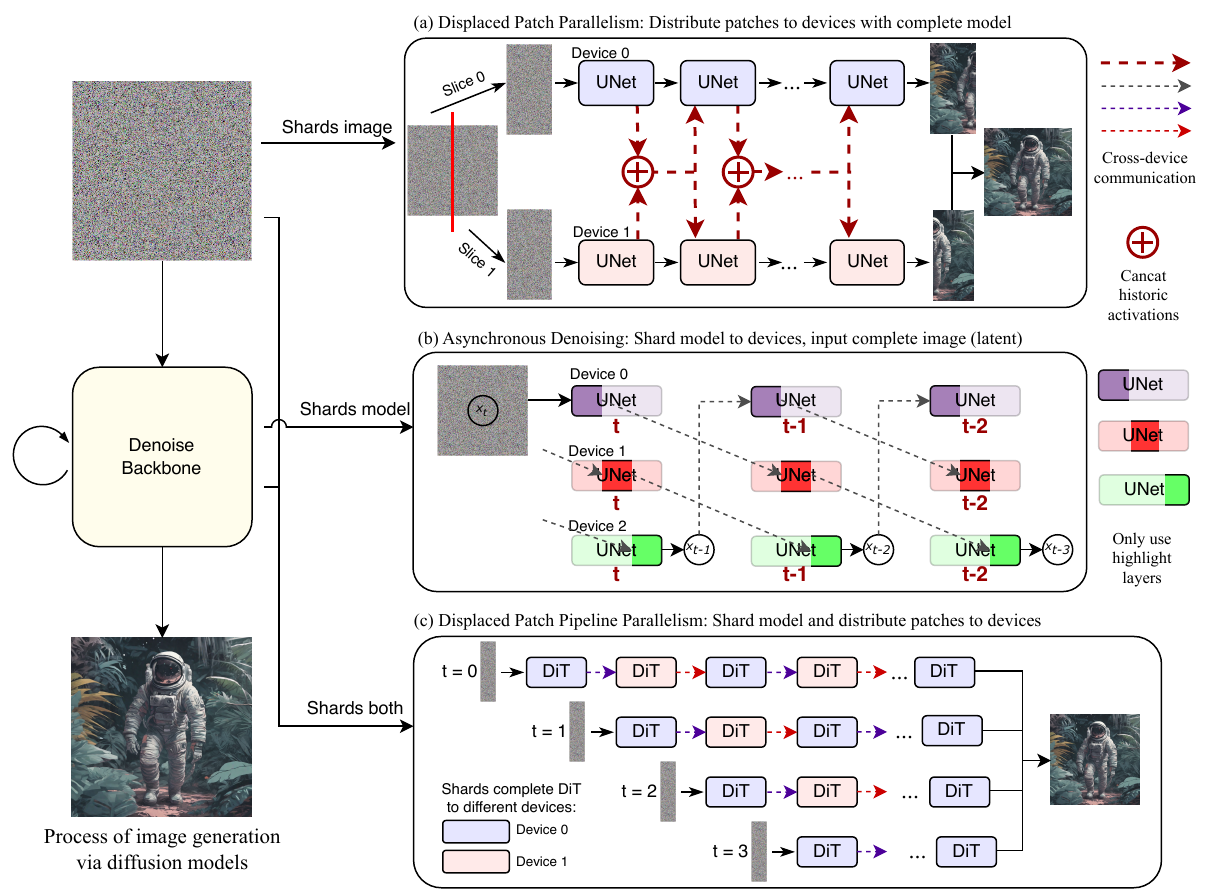}
\caption{Efficient cloud-based deployment strategies for diffusion models.}
\label{fig:deploy-cloud}
\end{figure}

However, due to the limited computational resources of mobile devices, it is difficult to achieve fast generation of high-quality, high-resolution images. Applications that need to handle large-scale tasks while requiring high-speed generation often rely on efficient deployment on cloud-based infrastructure of service providers. Cloud deployment not only leverages more powerful hardware resources to handle complex tasks but also improves the efficiency of concurrent inference through distributed computing and elastic scaling, as shown in Figure.\ref{fig:deploy-cloud}.


To achieve low-latency, high-resolution image generation without compromising image quality, DistriFusion~\cite{li2024distrifusion} focuses on parallelism across multiple GPUs. Observing the high similarity between inputs from adjacent diffusion steps, it reuses activations from previous steps to provide global context and inter-block interaction. Based on this, DistriFusion proposes Displaced Patch Parallelism, where the input image is divided into multiple patches and processed in parallel by SD-XL on different GPUs. The global results from the previous step are reused to approximate the context for the current step, while asynchronous communication prepares the global context for the next step, effectively hiding communication latency. In practice, DistriFusion achieves speedups of approximately $2.8\times$, $4.9\times$, and $6.1\times$ for generating images at $1024\text{px}$, $2048\text{px}$, and $3840\text{px}$ resolutions, respectively, using 8 A100 GPUs, without sacrificing image quality, compared to single A100 GPU processing.

To address the computational and latency challenges of generating high-resolution images with Diffusion Transformers (DiT) across multiple GPUs, PipeFusion~\cite{wang2024pipefusion} also leverages the high similarity between inputs from adjacent steps. However, applying DistriFusion method to DiT can result in inefficient memory usage due to the need for large communication buffers. To overcome this, PipeFusion introduces Displaced Patch Pipeline Parallelism. This method divides the image into patches and distributes transformer layers across different GPUs, using pipeline parallelism for computation and communication. By transmitting only the input activations of the initial layer and the output activations of the final layer via asynchronous point-to-point (P2P) communication between adjacent devices, it significantly reduces data transfer and memory usage. Tested on three GPU clusters using PCIe or NVLink, PipeFusion outperforms other parallelization techniques in terms of end-to-end latency at various resolutions. For instance, in a 4 A100 (PCIe) cluster, PipeFusion achieves latency reductions of $2.01\times$, $1.48x\times$, and $1.10\times$ at $1024\text{px}$, $2048\text{px}$, and $8192\text{px}$ resolutions, respectively. This is especially significant at $8192\text{px}$, where other methods often face ``Out Of Memory'' issues. PipeFusion dramatically lowers the required communication bandwidth, enabling the DiT model to run efficiently on GPUs connected via PCIe, without the need for costly NVLink infrastructure, thus significantly reducing operational costs for service providers.

Unlike patch-based parallel methods, AsyncDiff~\cite{chen2024asyncdiff} focuses on asynchronous parallel inference. In traditional diffusion models, denoising steps are performed sequentially, where each step’s input depends on the previous step’s output. AsyncDiff breaks this dependency chain by also leveraging the high similarity between inputs from adjacent diffusion steps, enabling parallel computation of denoising components. It introduces asynchronous denoising, model parallel strategies, and stride denoising, allowing multiple denoising steps to be processed concurrently in a single parallel round, reducing the number of parallel computation rounds and communication frequency between devices. This approach significantly improves inference speed while maintaining image quality. On four NVIDIA A5000 GPUs, AsyncDiff achieved a $4\times$ speedup on SDv2.1 with only a 0.38 reduction in CLIP score. Additionally, this method is also effective for video diffusion models, significantly reducing latency while maintaining high video quality.

\section{Applications}
\label{sec:7}
In the above analyses, we summarize efficient diffusion models by focusing on five critical components. Next, we conduct a comprehensive review of previous work, showcasing how these models have been applied in various contexts, including image synthesis, image editing, video generation, video editing, 3D synthesis, medical imaging, and bioinformatics engineering, while assessing their strengths and limitations. Based on this foundation, we propose potential development directions aimed at enhancing the efficiency and effectiveness of diffusion models in future applications.
\subsection{Image Synthesis}

\begin{figure}[htbp]
	\centering
	\subfigure[The number of relevant research works] {\includegraphics[width=.48\textwidth]{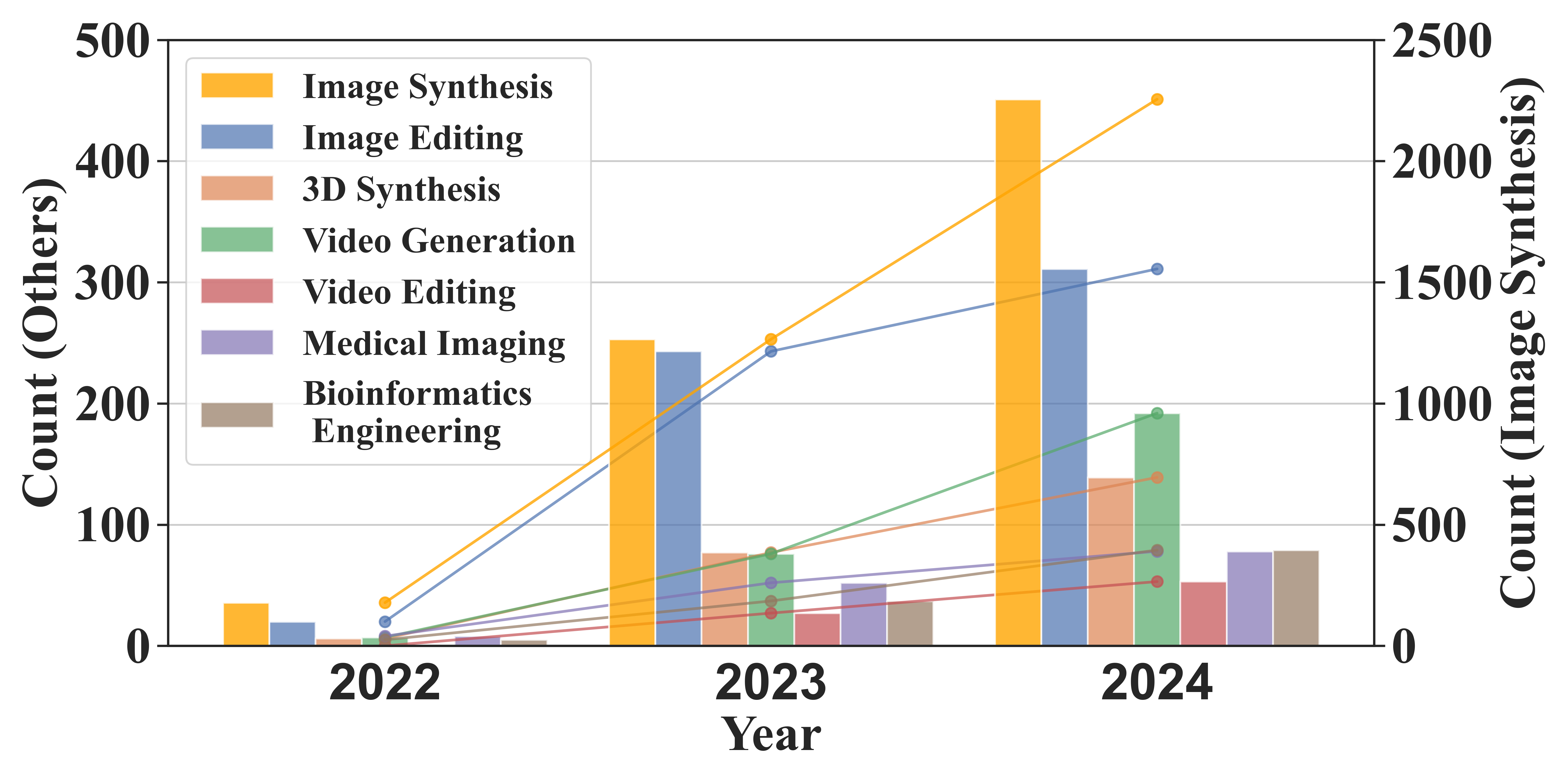}}
	\subfigure[Ratios of different tasks] {\includegraphics[width=.48\textwidth]{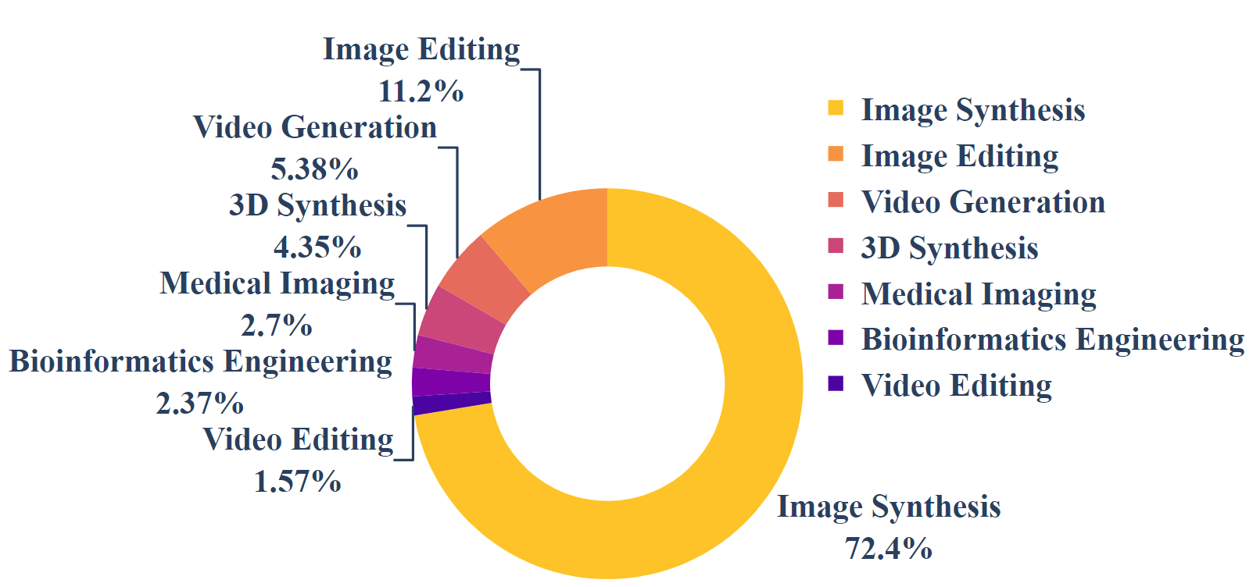}}
	\caption{The number of research papers on Efficient Diffusion Models published between 2022 and 2024}
	\label{fig_E1}
\end{figure}

\begin{table}[]
\centering
\label{tab:my-table}
\resizebox{\columnwidth}{!}{%
\begin{tabular}{@{}ccccccc@{}}
\toprule
\textbf{Application}                                                                    & \textbf{Name}                                     & \textbf{Organization}    & \textbf{State}                       & \textbf{Demo}                                                                                                                                            & \textbf{Program}                                                                                                                            & \textbf{Weight}                                                                                                                                                                                                                                                                                                                                                                 \\ \midrule
                                                                                        & FLUX.1 dev                                        & Black Forest Labs        & {\color[HTML]{009901} Open source}   & \href{https://fal.ai/models}{[demo]}                                                                                                                     & \href{https://github.com/black-forest-labs/flux}{[program]}                                                                                 & \href{https://huggingface.co/black-forest-labs/FLUX.1-dev}{[weight]}                                                                                                                                                                                                                                                                                                            \\
                                                                                        & FLUX.1 pro                                        & Black Forest Labs        & {\color[HTML]{656565} Closed Source} & \href{https://fal.ai/models}{[demo]}                                                                                                                     & -                                                                                                                                           & -                                                                                                                                                                                                                                                                                                                                                                               \\
                                                                                        & SD3-Ultra                                         & Stability AI             & {\color[HTML]{656565} Closed Source} & \href{https://colab.research.google.com/github/stability-ai/stability-sdk/blob/main/nbs/Stable\_Image\_API\_Public.ipynb\#scrollTo=ooqXnYhFhLUP}{[demo]} & -                                                                                                                                           & -                                                                                                                                                                                                                                                                                                                                                                               \\
                                                                                        & Ideogram                                          & Ideogram AI              & {\color[HTML]{656565} Closed Source} & \href{https://ideogram.ai/}{[demo]}                                                                                                                      & -                                                                                                                                           & -                                                                                                                                                                                                                                                                                                                                                                               \\
                                                                                        & FLUX.1 schnell                                    & Black Forest Labs        & {\color[HTML]{009901} Open source}   & \href{https://fal.ai/models}{[demo]}                                                                                                                     & \href{https://github.com/black-forest-labs/flux}{[program]}                                                                                 & \href{https://huggingface.co/black-forest-labs/FLUX.1-schnell}{[weight]}                                                                                                                                                                                                                                                                                                        \\
                                                                                        & Midjourney 6.0                                    & Midjourney               & {\color[HTML]{656565} Closed Source} & \href{https://docs.midjourney.com/docs/quick-start}{[demo]}                                                                                              & -                                                                                                                                           & -                                                                                                                                                                                                                                                                                                                                                                               \\
                                                                                        & DALL-E 3 HD~\cite{betker2023improving}            & OpenAI                   & {\color[HTML]{656565} Closed Source} & \href{https://chatgpt.com/}{[demo]}                                                                                                                      & -                                                                                                                                           & -                                                                                                                                                                                                                                                                                                                                                                               \\
\multirow{-8}{*}{Image Synthesis}                                                       & SD3 Medium~\cite{esser2024scaling}                & Stability AI             & {\color[HTML]{009901} Open source}   & \href{https://fireworks.ai/models/stability/sd3-medium}{[demo]}                                                                                          & \href{https://github.com/huggingface/diffusers/blob/main/docs/source/en/api/pipelines/stable\_diffusion/stable\_diffusion\_3.md}{[program]} & \href{https://huggingface.co/stabilityai/stable-diffusion-3-medium/tree/main}{[weight]}                                                                                                                                                                                                                                                                                         \\ \midrule
                                                                                        & OutfitAnyone~\cite{sun2024outfitanyone}           & Alibaba Group            & {\color[HTML]{656565} Closed Source} & \href{https://huggingface.co/spaces/HumanAIGC/OutfitAnyone}{[demo]}                                                                                      & -                                                                                                                                           & -                                                                                                                                                                                                                                                                                                                                                                               \\
                                                                                        & M\&M VTO~\cite{zhu2024m}                          & Google Research          & {\color[HTML]{330001} {\color[HTML]{656565} Closed Source}}   & -                                                                                                                                                        & -                                                                                                                                           & -                                                                                                                                                                                                                                                                                                                                                                               \\
                                                                                        & Diffuse to Choose                                 & Amazon                   & {\color[HTML]{656565} Closed Source} & \href{https://diffuse2choose.github.io/static/videos/Diffuse\_to\_Choose\_DemoReel.mp4}{[demo]}                                                          & -                                                                                                                                           & -                                                                                                                                                                                                                                                                                                                                                                               \\
                                                                                        & DEADiff~\cite{qi2024deadiff}                      & ByteDance                & {\color[HTML]{009901} Open source}   & \href{https://tianhao-qi.github.io/DEADiff/}{[demo]}                                                                                                     & \href{https://github.com/bytedance/DEADiff}{[program]}                                                                                      & \href{https://huggingface.co/qth/DEADiff/tree/main}{[weight]}                                                                                                                                                                                                                                                                                                                   \\
\multirow{-5}{*}{Image Editing}                                                         & DragDiffusion~\cite{mou2023dragondiffusion}       & ByteDance                & {\color[HTML]{009901} Open source}   & \href{https://yujun-shi.github.io/projects/dragdiffusion.html}{[demo]}                                                                                   & \href{https://github.com/Yujun-Shi/DragDiffusion}{[program]}                                                                                & -                                                                                                                                                                                                                                                                                                                                                                               \\ \midrule
                                                                                        & Sora                                              & OpenAI                   & {\color[HTML]{656565} Closed Source}                          & \href{https://openai.com/index/sora/}{[demo]}                                                                                                            & -                                                                                                                                           & -                                                                                                                                                                                                                                                                                                                                                                               \\
                                                                                        & Gen-3 Alpha                                       & Runway                   & {\color[HTML]{656565} Closed Source} & \href{https://www.youtube.com/watch?v=nByslCkykj8}{[demo]}                                                                                               & -                                                                                                                                           & -                                                                                                                                                                                                                                                                                                                                                                               \\
                                                                                        & Stable Video Diffusion~\cite{blattmann2023stable} & Stability AI             & {\color[HTML]{009901} Open source}   & \href{https://huggingface.co/spaces/multimodalart/stable-video-diffusion}{[demo]}                                                                        & \href{https://github.com/Stability-AI/generative-models}{[program]}                                                                         & \href{https://huggingface.co/stabilityai/stable-video-diffusion-img2vid-xt}{[weight]}                                                                                                                                                                                                                                                                                           \\
                                                                                        & Open-Sora                                         & HPC-AI Technology        & {\color[HTML]{009901} Open source}   & \href{https://hpcaitech.github.io/Open-Sora/}{[demo]}                                                                                                    & \href{https://github.com/hpcaitech/Open-Sora}{[program]}                                                                                    & \href{https://huggingface.co/hpcai-tech/Open-Sora}{[weight]}                                                                                                                                                                                                                                                                                                                    \\
                                                                                        & VideoCrafter~\cite{chen2023videocrafter1}         & Tencent AI Lab           & {\color[HTML]{009901} Open source}   & \href{https://huggingface.co/spaces/VideoCrafter/VideoCrafter}{[demo]}                                                                                   & \href{https://github.com/AILab-CVC/VideoCrafter}{[program]}                                                                                 & \href{https://huggingface.co/VideoCrafter/VideoCrafter2}{[weight]}                                                                                                                                                                                                                                                                                                              \\
                                                                                        & Latte~\cite{ma2024latte}                          & Shanghai AI Lab          & {\color[HTML]{009901} Open source}   & \href{https://huggingface.co/spaces/maxin-cn/Latte-1}{[demo]}                                                                                            & \href{https://github.com/Vchitect/Latte}{[program]}                                                                                         & \href{https://huggingface.co/maxin-cn/Latte-1/tree/main/transformer}{[weight]}                                                                                                                                                                                                                                                                                                  \\
                                                                                        & MagicVideo-V2~\cite{wang2024magicvideo}           & ByteDance                & {\color[HTML]{656565} Closed Source}                          & \href{https://magicvideov2.github.io/}{[demo]}                                                                                                           & -                                                                                                                                           & -                                                                                                                                                                                                                                                                                                                                                                               \\
                                                                                        & NUWA-XL~\cite{yin2023nuwa}                        & Microsoft Research Asia  & {\color[HTML]{656565} Closed Source}                          & \href{https://msra-nuwa.azurewebsites.net/\#/}{[demo]}                                                                                                   & -                                                                                                                                           & -                                                                                                                                                                                                                                                                                                                                                                               \\
                                                                                        & W.A.L.T~\cite{saharia2022photorealistic}          & Google Research          & {\color[HTML]{656565} Closed Source}                          & \href{https://walt-video-diffusion.github.io/samples.html}{[demo]}                                                                                       & -                                                                                                                                           & -                                                                                                                                                                                                                                                                                                                                                                               \\
                                                                                        & GenTron~\cite{chen2023gentron}                    & Meta AI                  & {\color[HTML]{656565} Closed Source}                          & \href{https://www.shoufachen.com/gentron\_website/}{[demo]}                                                                                              & -                                                                                                                                           & -                                                                                                                                                                                                                                                                                                                                                                               \\
\multirow{-11}{*}{Video Generation}                                                     & Text2Video-Zero~\cite{khachatryan2023text2video}  & Picsart AI Resarch       & {\color[HTML]{009901} Open source}   & \href{https://huggingface.co/spaces/PAIR/Text2Video-Zero}{[demo]}                                                                                        & \href{https://github.com/Picsart-AI-Research/Text2Video-Zero}{[program]}                                                                    & -                                                                                                                                                                                                                                                                                                                                                                               \\ \midrule
                                                                                        & ViViD~\cite{fang2024vivid}                        & Alibaba Group            & {\color[HTML]{009901} Open source}   & \href{https://alibaba-yuanjing-aigclab.github.io/ViViD/}{[demo]}                                                                                         & \href{https://github.com/alibaba-yuanjing-aigclab/ViViD}{[program]}                                                                         & \href{https://huggingface.co/guoyww/animatediff/blob/main/mm\_sd\_v15\_v2.ckpt}{[weight]}                                                                                                                                                                                                                                                                                       \\
                                                                                        & MotionEditor~\cite{tu2024motioneditor}            & Fudan University         & {\color[HTML]{009901} Open source}   & \href{https://francis-rings.github.io/MotionEditor/}{[demo]}                                                                                             & \href{https://github.com/Francis-Rings/MotionEditor?tab=readme-ov-file}{[program]}                                                          & -                                                                                                                                                                                                                                                                                                                                                                               \\
                                                                                        & FLATTEN~\cite{cong2023flatten}                    & Meta AI                  & {\color[HTML]{009901} Open source}   & -                                                                                                                                                        & \href{https://github.com/yrcong/flatten}{[program]}                                                                                         & \href{https://huggingface.co/stabilityai/stable-diffusion-2-1-base}{[weight]}                                                                                                                                                                                                                                                                                                   \\
                                                                                        & Dreamix~\cite{molad2023dreamix}                   & Google Research          & {\color[HTML]{656565} Closed Source}                          & \href{https://dreamix-video-editing.github.io/}{[demo]}                                                                                                  & -                                                                                                                                           & -                                                                                                                                                                                                                                                                                                                                                                               \\
                                                                                        & ControlVideo~\cite{zhang2023controlvideo}         & Huawei Cloud             & {\color[HTML]{009901} Open source}   & \href{https://replicate.com/cjwbw/controlvideo}{[demo]}                                                                                                  & \href{https://github.com/YBYBZhang/ControlVideo}{[program]}                                                                                 & -                                                                                                                                                                                                                                                                                                                                                                               \\
\multirow{-6}{*}{Video Editing}                                                         & Rerender\_A\_Video~\cite{yang2023rerender}        & NTU                      & {\color[HTML]{009901} Open source}   & \href{https://huggingface.co/spaces/Anonymous-sub/Rerender}{[demo]}                                                                                      & \href{https://github.com/williamyang1991/Rerender\_A\_Video}{[program]}                                                                     & -                                                                                                                                                                                                                                                                                                                                                                               \\ \midrule
                                                                                        & RodinHD~\cite{zhang2024rodinhd}                   & Microsoft Research Asia  & {\color[HTML]{009901} Open source}   & \href{https://rodinhd.github.io/}{[demo]}                                                                                                                & \href{https://github.com/RodinHD/RodinHD}{[program]}                                                                                        & -                                                                                                                                                                                                                                                                                                                                                                               \\
                                                                                        & CAT3D~\cite{gao2024cat3d}                         & Google DeepMind          & {\color[HTML]{656565} Closed Source}                          & \href{https://cat3d.github.io/gallery.html}{[demo]}                                                                                                      & -                                                                                                                                           & -                                                                                                                                                                                                                                                                                                                                                                               \\
                                                                                        & DreamFusion~\cite{poole2022dreamfusion}           & Google Research          & {\color[HTML]{656565} Closed Source}                          & \href{https://dreamfusion3d.github.io/gallery.html}{[demo]}                                                                                              & -                                                                                                                                           & -                                                                                                                                                                                                                                                                                                                                                                               \\
                                                                                        & SV3D~\cite{voleti2024sv3d}                        & Stability AI             & {\color[HTML]{656565} Closed Source}                          & \href{https://huggingface.co/stabilityai/sv3d}{[demo]}                                                                                                   & -                                                                                                                                           & -                                                                                                                                                                                                                                                                                                                                                                               \\
                                                                                        & DiffPortrait3D~\cite{gu2024diffportrait3d}        & ByteDance                & {\color[HTML]{009901} Open source}   & \href{https://freedomgu.github.io/DiffPortrait3D/}{[demo]}                                                                                               & \href{https://github.com/FreedomGu/DiffPortrait3D}{[program]}                                                                               & \href{https://drive.google.com/file/d/14qzipHghFrs4CFpVo1xW9CZt8OGLJf9t/view}{[weight]}                                                                                                                                                                                                                                                                                         \\
                                                                                        & Inpaint3D~\cite{prabhu2023inpaint3d}              & Google Research          & {\color[HTML]{656565} Closed Source}                          & \href{https://inpaint3d.github.io/}{[demo]}                                                                                                              & -                                                                                                                                           & -                                                                                                                                                                                                                                                                                                                                                                               \\
                                                                                        & TextureDreamer~\cite{yeh2024texturedreamer}       & Meta AI                  & {\color[HTML]{656565} Closed Source}                          & \href{https://texturedreamer.github.io/\#results}{[demo]}                                                                                                & -                                                                                                                                           & -                                                                                                                                                                                                                                                                                                                                                                               \\
\multirow{-8}{*}{3D Synthesis}                                                          & ViewCrafter~\cite{yu2024viewcrafter}              & Tencent AI Lab           & {\color[HTML]{009901} Open source}   & \href{https://huggingface.co/spaces/Doubiiu/ViewCrafter}{[demo]}                                                                                         & \href{https://github.com/Drexubery/ViewCrafter}{[program]}                                                                                  & \href{https://huggingface.co/Drexubery/ViewCrafter\_25/blob/main/model.ckpt}{[weight]}                                                                                                                                                                                                                                                                                          \\ \midrule
                                                                                        & AlphaFold3~\cite{AlphaFold2021}                   & DeepMind                 & {\color[HTML]{009901} Open source}   & \href{https://deepmind.google/technologies/alphafold/}{[demo]}                                                                                           & \href{https://github.com/lucidrains/alphafold3-pytorch}{[program]}                                                                          & -                                                                                                                                                                                                                                                                                                                                                                               \\
                                                                                        & DiffDock~\cite{corso2022diffdock}                 & MIT                      & {\color[HTML]{009901} Open source}   & \href{https://huggingface.co/spaces/reginabarzilaygroup/DiffDock-Web}{[demo]}                                                                            & \href{https://github.com/gcorso/DiffDock}{[program]}                                                                                        & -                                                                                                                                                                                                                                                                                                                                                                               \\
                                                                                        & RFdiffusion~\cite{watson2023novo}                 & University of Washington & {\color[HTML]{009901} Open source}   & \href{https://neurosnap.ai/service/RFdiffusion-v2}{[demo]}                                                                                               & \href{https://github.com/RosettaCommons/RFdiffusion}{[program]}                                                                             & -                                                                                                                                                                                                                                                                                                                                                                               \\
\multirow{-4}{*}{\begin{tabular}[c]{@{}c@{}}Bioinformatics \\ Engineering\end{tabular}} & DiffAb~\cite{luo2022antigen}                      & Helixon Research         & {\color[HTML]{009901} Open source}   & \href{https://neurosnap.ai/service/DiffAb\%20Antibody\%20Design}{[demo]}                                                                                 & \href{https://github.com/luost26/diffab}{[program]}                                                                                         & -                                                                                                                                                                                                                                                                                                                                                                               \\ \midrule
                                                                                        & DiffMa~\cite{wang2024soft}                        & Sichuan University       & {\color[HTML]{009901} Open source}   & -                                                                                                                                                        & \href{https://github.com/wongzbb/DiffMa-Diffusion-Mamba}{[program]}                                                                         & \href{https://huggingface.co/ZhenbinWang/DiffMa/tree/main}{[weight]}                                                                                                                                                                                                                                                                                                            \\
                                                                                        & DDM2~\cite{xiang2023ddm}                          & Stanford University      & {\color[HTML]{656565} Closed Source} & -                                                                                                                                                        & \href{https://github.com/StanfordMIMI/DDM2}{[program]}                                                                                      & -                                                                                                                                                                                                                                                                                                                                                                               \\
                                                                                        & ScoreInverseProblems~\cite{song2021solving}       & Stanford University      & {\color[HTML]{009901} Open source}   & -                                                                                                                                                        & \href{https://github.com/yang-song/score\_inverse\_problems}{[program]}                                                                     & \href{https://drive.google.com/drive/folders/19G2zfKHX2ZCVh7H\_B2BTPBNhMECZEE8H}{[weight]}                                                                                                                                                                                                                                                                                      \\
\multirow{-4}{*}{Medical Imaging}                                                       & BrLP                                              & University of Catania    & {\color[HTML]{009901} Open source}   & \href{https://www.youtube.com/watch?v=6YKz2MNM4jg}{[demo]}                                                                                               & \href{https://github.com/LemuelPuglisi/BrLP}{[program]}                                                                                     & \href{https://studentiunict-my.sharepoint.com/personal/uni399517\_studium\_unict\_it/\_layouts/15/onedrive.aspx?id=\%2Fpersonal\%2Funi399517\%5Fstudium\%5Funict\%5Fit\%2FDocuments\%2FPhD\%2FProjects\%2FBrLP\%2Dshare\%2Flatentdiffusion\%2Epth\&parent=\%2Fpersonal\%2Funi399517\%5Fstudium\%5Funict\%5Fit\%2FDocuments\%2FPhD\%2FProjects\%2FBrLP\%2Dshare\&ga=1}{[weight]} \\ \bottomrule
\end{tabular}%
}
\vspace{2pt}
\caption{State-of-the-art models across various applications}

\end{table}

Image synthesis plays an important role in computer vision and has widespread applications in fields such as artistic creation and personalized content generation. The application of diffusion models to image synthesis gains prominence with the emergence of text-to-image diffusion models ~\cite{nichol2021glide,ramesh2022hierarchical,saharia2022photorealistic,rombach2022high,balaji2022ediff,xue2024raphael}, enabling the generation of high-quality images from natural language descriptions. Subsequently, efficient fine-tuning techniques expand the application of diffusion models to various conditional image generation tasks, including the  structures ~\cite{zhang2023adding,mou2024t2i} and content~\cite{ruiz2023dreambooth,li2024blip}. Meanwhile, research into efficient sampling methods further facilitates the practical application of these technologies, driving the broader advancement of image synthesis.\par


Customized generation is an important research direction in image synthesis, aiming to achieve tailored outputs that meet specific user needs. Dreambooth ~\cite{ruiz2023dreambooth} introduces subject-driven customized generation ~\cite{chen2024subject,zhang2024ssr,ma2024subject,kumari2023multi,wei2023elite,li2024blip,han2023svdiff,xiao2023fastcomposer}, which faithfully preserves the visual content of the themes depicted in the provided samples. In addition, identity customization ~\cite{wang2024instantid,valevski2023face0,chen2023dreamidentity,ruiz2024hyperdreambooth,yan2023facestudio,li2024photomaker,peng2024portraitbooth} is achieved through the high-fidelity preservation of facial features. Moreover, some work focuses on visual text generation ~\cite{chen2024textdiffuser, zhu2023conditional, ma2023glyphdraw, yang2024glyphcontrol, zhang2024brush,tuo2023anytext,hsu2023posterlayout}
, emphasizing accurate text creation within images, which aids in producing high-quality posters. At the same time, there are also interesting developments in visual storytelling~\cite{pan2024synthesizing, liu2024intelligent, gong2023talecrafter, jeong2023zero, shen2024boosting} applications, which aim to generate a coherent series of images, such as comics, to enhance the efficiency of artistic creation. Finally, in the field of safe image generation, privacy and copyright protection techniques ~\cite{he2024diff, gandikota2023erasing, cui2023diffusionshield, kumari2023ablating, carlini2023extracting, somepalli2023understanding, somepalli2023diffusion, fernandez2023stable,ma2024safe} have become key research priorities.

\subsection{Image Editing}
Diffusion models have demonstrated powerful controllable generation capabilities, which are inherently well-suited for editing tasks that require adjustments during the generation process. Among these methods, instruction-based editing techniques ~\cite{brooks2023instructpix2pix,guo2024focus,chakrabarty2023learning,geng2024instructdiffusion,sheynin2024emu,yildirim2023inst,zhang2024hive,yang2024imagebrush,fu2023guiding,ma2024adapedit,huang2024smartedit} have the broadest applicability and align most closely with human habits. However, they are constrained by the expensive fine-tuning costs to learn the editing instructions. Therefore, some researchers have concentrated on domain-specific editing techniques ~\cite{kim2022diffusionclip, kwon2022diffusion, huang2024diffstyler, wang2023stylediffusion, xu2024cyclenet, preechakul2022diffusion, zhao2022egsde} to address this issue. On the other hand, some work focuses on fine-tuning during the inference stage to further enhance editing efficiency. This includes techniques such as text embedding fine-tuning ~\cite{mokady2023null, yang2023dynamic, wu2023uncovering, dong2023prompt}, latent variable optimization ~\cite{ mou2023dragondiffusion, shi2024dragdiffusion, hertz2023delta, kwon2022diffusion, nam2024contrastive}, and fine-tuning of the diffusion model itself ~\cite{ valevski2022unitune, choi2023custom, kawar2023imagic, zhang2023sine}. Currently, finetuning-free methods have shown significant potential for efficient editing, attracting increased research attention. To avoid fine-tuning, researchers have closely analyzed the attention layers that interact most frequently with editing control conditions and proposed the classic attention modification methods ~\cite{hertz2022prompt, parmar2023zero, cao2023masactrl, tumanyan2023plug, lu2023tf, patashnik2023localizing, lee2024conditional, park2024energy, park2024shape}.
Subsequently, sampling modification ~\cite{huberman2024edit, nie2023blessing, brack2024ledits++,miyake2023negative, han2024proxedit, zhao2023null, wallace2023edict, pan2023effective, wu2023latent, jeong2024training, epstein2023diffusion} and mask guidance ~\cite{Yu2023FISEditAT, avrahami2023blended, couairon2022diffedit, avrahami2022blended, li2024zone, yu2023inpaint} techniques were introduced to further enhance accuracy.\par

These techniques have been widely adopted across various editing scenarios. For example, the recently popular virtual try-on technology        ~\cite{zhu2024m,huang2023composer,morelli2023ladi,baldrati2023multimodal,gou2023taming,kim2024stableviton,xu2024ootdiffusion,zhu2023tryondiffusion,yang2023paint,lee2022high,chen2024anydoor,sun2024outfitanyone} on e-commerce platforms allows users to better visualize how garments will look when worn. Additionally, image style transfer technology ~\cite{zhang2023adding,mou2024t2i,ye2023ip,li2024blip,qi2024deadiff,wang2023stylediffusion,zhang2023inversion,sohn2023styledrop,huang2023composer,wang2023styleadapter} allows for the flexible generation of stylized and customized images, preserving the original content while showcasing a diverse range of visual styles. On the other hand, diffusion model-based methods have also shown outstanding performance in solving low-level vision tasks, such as super-resolution~\cite{kawar2022denoising,chung2022come,saharia2022image,li2022srdiff,shang2024resdiff,ho2022cascaded,sahak2023denoising,gao2023implicit,xia2023diffir,delbracio2023inversion,choi2021ilvr,kawar2021snips,chung2022diffusion,fei2023generative,song2023pseudoinverse,zhu2023denoising}, deblurring~\cite{kawar2022denoising,xia2023diffir,delbracio2023inversion,kawar2021snips,chung2022diffusion,fei2023generative,zhu2023denoising,wang2022zero,chen2024hierarchical,ren2022image,whang2022deblurring,murata2023gibbsddrm,yuan2018unsupervised,feng2023score}, inpainting~\cite{
chung2022come,chung2022diffusion,fei2023generative,song2023pseudoinverse,zhu2023denoising,wang2022zero,saharia2022palette,lugmayr2022repaint,zhang2023towards,mardani2023variational}, and compression artifact removal~\cite{welker2022driftrec,luo2023refusion,jin2022shadowdiffusion,guo2023shadowdiffusion}. These can be seen as a broader form of the editing process.

\subsection{Video Generation}
The essence of video is a sequence of images ordered temporally. Consequently, text-to-video synthesis techniques ~\cite{he2022latent,wang2024magicvideo,zhou2022magicvideo,esser2023structure,blattmann2023align,ho2022imagen,singer2022make,ho2022video,hong2022cogvideo,khachatryan2023text2video,qi2023fatezero,wu2023tune,chen2023gentron} based on diffusion models greatly benefit from advancements in text-to-image synthesis technology, including shared aspects such as model architecture~\cite{rombach2022high,zhou2022magicvideo}and training methods~\cite{zhang2023adding,lin2024ctrl,guo2024i2v,xu2024facechain}. 
In addition, similar to controllable image generation techniques, video generation has also integrated various control conditions, such as image-guided~\cite{chen2023videocrafter1,guo2023animatediff,wang2024videocomposer,zhang2023i2vgen}, pose-guided~\cite{zhu2024champ, xu2024magicanimate, hu2024animate, wang2023disco, ma2024follow}, motion-guided~\cite{zhu2024champ, jeong2024vmc}, sound-guided~\cite{xu2024hallo, tian2024emo, jeong2023power}, depth-guided~\cite{he2023animate, xing2024make}, and multi-modal guided~\cite{zhang2024moonshot, tang2024any, ruan2023mm} approaches. These advancements further enhance controllability and improve the efficiency of custom content creation.

As a dynamic form of images, video emphasizes the controllability of motion~\cite{guo2023animatediff,wu2023tune,wang2024videocomposer,wang2024motionctrl,yang2024direct,yin2023dragnuwa,zhao2023motiondirector,chen2023motion}, making it a crucial research direction in video generation. It allows users to precisely control motion trajectories and dynamic effects, providing greater creative freedom and more accurate visual expression. Meanwhile, character animation~\cite{hu2024animate,zhu2024champ,shafir2023human,ren2024insactor,xie2024x} is a fascinating task that aims to generate character videos from static images using driving signals. Through this process, characters can exhibit natural movements and expressions, resulting in lively and dynamic content. Additionally, world models have become a significant research focus, particularly for the field of autonomous driving~\cite{wang2023drivedreamer,wang2024driving,wen2024panacea,gao2023magicdrive,li2023drivingdiffusion}. These models show great potential for generating high-quality driving videos and designing safe driving strategies by simulating real-world scenarios.Currently, generating longer videos ~\cite{yin2023nuwa, he2022latent, wang2023gen, liu2024sora, ma2024latte} is a highly challenging task, but it holds the potential to create more complex and content-rich visual works.

\subsection{Video Editing}
Text-guided video editing aims to achieve similar goals to image editing, but with videos as the target for editing. These techniques can be categorized based on their efficiency in achieving editing capabilities. The first category involves training on large-scale video-text datasets to develop generalized editing capabilities, which is the most straightforward approach to developing generalized editing capabilities. The second category, one-shot tuning methods, refines pre-trained models using specific video instances to provide more accurate and contextually relevant video editing, offering a balanced trade-off between effectiveness and efficiency. Finally, training-free methods adapt pre-trained models in a zero-shot manner but often face challenges with spatio-temporal distortions. These issues are addressed through techniques such as feature propagation, hierarchical constraints, and attention mechanisms.\par

One of the fundamental goals of video editing is to maintain temporal consistency between frames, ensuring that the generated video appears smooth and natural. Building on this foundation, virtual try-on for videos represents a significant application that aims to enhance the user's ability to edit the content and appearance of objects~\cite{peruzzo2024vase, fang2024vivid, yang2024eva, zi2024cococo, kara2024rave, chai2023stablevideo}, allowing for a more realistic experience of different garments or accessories. Concurrently, video action editing has also garnered considerable attention~\cite{mou2024revideo, zuo2024edit, jeong2024vmc, tu2024motioneditor}, focusing on the flexible manipulation of character or object movements. Recently, research has introduced unified models that integrate these two aspects, aiming to achieve more efficient editing~\cite{kwon2024unified, bai2024uniedit, liew2023magicedit, geyer2023tokenflow,cong2023flatten,molad2023dreamix,zhang2023controlvideo,cong2023flatten}. This approach not only enhances the flexibility of editing processes but also preserves video coherence, ultimately providing users with a superior editing experience.

\subsection{3D Synthesis}
3D synthesis~\cite{poole2022dreamfusion,lin2023magic3d,wang2024prolificdreamer,liu2023zero,long2024wonder3d,chen2023fantasia3d,tang2023make,anciukevivcius2023renderdiffusion,gao2024cat3d,voleti2024sv3d,yeh2024texturedreamer,yu2024viewcrafter} is a technique used to create and combine three-dimensional images or scenes~\cite{kim2023neuralfield,huang2023diffusion,shue20233d,zhai2024commonscenes,lei2023rgbd2,po2024compositional,gao2024magicdrive3d,prabhu2023inpaint3d}, typically involving the integration of multiple 3D models, textures, and lighting effects to generate realistic 3D visuals. This technology is widely used in film production, video games, virtual reality, augmented reality, and computer graphics. Through 3D compositing, users can create highly detailed and dynamic 3D environments, enhancing visual immersion and interactive experiences.\par 
Human motion modeling~\cite{yuan2023physdiff,xu2023interdiff,lin2024motion,karunratanakul2023guided,liang2024intergen,zhang2022motiondiffuse,barquero2023belfusion,du2023avatars,chen2023executing,wei2023human} is a crucial component in animating virtual characters to mimic lifelike and dynamic human movements. This area becomes an important focus in various applications, including film production and game development. Building on the foundation of human motion modeling, 3D digital avatars~\cite{zhang2024rodinhd,wang2023rodin,cao2024dreamavatar,han2024headsculpt,huang2024dreamwaltz,gu2024diffportrait3d} take this concept further by creating digital representations of three-dimensional virtual characters. These avatars not only feature realistic appearances and personalized traits but also possess the ability to perform a wide range of actions and behaviors. Moreover, to enhance the realism and interactivity of these avatars, speech-driven gesture~\cite{zhang2023diffmotion,chen2024diffsheg,zhu2023taming,alexanderson2023listen} plays a vital role. By integrating hand, arm, and body movements with speech, this aspect of embodied human communication significantly improves both human-computer interaction and human-human digital communication.

\subsection{Medical Imaging}
Medical imaging is significantly different from traditional imaging due to its complexity and multimodality, involving various techniques such as MRI, CT, and ultrasound. Researchers optimize diffusion models based on the unique characteristics of these imaging modalities, enabling the generation of high-quality medical images that alleviate data scarcity issues and enhance the accuracy of image analysis. One of the most prominent research directions is the generation of missing imaging types, such as translating CT images to MRI ~\cite{lyu2022conversion, meng2022novel, ozbey2023unsupervised,wang2024soft}, which improves diagnostic consistency while reducing time and costs. 
Furthermore, diffusion models directly synthesize high-quality medical images ~\cite{moghadam2023morphology, pinaya2022brain, dorjsembe2022three, kim2022diffusion,  packhauser2023generation, jiang2023cola,waibel2023diffusion,puglisi2024enhancing}, thereby effectively alleviating data scarcity issues. By annotating real and high-quality synthetic data, self-supervised learning methods have shown broad potential in tasks such as medical image classification~\cite{yang2023diffmic, wolleb2022diffusion, wyatt2022anoddpm, sanchez2022healthy, wolleb2022swiss, pinaya2022fast}, segmentation~\cite{fernandez2022can, kim2022diffusion, rahman2023ambiguous, bieder2023memory}, reconstruction~\cite{chung2022score, song2021solving, xie2022measurement, peng2022towards, gungor2023adaptive, luo2022mri, cui2022self, chung2022improving}, and denoising~\cite{gong2024pet, hu2022unsupervised}, thus advancing the development of medical image analysis. Additionally, diffusion models excel in medical anomaly detection ~\cite{wolleb2022diffusion, wyatt2022anoddpm, sanchez2022healthy, wolleb2022swiss, pinaya2022fast} by generating healthy images and identifying abnormal regions, enhancing diagnostic accuracy and efficiency. These applications indicate that diffusion models hold significant promise and value in the realm of medical imaging.

\subsection{Bioinformatics Engineering}
Diffusion models emerge as highly promising tools in bioinformatics due to their strong capacity for processing high-dimensional data, generating diverse synthetic data, and flexibly adapting to various bioinformatics tasks. Diffusion models provide a more versatile approach to protein design and generation~\cite{lee2022proteinsgm, wu2024protein, gao2023diffsds, lin2023generating, trippe2022diffusion, watson2023novo, yim2023se, ingraham2023illuminating, zhang2024diffpack,AlphaFold2021,luo2022antigen,zhang2024ultramedical}, overcoming the limitations of traditional generative models that can only produce small proteins or specific domains. This capability enables scientists to design proteins with specific functional or structural characteristics, which is crucial for protein engineering and drug discovery. Additionally, in molecular design~\cite{hoogeboom2022equivariant, huang2023conditional, wu2022diffusion, luo2021predicting, zhang2023sdegen, wu2023diffmd, igashov2024equivariant,weiss2023guided, abramson2024accurate,guan2024decompdiff}, diffusion models effectively model linkers between fragmented molecular components to identify and optimize small molecules that interact with biological targets, such as enzymes or receptors, thereby accelerating the generation and evaluation of potential drug candidates. Lastly, predicting the conformations of ligands bound to proteins is essential for studying protein-ligand interactions~\cite{lin2022diffbp, schneuing2022structure, corso2022diffdock, qiao2022dynamic, jin2024unsupervised}. Diffusion models facilitate this by predicting the interaction patterns of small molecules (ligands) with specific binding sites on proteins, helping researchers quickly identify potential drug candidates and optimize their designs to enhance drug efficacy and selectivity. In summary, diffusion models play a critical role in protein design and generation, small molecule and drug design, and modeling protein-ligand interactions, significantly advancing the application of bioinformatics.


\section{Discussion and Conclusion}
\label{sec:8}

\subsection{Limitations and Future work}

Although the theories and applications of diffusion models have emerged rapidly in recent years, there are still many limitations that deserve our attention and solution for efficient diffusion models. We briefly sort out these concerns below, hoping that they can serve as potential directions for future exploration:
\begin{itemize}
    \item Despite the excellent performance, the architecture of the current diffusion models still suffer from high computational complexity caused by the attention computing, especially in 3D and video diffusion models, which leads to heavy FLOPs and inference latency. Moreover, current fine-tuned models still struggle with poor task and domain generalization. For instance, ControlNets solely excels at handling vision-conditional controllable generation and Adapters require structural redesign for each new task, which reduce their flexibility. A feasible future direction would involve designing a unified framework that can handle multimodal and multi-granularity control, enabling more efficient and versatile task adaptation without the need for extensive architectural adjustments. 
    \item By integrating MoE~\citep{cai2024survey,vats2024evolution} designs into diffusion models, dynamic networks can be created to support a more flexible generation process~\citep{zhao2024dynamic,ganjdanesh2024mixture}. By dynamically activating different expert modules according to the input data’s characteristics, the model can allocate computational resources more efficiently, thereby improving its adaptability to diverse and complex tasks.
    \item The trade-off between training cost and sampling quality is another critical issue. Up to now, most existing training-free methods still require over $10$ sampling NFEs to ensure high-fidelity. Despite training-based methods can accelerate the process, they are still limited by the high training costs, such as time, data, and GPU resources. An effective route is to integrate these above two methods for high-quality and low-consumption model training.
    \item In terms of deployment and practical application, current diffusion models still take considerable time, high computational cost and storage requirements, especially in low-resource environments. Moreover, model compression and high-quality generation is still an important balance. Last but not least, communication overhead and memory efficiency in large-scale inference require further optimization. Researchers could focus on more efficient model compression or distillation, improved cross-device communication strategies, and leveraging hardware acceleration to enhance inference speed, stability, and image quality in future works.
\end{itemize}

In addition to the above mentioned directions, researchers can also investigate speculative decoding~\citep{leviathan2023fast,chen2023accelerating} by employing a collaborative approach between small and large models to improve the efficiency of diffusion models~\citep{christopher2024speculative,pan2024t}. In this direction, a small diffusion model rapidly generates preliminary structures or rough sketches, which the larger model subsequently refines and completes them. The division of tasks will reduce the computational load on the larger model, enhancing overall efficiency while preserving generation quality.

\subsection{Conclusion}

In this study, we conduct an in-depth and comprehensive review and take a deep dive into the realm of efficient DMs literature, providing an all-encompassing view of its central challenges and themes, including foundational theories and principles, as well as their extensive practices. Our goal is to identify and highlight areas that require further research and suggest potential avenues for future studies. This survey aim to provide a comprehensive perspective on the current state of efficient diffusion models, with the hope of inspiring additional research and exciting works.
Given the dynamic nature of this field, it's possible that some recent developments may not be fully covered. To counter this, we will set up a dedicated website that uses crowdsourcing to keep up with the latest advancements. This platform is intended to serve as a continually updated source of information, promoting ongoing growth in the field.
Due to space constraints, we can't cover all technical details in depth but have provided brief overviews of the key contributions in the field. In the future, we plan to continuously update and enhance the information on our website, adding new insights as they come to light.

\newpage
{
    \small
    \bibliographystyle{unsrt}
    \bibliography{neurips_2023,arXiv-2405.10739v1/bib/2-principles,arXiv-2405.10739v1/bib/3-architecture,arXiv-2405.10739v1/bib/4-training,arXiv-2405.10739v1/bib/5-inference,arXiv-2405.10739v1/bib/6-deployment,arXiv-2405.10739v1/bib/7-application,arXiv-2405.10739v1/bib/mm_ref,arXiv-2405.10739v1/bib/llm_ref}

}

\newpage

\end{document}